\documentclass{article}

\PassOptionsToPackage{numbers, compress}{natbib}

\usepackage[preprint]{neurips_2022}
\usepackage{caption}
\usepackage{soul}
\renewcommand\parallel{\mathrel{/\mskip-2.5mu/}}

\usepackage[utf8]{inputenc} 
\usepackage[T1]{fontenc}    
\usepackage{hyperref}       
\usepackage{url}            
\usepackage{booktabs}       
\usepackage{amsfonts}       
\usepackage{nicefrac}       
\usepackage{microtype}      

\usepackage{graphicx}
\usepackage{amsmath}
\usepackage{amssymb}
\usepackage{booktabs}
\usepackage[utf8]{inputenc} 
\usepackage[T1]{fontenc}    
\usepackage{url}            
\usepackage{amsfonts}       
\usepackage{nicefrac}       
\usepackage{microtype}      
\usepackage[dvipsnames]{xcolor}
\usepackage{bbding}
\usepackage{pifont}
\usepackage{bbm}
\usepackage{soul}
\usepackage{multirow}
\usepackage{framed}
\usepackage{amsthm}
\usepackage{makecell}
\usepackage{subfigure}
\usepackage{hyperref}
\usepackage{footnotebackref}

\usepackage{graphicx}
\usepackage{verbatim}
\usepackage{amsmath}
\usepackage{amsthm}
\usepackage{amssymb}
\usepackage{mathtools} 
\usepackage{bm}
\usepackage{caption}  
\usepackage{multirow}
\usepackage{comment}
\usepackage{xcolor}

\usepackage{amsmath}
\usepackage{amssymb}
\usepackage{mathtools}
\usepackage{amsthm}

\usepackage[capitalize,noabbrev]{cleveref}

\theoremstyle{plain}
\newtheorem{theorem}{Theorem}

\newtheorem{lemma}{Lemma}

\theoremstyle{definition}

\newtheorem{assumption}{Assumption}
\theoremstyle{remark}

\makeatletter
\newtheoremstyle{namedtheoremstyle}%
  {\topsep}{\topsep}
  {\itshape}{}
  {\bfseries}{.}
  {0.5em}
  {\thmname{\@ifempty{#3}{#1}\@ifnotempty{#3}{#3}}}
\makeatother

\theoremstyle{namedtheoremstyle}

\usepackage[textsize=tiny]{todonotes}

\title{Trap of Feature Diversity in the Learning of MLPs}

\author{Dongrui Liu$^{a*}$ \quad Shaobo Wang$^b$\thanks{Equal contribution} \quad Jie Ren$^a$ \quad Kangrui Wang$^a$ \quad  Sheng Yin$^a$ \\\quad  \textbf{Huiqi Deng}$^a$ \quad \textbf{Quanshi Zhang}$^a$ \thanks{This research is done under the supervision of Dr. Quanshi Zhang. He is with the Department of Computer Science and Engineering,
the John Hopcroft Center and the MoE Key Lab of Artificial Intelligence, AI Institute, at the Shanghai Jiao Tong University, China. Correspondence to: Quanshi Zhang <zqs1022@sjtu.edu.cn>.}\\
	$^a$Shanghai Jiao Tong University \quad $^b$Harbin Institute of Technology\\
}

\begin{document}

\maketitle

\begin{abstract}
In this paper, we focus on a typical two-phase phenomenon in the learning of multi-layer perceptrons (MLPs), and we aim to explain the reason for the decrease of feature diversity in the first phase. Specifically, people find that, in the training of MLPs, the training loss does not decrease significantly until the second phase. To this end, we further explore the reason why the diversity of features over different samples keeps decreasing in the first phase, which hurts the optimization of MLPs. We explain such a phenomenon in terms of the learning dynamics of MLPs. Furthermore, we theoretically explain why four typical operations can alleviate the decrease of the feature diversity. \textit{The code will be released when the paper is accepted.}
\end{abstract}

\section{Introduction}
Deep neural networks (DNNs) have achieved significant success in various tasks. However, the essential reason for the superior performance of DNNs has not been fully investigated. Many studies aim to explain typical phenomena in DNNs, \emph{e.g.} investigating the phenomenon of the lottery ticket hypothesis \cite{frankle2018lottery}, explaining the double-descent phenomenon \cite{jacot2020implicit, nakkiran2019deep, heckel2020early}, understanding the information bottleneck hypothesis \cite{wolchover2017new, shwartz2017opening}, exploring the gradient noise and regularization \cite{simsekli2019tail, mahoney2019traditional}, and analyzing the nonlinear learning dynamics \cite{lampinen2018analytic, pennington2017resurrecting}.

\begin{figure}[t]
	\centering
	\vspace{-5pt}
	\includegraphics[width=0.98\linewidth]{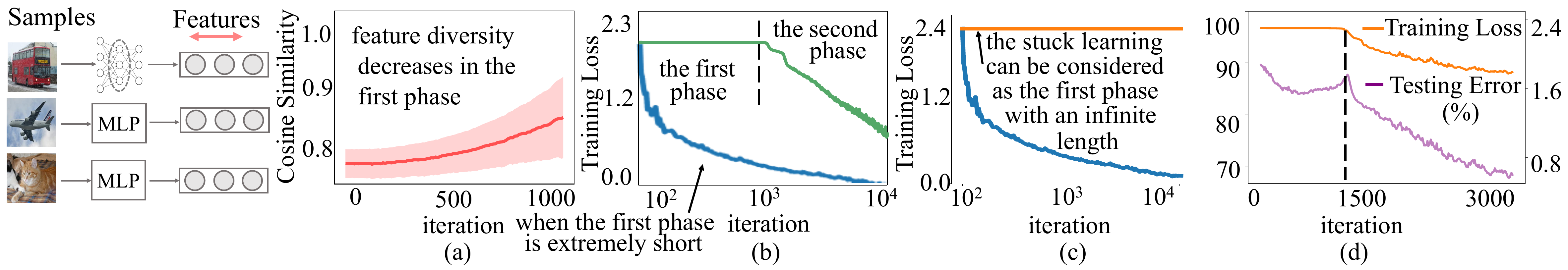}
	\vspace{-5pt}
	\caption{(a, b) We aim to explain the two-phase phenomenon in the training of MLPs, \emph{i.e.}, the reason why the cosine similarity between features of different categories keeps increasing in the first phase in (a). (c) When the loss minimization gets stuck (orange curve), we can consider it as the first phase with an infinite length. The learning-sticking problem can be solved by techniques of shortening the first phase (blue curve). (d) The two-phase phenomenon aligns the double-descent behavior.}
	\label{fig:phenomenon}
	\vspace{-17pt}
\end{figure}

In this paper, we focus on the learning dynamics of multi-layer perceptrons (MLPs). It has been widely discovered that when MLPs have many layers and the optimization is difficult, the learning process of MLPs is more likely to have two phases. As Figure \ref{fig:phenomenon}(b) shows, the first phase is usually not long, in which the training loss does not decrease. Then, in the second phase, the training loss suddenly begins to decrease. Note that when the task is simple the first phase may be extremely short (\emph{e.g.} a few iterations within an epoch), and thus cannot be observed.

To this end, we discover an interesting phenomenon in the first phase, \emph{i.e.}, as Figure \ref{fig:phenomenon}(a) shows, \textit{features of different categories become increasingly similar to each other.}
The feature diversity keeps decreasing (the cosine similarity between features keeps increasing) until the second phase.

\begin{figure*}[t]
	\centering
	\includegraphics[width=0.98\linewidth]{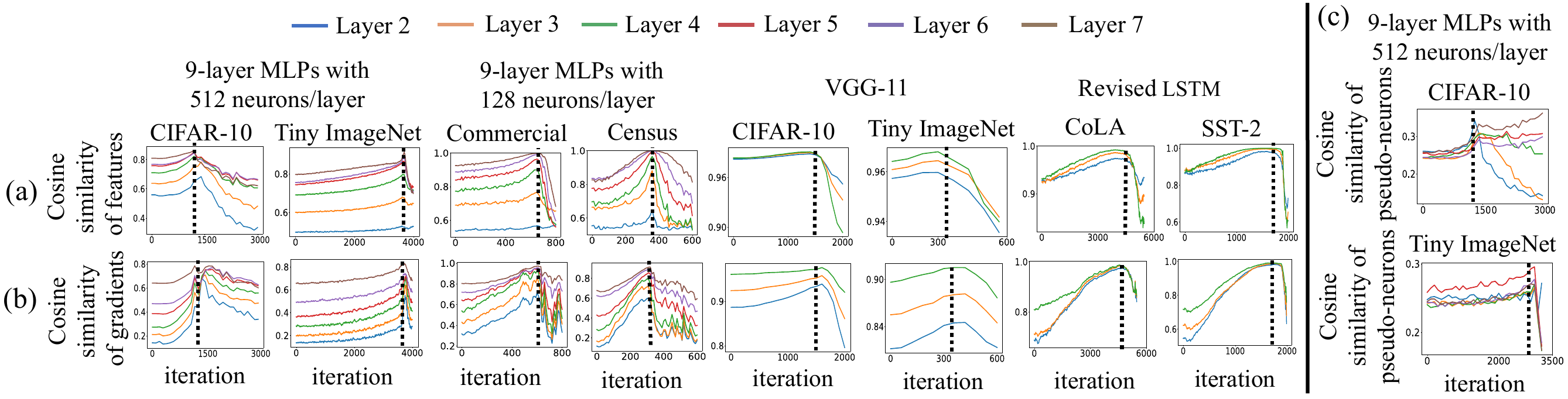}
	\vspace{-5pt}
	\caption{
	The two-phase phenomenon. (a) Cosine similarity of features between samples in different categories {\small $\mathbb{E}_{x,x'\in X} [\cos(F_t^{(l)}|_x,F_t^{(l)}|_{x'})]$} keeps increasing in the first phase (left to the dotted line), until the second phase. The low cosine similarity indicates the high diversity. (b) Cosine similarity of gradients \emph{w.r.t.} features between different samples of a category {\small $\mathbb{E}_{x,x'\in X_c} [\cos(\dot F_t^{(l)}|_x,\dot F_t^{(l)}|_{x'})]$} keeps increasing in the first phase until the second phase, where {\small$X_c$} denotes samples of the category {\small $c$}. (c) Cosine similarity of weight changes between ``pseudo-neurons'' in a layer {\small $\mathbb{E}_{x\in X} \cos(\Delta w_{t,i}^{(l)}|_x,\Delta w_{t,j}^{(l)}|_x)$} keeps increasing in the first phase.
	\textit{Please see the supplementary material for results on more DNNs.}}
	\label{fig:curve}
	\vspace{-15pt}
\end{figure*}

This phenomenon is widely shared by MLPs, convoluational neural networks, and recurrent neural networks (see both Figure \ref{fig:curve} and the supplementary material). Neural networks trained with different depths and widths, different activation functions, and different learning rates, on different types of data may all exhibit such a phenomenon, which hurts the optimization.

More crucially, the investigation and alleviation of the two-phase phenomenon are of considerable value, because this is a typical learning-sticking problem with MLPs. As Figure \ref{fig:phenomenon}(c) shows, when the loss minimization gets stuck, we can consider it as a strong first phase with an infinite length.

Therefore, we aim to investigate the optimization behavior in the first phase, \emph{i.e.}, why and how the feature diversity decreases during the early training process of the MLP. To this end, we find that in intermediate layers of the MLP, both features and parameters are mainly optimized towards a special direction, namely the primary common direction. In this study, we theoretically clarify certain dynamics of the optimization, which increases the likelihood of further enhancing such a primary common direction, just like a self-enhanced system. This may explain the decrease of the feature diversity in early iterations.

 Based on our theoretical analysis, we further discover and explain the reason why four typical operations (\emph{i.e.}, batch normalization, momentum, initialization, and $L_2$ regularization) can effectively alleviate the two-phase phenomenon. The above findings provide theoretical supports for heuristic solutions to the learning-sticking problem.

Although the decrease of feature diversity can be alleviated by traditional operations, the core contribution of
this study is to \textbf{discover and explain a fundamental yet
counter-intuitive two-phase phenomenon with the MLP, which has not been theoretically explained} for a long time.
Instead, previous studies simply owed the learning-sticking problem in the first phase to the difficulty of the training task, without insightful analysis or theoretically supported solutions.

Contributions of this study can be summarized as follows. (1) We discover the common phenomenon of feature diversity decreasing in early learning of the MLP, which has been ignored for a long time. (2) We explain this phenomenon from the perspective of learning dynamics. (3) We explain why
four types of operations can alleviate the decrease of feature diversity.

\section{Related Work}
Understanding the optimization and the representation capacity of DNNs is an important direction to explain DNNs. The information bottleneck theory \cite{wolchover2017new, shwartz2017opening} quantitatively explained the information encoded by features in intermediate layers of DNNs. \citet{xu2017information}, \citet{achille2018information}, and \citet{cheng2018evaluating} used the information bottleneck theory to evaluate and improve the DNN's representation capacity. \citet{arpit2017closer} analyzed the representation capacity of DNNs with real training data and noises. In addition, several metrics were proposed to measure the generalization capacity or robustness of DNNs, including the stiffness \cite{fort2019stiffness}, the sensitivity metrics \cite{novak2018sensitivity}, the Fourier analysis \cite{xu2018understanding}, and the CLEVER score \cite{weng2018evaluating}. Some studies focused on proving the generalization bound of DNNs \cite{long2019generalization, li2018tighter}. In comparison, we explain the MLP from the perspective of the learning dynamics, \emph{i.e.}, we explain the decrease of feature diversity in early iterations of the MLP.

Analyzing the learning dynamics is another perspective to understand DNNs. Many studies analyzed the local minima in the optimization landscape of linear networks \cite{baldi1989neural, saxe2013exact, hardt2016identity, daniely2016toward} and nonlinear networks \cite{choromanska2015loss, kawaguchi2016deep, safran2018spurious}.
Some studies discussed the convergence rate of gradient descent on separable data \cite{soudry2018implicit, xu2018convergence, nacson2019convergence}. \citet{hoffer2017train} and \citet{jastrzkebski2017three} have investigated the effects of the batch size and the learning rate on SGD dynamics. In addition, some studies analyzed the dynamics of gradient descent in the overparameterization regime \cite{arora2018optimization, jacot2018neural, lee2018deep, du2018gradient}. Unlike previous studies, we analyze the learning dynamics of features and weights of the MLP, in order to explain the decrease of feature diversity in the first phase.

\section{Discovering the decrease of feature diversity}
\label{se:phenomenon}

The training process of the MLP can usually be divided into the following two phases, according to the training loss. As Figure \ref{fig:phenomenon}(b) shows, the training loss does not decrease significantly in the first phase, and the training loss suddenly begins to decrease in the second phase. In this paper, \textbf{we discover an interesting phenomenon in the first phase that both the diversity of intermediate-layer features over different samples and the diversity of gradients \emph{w.r.t.} features keep decreasing.} In other words, in the first phase, the cosine similarity between features and the cosine similarity between feature gradients keep increasing.

We consider an MLP $f$ with {\small $L$} concatenated linear layers, each being followed by a ReLU layer. Only the last linear layer is followed by a softmax operation. Let {\small $W_t^{(l)} \in \mathbb{R}^{h\times d}$} denote the weight matrix of the $l$-th linear layer with $h$ neurons {\small $(1 \leqslant l \leqslant L)$}, and {\small $W_t^{(l)}$} has been learned for $t$ iterations. 
Given a specific input sample $x$,
the layer-wise forward propagation in the $l$-th layer is represented as
\vspace{-4pt}
\begin{equation}
\label{eq:140}
\small
    F_{t}^{(l)}=\text{ReLU}(W_{t}^{(l)}F_{t}^{(l-1)})=D_t^{(l)}W_{t}^{(l)}F_{t}^{(l-1)},
\end{equation}

\vspace{-6pt}
where {\small $F_{t}^{(l)} \!\!\! \in\! \mathbb{R}^h$} denotes the output feature of the $l$-th layer after the $t$-th iteration. {\small $D_t^{(l)}$} denotes a diagonal matrix, which represents gating states in the ReLU layer and {\small $D^{(l)}_{t,(i,i)}\in\{0,1\}$}.

\textbf{Visualizing that the two-phase phenomenon is widely shared by different DNNs learned for different tasks.} We observed such a two-phase phenomenon on  MLPs, VGG-11 \cite{simonyan2014very}, and the revised long short-term memory (LSTM) on different types of data, including image data (MNIST \cite{lecun1998gradient}, CIFAR-10 \cite{krizhevsky2009learning}, and the Tiny ImageNet dataset \cite{le2015tiny}), tabular data (two UCI datasets of census income and TV news \cite{asuncion2007uci}), and natural language data (CoLA \cite{warstadt2019neural}, SST-2 \cite{socher2013recursive}, and AGNews \cite{del2005ranking}). We also observed MLPs with Leaky ReLU layers \cite{maas2013rectifier}, with different learning rates, and with different batch sizes. Figure~\ref{fig:curve}(a,b) shows the two-phase phenomenon on some DNNs, and please see the supplementary material for results on more DNNs. Specifically, given two input samples {\small $x_1$} and {\small $x_2$},
the feature similarity between {\small $x_1$} and {\small $x_2$} {\small $\cos(F_t^{(l)}|_{x_1}, F_t^{(l)}|_{x_2})$}, and the cosine similarity of gradients {\small $\cos(\dot F_t^{(l)}|_{x_1}, \dot F_t^{(l)}|_{x_2})$} keep increasing, which demonstrates the phenomenon. {\small $\dot F_{t}^{(l)}$} denotes the gradient of the loss \emph{w.r.t.} the feature {\small $F_{t}^{(l)}$} in Eq. \eqref{eq:140}.

Note that such a decrease of feature diversity sometimes appears in very early epochs (or iterations) of the training process. More crucially, as Figure \ref{fig:phenomenon}(c) shows, when the loss directly decreases in the first epoch, it can be considered as a short first phase in very early iterations. Besides, the learning process is sometimes stuck when the task is difficult. Then \textbf{we can consider the learning-sticking problem} as the first phase with an infinite length (please see the supplementary material for more discussions).

\textbf{Connection to the epoch-wise double descent.} Figure \ref{fig:phenomenon}(d) shows that the epoch-wise double-descent behavior \cite{nakkiran2019deep, heckel2020early} is temporally aligned with the first phase in the aforementioned two-phase phenomenon. Please see the supplementary material for more discussions. Instead of explaining the epoch-wise double descent, in this paper, we mainly explain the decrease of the feature diversity.

\section{Explaining the dynamics of the decrease of feature diversity}
\label{contribuation}

\begin{figure}[t]
    \centering
    \begin{minipage}[c]{.8\linewidth}
    \centering
        \includegraphics[width=0.99\linewidth]{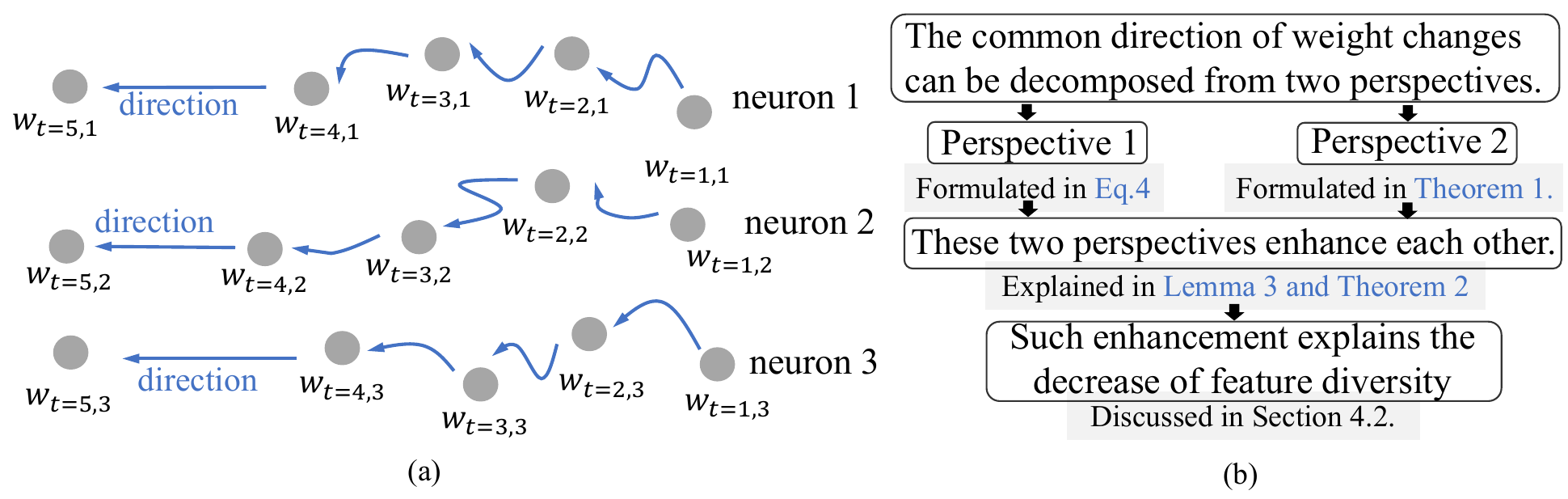}
    \end{minipage}
    \vspace{-5pt}
    \hfill
    \begin{minipage}{.185\linewidth}
        \vspace{5pt}
        \caption{(a) Weights of neurons are changed towards a common direction. (b) The logic of explaining the decrease of feature diversity.}
        \label{fig:logic}
    \end{minipage}
    \vspace{-5pt}
\end{figure}

In this section, we aim to investigate certain dynamics of network parameters, so as to explain the condition that may boost the likelihood of decreasing the feature diversity in early epochs under some common assumptions. In Section \ref{disentangling}, we find that the decreasing diversity of feature gradients over different samples is owing to the phenomenon that different neurons in a layer are optimized towards a common direction in the first phase. Then, in Section \ref{enhance}, we further clarify learning dynamics that may potentially enhance the significance of the common direction, just like a self-enhanced system. The self-enhanced common direction boosts the likelihood of the feature diversity's decreasing.
The overall logic of the explanation is illustrated in Figure \ref{fig:logic}(b). In Section \ref{alleviate}, we explain why four types of operations can alleviate the decrease of feature diversity based on our analysis.

\subsection{Two perspectives to analyze the common direction of learning effects}
\label{disentangling}

\textbf{The decreasing diversity of feature gradients over different samples is owing to the phenomenon that different neurons in a layer are optimized towards a common direction in the first phase.} For example, as Figure \ref{fig:logic}(a) shows, at the beginning of the learning, different neurons are optimized towards different directions. Along with the learning process, different neurons gradually tend to be optimized along a similar direction. According to the forward propagation in Eq. \eqref{eq:140}, feature gradients {\small $\dot F_{t}^{(l)}$} at the $l$-th layer during the back propagation can be rewritten as
\vspace{-2pt}
\begin{equation}
\small
\dot F_t^{(l-1)}=W_t^{(l)^\top}D_t^{(l)}\dot F_t^{(l)}.
\label{eq:138}
\end{equation}

\vspace{-8pt}
In this way, we can consider feature gradients {\small $\dot F_{t}^{(l-1)}$} as the result of a pseudo-forward propagation. In the pseudo-forward propagation, the input is the gradient \emph{w.r.t.} features of the {$l$}-th layer {\small $\dot F_{t}^{(l)}$}, and the output is the gradient of the {\small $(l\!-\!1)$}-th layer {\small $\dot F_{t}^{(l-1)}  \in \mathbb{R}^{d}$}. Accordingly, the equivalent weight matrix is given as {\small $W_t^{(l)^\top}\!\!\!=[{w_{t,1}^{(l)}}, w_{t,2}^{(l)}, \cdots, w_{t,d}^{(l)}]^{\top}\!\!\! \in \mathbb{R}^{d\times h}$}, which consists of $d$ ``pseudo-neurons.''

Before explaining the common direction phenomenon, let us first clarify that the increasing similarity between feature gradients {\small $\dot F_{t}^{(l-1)}$} of different samples is caused by the experimental observation that different ``pseudo-neurons'' are approximately optimized to a common direction. Let us consider the following conjecture that different ``pseudo-neurons'' {\small $[{w_{t,1}^{(l)}}, w_{t,2}^{(l)}, \cdots, w_{t,d}^{(l)}]^{\top}$} have been roughly optimized towards a common direction {\small $ C^{(l)}\!\! \in \mathbb{R}^{h}$} for many epochs. Then, we can roughly represent
{\small $w_{t,i}^{(l)} =\beta_i C^{(l)} + \bm{\epsilon}_i$}, where {\small $\beta_i \in \mathbb{R}$}, and {\small $\bm{\epsilon}_i \in \mathbb{R}^{h}$}. Here, {\small $\bm{\epsilon}_i \in \mathbb{R}^{h}$} denotes a small residual. According to Eq. \eqref{eq:138}, we have
\begin{equation}
\small
\label{eq:232}
\dot F_t^{(l-1)}=(C^{(l)^\top}D_t^{(l)}\dot F_t^{(l)}) \cdot \bm{\beta} +  \bm{\epsilon} D_t^{(l)}\dot F_t^{(l)}
\end{equation}
Thus, if {\small $\bm{\beta}$} is large enough (\emph{i.e.}, keeping optimizing {\small $W_t^{(l)^\top}$} along the common direction {\small $ C^{(l)}$} for a long time), then the feature gradients {\small $ \dot F_t^{(l-1)}$} of different samples will be roughly parallel to the same vector {\small $\bm{\beta}$}. This is because {\small $C^{(l)^\top}D_t^{(l)}\dot F_t^{(l)}$} is a scalar and the term {\small $\bm{\epsilon} D_t^{(l)}\dot F_t^{(l)}$} is small. In other words, the diversity between feature gradients {\small $\dot F_{t}^{(l-1)}$} of different samples decreases. Here, {\small $\bm{\beta} = [\beta_1, \beta_2, \cdots, \beta_d]$}, and {\small $\bm{\epsilon} = [\bm{\epsilon}_1, \bm{\epsilon}_2, \cdots, \bm{\epsilon}_d]^{\top}$}.

\textbf{Therefore, the first core task of proving the decreasing diversity of feature gradients is to explain the conjecture of the common optimization direction shared by different ``pseudo-neurons.''} To this end, let us first propose two perspectives to disentangle such a common direction.

\textbf{Perspective 1.} Perspective 1 focuses on the weight change in a certain layer. For clarity, we omit the superscript $(l)$ to simplify the notation in the following paragraphs in Section \ref{disentangling}, \emph{i.e.}, {\small $\Delta w^{(l)}_{t,i} \overset{\text{simplify}}{\longrightarrow}\Delta w_{t,i},\Delta W^{(l)}_{t} \!\overset{\text{simplify}}{\longrightarrow}\!\Delta W_{t}, C^{(l)} \overset{\text{simplify}}{\longrightarrow} C$}.
Let {\small $\Delta W_{t}^{\top}=[\Delta w_{t,1},\Delta w_{t,2}, \cdots, \Delta w_{t,d}]^\top$} denote weight changes of $d$ ``pseudo-neurons'' in the $l$-th layer. We decompose {\small $\Delta W_{t}^{\top}$} into the component along a common direction {\small $C$} and a component along other directions as follows.
\vspace{-3pt}
\begin{small}
\begin{equation}
	\Delta W_{t}^{\top}=\Delta V_{t}C^\top+\Delta \varepsilon_{t}, 
	\label{eq:146}
\end{equation}
\end{small}

\vspace{-6pt}
where {\small $\! \Delta V_{t} =[\Delta v_{t,1},\Delta v_{t,2},\cdots,\Delta v_{t,d}]\in \mathbb{R}^d$} denotes the coefficient vector for weight changes of different ``pseudo-neurons'' along the common direction {\small$C$}, \emph{i.e.}, using {\small $\Delta v_{t,i}C^\top$} to approximate {\small $\Delta w_{t,i}^\top$}. Specifically, {\small $\Delta \varepsilon_{t}$} is relatively small ``noise'' term, which is orthogonal to {\small $C$}, \emph{i.e.}, {\small $\Delta \varepsilon_{t}C=\bm 0$}. In this way, the common direction {\small $C$} can determined by minimizing the ``noise'' term over different samples across different iterations, as follows.
\begin{equation}
\small
{\mathop{\min}}_{C, \Delta V_{t}|_x} \left(\mathbb{E}_{t\in [T_\text{start}, T_\text{end}]}\mathbb{E}_{x \in X} \left\|\Delta\varepsilon_{t}|_x\right\|_{F}^{2} \right), ~~~
\text{s.t.}~~~\Delta \varepsilon_{t}|_x= \Delta W_{t}^\top|_x -\Delta V_{t}|_xC^\top
\label{eq:215}
\end{equation}

\vspace{-2pt}
\begin{lemma}
	(Proof in the supplementary material) For the decomposition {\small $\Delta W_{t}^{\top}\!\!=\!\!\Delta V_{t}C^\top+\Delta \varepsilon_{t}$},
	given weight changes over different samples {\small $\Delta W_{t}^{\top}$}, we can compute the common direction {\small $C$} in Eq. \eqref{eq:215} and obtain {\small $\Delta V_{t}=\frac{\Delta W_{t}^{\top} C}{C^{\top}C}$} and {\small $\Delta\varepsilon_{t}$}= {\small $\Delta W_{t}^{\top}-\Delta W_{t}^{\top}\frac{C C^{\top}}{C^{\top} C} $}  \emph{s.t.} {\small $\Delta\varepsilon_{t}C=\bm 0$}. Such settings minimize {\small $\|\Delta\varepsilon_{t}\|_{F}$}.
	\label{lemma1}
\end{lemma}

\begin{lemma}
	(\textbf{We can also decompose the weight {\small $W_{t}^{(l)}$} into the component along the common direction {\small $C$} and the component {\small $\varepsilon_{t}$} in other directions}. Proof is in the supplementary material.)
	Given the weight {\small $ W_{t}^{\top}$} and the common direction {\small $C$}, the decomposition {\small $ W_{t}^{\top}= V_{t}C^\top+\varepsilon_{t}$} can be conducted as {\small $ V_{t}=\frac{ W_{t}^{\top} C}{C^{\top}C}$} and {\small $\varepsilon_{t}$}= {\small $ W_{t}^{\top}-W_{t}^{\top}\frac{CC^{\top}}{C^{\top} C}$}  \emph{s.t.} {\small $\varepsilon_{t}C=\bm 0$}. Such settings minimize {\small $\|\varepsilon_{t}\|_{F}$}.
	\label{lemma2}
\end{lemma}

\begin{figure}[t]
	\centering
	\includegraphics[width=0.95\linewidth]{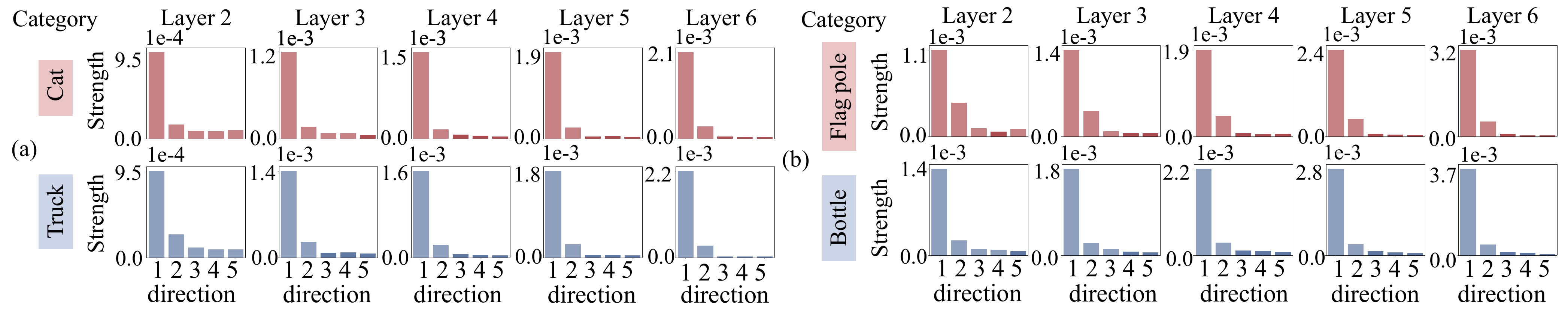}
	\vspace{-3pt}
	\caption{The strength of different common directions in the (a) CIFAR-10 dataset, (b) Tiny ImageNet dataset. We trained 9-layer MLPs, where each layer of the MLP had 512 neurons. We illustrated results on the two categories with the highest training accuracies. {\small $s_{i} = \|C_{i}\Delta \overline V_{i}^{\top}\|_{F} $} measures the strength of weight changes along the $i$-th common direction, where {\small $\Delta \overline V_{i}=\mathbb{E}_t [\Delta \overline V_{i,t}]$}.
 		The strength of the primary direction was much greater than the strength of other directions. Please see the supplementary material for results on the MNIST.}
	\label{fig:c}
	\vspace{-13pt}
\end{figure}

\textit{$\bullet$ Experimental verification of the strength of the primary common direction {\small $C$}}.
To this end, let us focus on the average weight change over different samples {\small $\Delta \overline W_t=\mathbb{E}_{x \in X} \Delta W_{t}|_x$.}
Then, we decompose {\small $\Delta \overline W_t$} into components along five common directions as {\small $\Delta \overline W_t= C_{1}\Delta \overline V_{1,t}^{\top} + C_{2} \Delta \overline V_{2,t}^{\top} + \cdots +  C_{5} \Delta \overline V_{5,t}^{\top} + \Delta \overline \varepsilon_{5,t}^{\top}$}, where {\small $C_{1}$}={\small $C$} is termed the \textit{primary common direction}. {\small $C_{2}, C_{3}, C_{4}$} and {\small $C_{5}$} represent the second, third, forth, and fifth common directions, respectively. {\small $C_{1}$}, {\small $C_{2}$}, {\small $C_{3}$}, {\small $C_{4}$}, and {\small $C_{5}$} are orthogonal to each other.
{\small $C_{i}$} and {\small $\Delta \overline V_{i,t}$} are computed based on Eq. (\ref{eq:215}) when we remove the first {\small $i-1$} components along the direction {\small $C, \cdots, C_{i-1}$} from the {\small $\Delta \overline W_t$}.
Figure \ref{fig:c} shows that the strength of the primary common component {\small $C_{1}\Delta \overline V_{1}^{\top}$} was approximately ten times greater than the strength of the secondary common component {\small $C_{2}\Delta \overline V_{2}^{\top}$}. Please see the supplementary material for more discussions.

\textbf{Perspective 2 by using gradients \emph{w.r.t.} features {\small $\dot F_t^{(l+1)}$}.}

We decompose the weight change by considering the influence of the common direction of the upper layer {\small $C^{(l+1)}$}. In order to distinguish variables belonging to different layers, we add the superscript {\small $(l)$} back to {\small$\Delta W_{t}^{(l)}, \Delta V_t^{(l)}$}, and {\small $\Delta \varepsilon_t^{(l)}$} to denote the layer in the following paragraphs.
\begin{theorem}
\label{theorem1}
	(Proof in the supplementary material) The weight change made by a sample  can be decomposed into {$(h+1)$} terms after the {$t$}-th iteration as follows.
	\vspace{-2pt}
	\begin{equation}
	\label{eq:340}
	 \small
	 \Delta W_{t}^{(l)}=  \Delta W_{\text{\rm{primary}},t}^{(l)}+ \sum\nolimits_{k=1}^h\Delta W_{\text{\rm{noise}}, t}^{(l,k)} \overset{\text{\rm{rewritten}}}{=\!\!=\!\!=\!\!=} \Gamma_t^{(l)} F_{t}^{(l-1)^{\top}}+ \kappa_t^{(l)^\top},
	\end{equation}
	
	\vspace{-6pt}
 where {\small $\Delta W_{\text{\rm{primary}},t}^{(l)}\!\!\!= D_{t}^{(l)}V_{t}^{(l+1)} $ $C^{(l+1)^{\top}} C^{(l+1)} \Delta V_{t}^{(l+1)^{\top}} F_{t}^{(l)} F_{t}^{(l-1)^{\top}}/ \|F_{t}^{(l)}\|_{2}^{2}$} denotes the component along the primary common direction,
 	and {\small $\! \Delta W_{\text{\rm{noise}}, t}^{(l,k)}\!\!=$ $D_{t}^{(l)}\varepsilon_{t}^{(l+1,k)} \Delta \varepsilon_{t}^{(l+1)^\top} F_{t}^{(l)}F_{t}^{(l-1)^{\top}}/\|F_{t}^{(l)}\|_{2}^{2}$} denotes the component along the $k$-th common direction in the noise term. {\small $\varepsilon_{t}^{(l+1, k)}\!\!\!= \boldsymbol{\Sigma}_{kk}\mathcal{U}_k\mathcal{V}_k^{\top}$}, where the SVD of {\small $ \varepsilon_{t}^{(l+1)} \in \mathbb{R}^{h\times h'} \!\!$} is given as {\small $\varepsilon_{t}^{(l+1)}=\mathcal{U}\boldsymbol{\Sigma}\mathcal{V}^{\top}$} {\small $(h \leq h')$}, and {\small $\boldsymbol{\Sigma}_{kk}$} denotes the $k$-th singular value {\small $\in \mathbb{R}$}. {\small $\varepsilon_{t}^{(l+1)}=\sum_k \varepsilon_{t}^{(l+1, k)}$}. {\small $\mathcal{U}_k$} and {\small $\mathcal{V}_k$} denote the $k$-th column of the matrix {\small $\mathcal{U}$} and {\small $\mathcal{V}$}, respectively. Besides, we have {\small $\forall k\in\{1,2,\ldots,h\},~ \mathcal{U}_k^{\top}C^{(l+1)}=0$.} Consequently, we have {\small $\Gamma_t^{(l)}\!\!\!= {D_{t}^{(l)}V_{t}^{(l+1)}C^{(l+1)^{\top}} C^{(l+1)} \Delta V_{t}^{(l+1)^{\top}} F_{t}^{(l)}} /\|F_{t}^{(l)}\|_{2}^{2} \in \mathbb{R}^{h}$}, and {\small $\kappa_t^{(l)^\top}\!\!\!=D_{t}^{(l)}$}{\small$ \varepsilon_{t}^{(l+1)}$}{\small $  \Delta \varepsilon_{t}^{(l+1)^\top}F_{t}^{(l)}F_{t}^{(l-1)^{\top}}\!\!\!\! /\|F_{t}^{(l)}\|_{2}^{2}$}{\small $  \in \mathbb{R}^{h \times d}$}.
\end{theorem}

\begin{table*}[]
	\renewcommand{\arraystretch}{1}
	\caption{ Strength of components of weight changes along the common direction and other directions. We trained 9-layer MLPs on the CIFAR-10 dataset and the Tiny ImageNet dataset, respectively. Each layer of the MLP had 512 neurons. The strength of the primary common direction was much greater than those of other directions. The supplementary material provides results on the MNIST dataset and explains the phenomenon that {\small $S_{1}^{(l)}$}, {\small $S_{2}^{(l)}$}, and {\small $S_{3}^{(l)}$} does not decrease monotonically.}
	
	\centering
	\resizebox{0.95\linewidth}{!}{
		
		\begin{tabular}{|c|c|ccccc|ccccc|}
			\hline
			\multirow{6}{*}{\rotatebox{90}{CIFAR-10$\quad\,$}}      & Category                     & \multicolumn{5}{c|}{Cat}                                                                                         & \multicolumn{5}{c|}{Truck}                                                                                       \\ \cline{2-12}
			& $S$ {\tiny ($\times 10^{-3}$)} & Layer 2              & Layer 3              & Layer 4              & Layer 5              & Layer 6              & Layer 2              & Layer 3              & Layer 4              & Layer 5              & Layer 6              \\ \cline{2-12}
			& $S_{\text{primary}}^{(l)}$   & ${154.0}_{\pm 17.1}$ & ${176.5}_{\pm 16.8}$ & ${201.6}_{\pm 18.7}$ & ${253.6}_{\pm 24.6}$ & ${277.4}_{\pm 25.6}$ & ${169.9}_{\pm 20.8}$ & ${208.1}_{\pm 21.5}$ & ${223.6}_{\pm 20.1}$ & ${248.4}_{\pm 19.2}$ & ${281.5}_{\pm 20.4}$ \\
			& $S_{1}^{(l)}$                & ${11.5}_{\pm 1.5}$   & ${13.0}_{\pm 0.9}$   & ${11.6}_{\pm 1.7}$   & ${16.1}_{\pm 1.8}$   & ${9.0}_{\pm 0.8}$    & ${15.6}_{\pm 2.1}$   & ${14.0}_{\pm 1.8}$   & ${14.3}_{\pm 1.1}$   & ${14.3}_{\pm 1.7}$   & ${10.0}_{\pm 1.1}$   \\
			& $S_{2}^{(l)}$                & ${12.7}_{\pm 1.7}$   & ${11.9}_{\pm 1.3}$   & ${10.9}_{\pm 1.3}$   & ${11.9}_{\pm 0.8}$   & ${8.8}_{\pm 1.1}$    & ${14.4}_{\pm 1.4}$   & ${15.1}_{\pm 2.0}$   & ${11.3}_{\pm 1.4}$   & ${12.3}_{\pm 0.9}$   & ${12.9}_{\pm 1.2}$   \\
			& $S_{3}^{(l)}$                & ${11.0}_{\pm 1.1}$   & ${14.4}_{\pm 1.7}$   & ${12.5}_{\pm 2.2}$   & ${13.9}_{\pm 1.7}$   & ${8.6}_{\pm 1.1}$    & ${14.3}_{\pm 2.2}$   & ${12.4}_{\pm 1.9}$   & ${12.8}_{\pm 1.6}$   & ${13.1}_{\pm 1.2}$   & ${9.7}_{\pm 1.0}$    \\ \hline
			\multirow{6}{*}{\rotatebox{90}{Tiny ImageNet$\,\,\,\,$  }} & Category                     & \multicolumn{5}{c|}{Flagpole}                                                                                    & \multicolumn{5}{c|}{Bottle}                                                                                      \\ \cline{2-12}
			& $S$ {\tiny ($\times 10^{-3}$)} & Layer 2              & Layer 3              & Layer 4              & Layer 5              & Layer 6              & Layer 2              & Layer 3              & Layer 4              & Layer 5              & Layer 6              \\ \cline{2-12}
			& $S_{\text{primary}}^{(l)}$   & ${97.8}_{\pm 3.7}$   & ${143.9}_{\pm 5.6}$  & ${198.9}_{\pm 8.1}$  & ${259.8}_{\pm 10.1}$ & ${322.8}_{\pm 12.7}$ & ${202.3}_{\pm 12.2}$ & ${234.4}_{\pm 13.1}$ & ${276.8}_{\pm 13.9}$ & ${345.2}_{\pm 16.6}$ & ${440.2}_{\pm 22.2}$ \\
			& $S_{1}^{(l)}$                & ${10.6}_{\pm 0.9}$   & ${9.5}_{\pm 0.8}$    & ${14.4}_{\pm 1.4}$   & ${24.9}_{\pm 1.3}$   & ${8.8}_{\pm 1.0}$    & ${10.3}_{\pm 1.4}$   & ${11.2}_{\pm 1.6}$   & ${12.2}_{\pm 1.3}$   & ${11.9}_{\pm 1.1}$   & ${13.2}_{\pm 1.6}$   \\
			& $S_{2}^{(l)}$                & ${7.5}_{\pm 0.9}$    & ${7.9}_{\pm 1.2}$    & ${9.7}_{\pm 1.2}$    & ${9.2}_{\pm 1.2}$    & ${8.3}_{\pm 0.6}$    & ${10.4}_{\pm 1.1}$   & ${11.6}_{\pm 1.0}$   & ${13.8}_{\pm 1.3}$   & ${10.0}_{\pm 0.8}$   & ${13.6}_{\pm 1.2}$   \\
			& $S_{3}^{(l)}$                & ${7.1}_{\pm 0.8}$    & ${9.1}_{\pm 1.1}$    & ${11.3}_{\pm 1.0}$   & ${17.9}_{\pm 2.2}$   & ${16.6}_{\pm 1.5}$   & ${11.6}_{\pm 1.4}$   & ${15.7}_{\pm 1.4}$   & ${10.7}_{\pm 1.1}$   & ${10.8}_{\pm 1.2}$   & ${19.8}_{\pm 1.6}$   \\ \hline
		\end{tabular}
	}
	\label{table:norm_cifar}
	\vspace{-10pt}
\end{table*}

Given weight changes {\small $\Delta W_{t}^{(l)}$} made by a sample $x$, the primary term {\small $\Delta W_{\text{primary},t}^{(l)}$} represents the component of weight changes along the common direction {\small $C^{(l+1)}$}. The $k$-th noise term {\small $\Delta W_{\textrm{noise}, t}^{(l,k)}$} represents the component along the $k$-th direction {\small $\mathcal{U}_k$}, which is orthogonal to {\small $C^{(l+1)}$}.

\textit{$\bullet$ Experimental verification of the significant strength of the component along the common direction {\small $C^{(l+1)}$}.}
To this end, we computed the average strength of the component along the common direction {\small $C^{(l+1)}$} over all samples in {\small $X$} as {\small $S_{\text{primary}}^{(l)}=\mathbb{E}_{t\in [T_\text{start}, T_\text{end}]}\mathbb{E}_{x \in X}\! [\|\Delta W_{\text{primary},t}^{(l)}|_x\|_F ]$}. Similarly, the strength of the component along the $k$-th noise direction was computed as {\small $S_{k}^{(l)}\!\!\!\!\!=\mathbb{E}_{t\in [T_\text{start}, T_\text{end}]}\mathbb{E}_{x \in X} [\|\Delta W_{\textrm{noise}, t}^{(l,k)}|_x\|_F ]$}. Table \ref{table:norm_cifar} illustrates that the strength of the primary component {\small $S_{\text{primary}}^{(l)}$} was more than ten times greater than the strength of components along other noise directions {\small $S_{1}^{(l)}, S_{2}^{(l)}$}, and {\small$ S_{3}^{(l)}$}.

\textit{Discussion about comparing with the sum of all other directions' significance.} According to Table \ref{table:norm_cifar}, it seems that the sum of strengths of components along other directions is also large. However, different directions decomposed by the above method are orthogonal to each other. Therefore, weight changes along different directions are independent, and their strengths cannot be summed up. Thus, we can directly compare the strength of the component of weight changes along each direction to verify the significant strength of the primary direction.

\subsection{Explaining the enhancement of the significance of the common direction}
\label{enhance}

The previous subsection owes the decreasing diversity of feature gradients to the typical phenomenon that there exists a common optimization direction shared by different ``pseudo-neurons.'' In the current subsection, we explain that the common optimization direction phenomenon is very likely to be further enhanced, just like a self-enhanced system, when features {\small $F_{t}^{(l-1)}$} of different samples have been pushed a little bit towards a specific common direction (see the following background assumption). The self-enhancement of the optimization direction will explain the decreasing diversity.

According to Eq.~\eqref{eq:146} and Eq.~\eqref{eq:340},  weight changes made by the sample $x$ can be given as
\begin{small}
\begin{equation}
    \text{Perspective 1: } \Delta W_{t}^{(l)}  =C^{(l)} \Delta V_{t}^{(l)^\top}+\Delta \varepsilon_{t}^{(l)^\top}  \quad
    \text{Perspective 2: } \Delta W_{t}^{(l)} = \Gamma_t^{(l)} F_{t}^{(l-1)^{\top}}\!\!+ \kappa_t^{(l)^\top}
    \label{eq:219}
\end{equation}
\end{small}

\begin{figure}[t]
	\centering
	\includegraphics[width=0.95\linewidth]{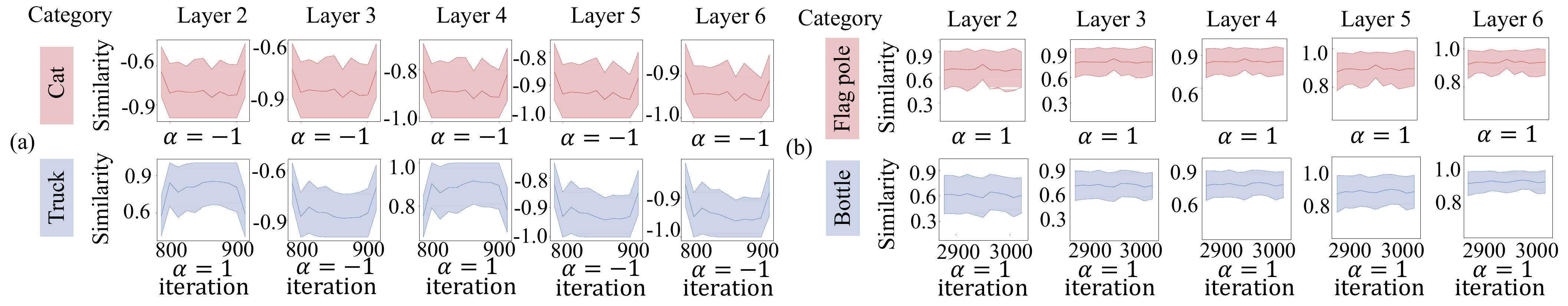}
	\vspace{-5pt}
	\caption{The average cosine similarity between the feature {\small $ F_{t}^{(l-1)}$} and the vector {\small $\Delta V_{t}^{(l)}$} over different samples in the first phase. We conducted experiments on 9-layer MLPs trained on the (a) CIFAR-10 dataset, and (b) the Tiny ImageNet dataset. The shade in each subfigure represents the standard deviation of the cosine similarity over different samples.}
	\label{fig:linear}
	\vspace{-11pt}
\end{figure}

\vspace{-4pt}
\textbf{Explaining the guess about the relationship between features and weights.} Figure \ref{fig:linear} shows a phenomenon that the feature {\small $F_{t}^{(l-1)}$} is in the similar direction of the vector {\small $\alpha \Delta V_{t}^{(l)}$}, where {\small $\alpha \in \{-1,+1\}$}. \textit{By comparing the above two perspectives of such a phenomenon, we guess that the common direction {\small $C^{(l)}$} is similar to {\small $\pm \Gamma_t^{(l)}$}, and the feature {\small $ F_{t}^{(l-1)}$} is similar to the vector {\small $\pm \Delta V_{t}^{(l)}$}.}

Therefore, in this subsection, we aim to explain that the feature {\small $F_{t}^{(l-1)}$} and the vector {\small $V_{t}^{(l)}$} become more and more similar to each other in the first phase. This explains the self-enhancement of the significance of the common direction.

\textbf{Discussions on the background assumption.} Directly proving how such a ``self-enhanced system'' emerges from the very beginning of training an initialized MLP is too difficult. Therefore, we hope to explain the significance of the common direction is probably further enhanced under the assumption that features {\small $F_{t}^{(l-1)}$} of different samples have been pushed a little bit towards a specific common direction. This assumption can derive that there exists at least one learning iteration in the first phase, in which {\small $\Delta F_t^{(l-1)}$} and {\small $ F_t^{(l-1)}$} of most samples have similar directions, and {\small $\Delta V_t^{(l)}$} and {\small $V_t^{(l)}$} have similar directions. The derivation is introduced in the supplementary material. The future self-enhancement of {\small $ F_t^{(l-1)}$}  and {\small $V_t^{(l)}$} is proved under this assumption.

\begin{lemma}
\label{lemma3}
	(Proof in the supplementary material) Given an input sample $x \in X$ and a common direction {\small $C^{(l)}$} after the {$t$}-th iteration, if the noise term {\small $\varepsilon_t^{(l)}$} is small enough to satisfy {\small $|\Delta V_{t}^{(l)^{\top}}F_{t}^{(l-1)}V_{t}^{(l)^{\top}}V_{t}^{(l)}C^{(l)^\top}C^{(l)}\Delta V_t^{(l)^\top} F_{t}^{(l-1)}|\gg|\Delta V_{t}^{(l)^{\top}}F_{t}^{(l-1)}V_{t}^{(l)^{\top}}\varepsilon_t^{(l)}\Delta \varepsilon_t^{(l)^\top}F_{t}^{(l-1)}|$}
	, we can obtain {\small $\cos(\Delta V_{t}^{(l)},F_{t}^{(l-1)}) \cdot \cos(V_{t}^{(l)}, \Delta F_{t}^{(l-1)})\geq0$}, where {\small $\Delta V_{t}^{(l)}=\frac{\Delta W_{t}^{(l)^\top} C^{(l)}}{C^{(l)^\top}C^{(l)}}$}, and {\small $V_{t}^{(l)}=\frac{W_{t}^{(l)^\top} C^{(l)}}{C^{(l)^\top}C^{(l)}}$}. {\small $\Delta F_{t}^{(l-1)}$} denotes the change of features {\small $\Delta F_{t}^{(l-1)}=F_{t+1}^{(l-1)}-F_{t}^{(l-1)}$} made by the training sample $x$ after the {\small $t$}-th iteration. To this end, we approximately consider the change of features {\small $\Delta F_{t}^{(l-1)}$} after the {\small $t$}-th iteration negatively parallel to feature gradients {\small $\dot{F}_{t}^{(l-1)}$}, although strictly speaking, the change of features is not exactly equal to the gradient \emph{w.r.t.} features.
\end{lemma}

\begin{theorem}
\label{theorem2}
	(Proof in the supplementary material) Under the background assumption, for any training samples $x, x' \!\in\! X_c$ in the category $c$, if {\small $[C^{(l)^\top}\!\!\!D_t^{(l)}|_x \dot F_t^{(l)}|_x]\cdot[C^{(l)^\top} D_t^{(l)}|_{x'} \dot F_t^{(l)}|_{x'}]>0$} (means that {\small $F_t^{(l)}|_{x}$} and {\small $ F_t^{(l)}|_{x'}$} have kinds of similarity in very early iterations), then {\small $\cos(\alpha_c \Delta V_{t}^{(l)}|_x, F_{t}^{(l-1)}|_{x})\!\geq\!0$}, and {\small $\cos(\alpha_c V_{t}^{(l)}\!\!\!, \Delta F_{t}^{(l-1)}|_{x})$ $\geq0$}, where {\small $\alpha_c\! \in\!\! \{-1,+1\}$} is a constant shared by all samples in category $c$.
\end{theorem}

\begin{figure}[t]
	\centering
	\includegraphics[width=0.95\linewidth]{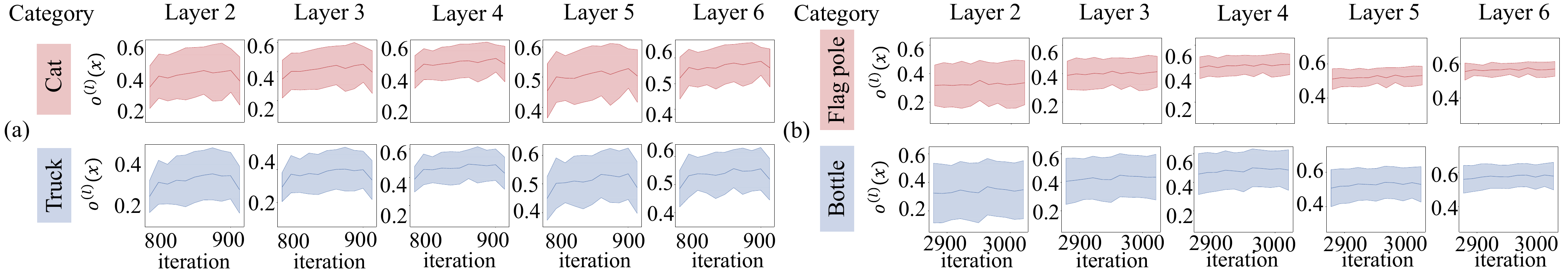}
	\vspace{-3pt}
	\caption{ The change of {\small $o^{(l)}$} in the first phase. We trained 9-layer MLPs on the (a) CIFAR-10 and the (b) Tiny ImageNet. Each layer of the MLP had 512 neurons. The supplementary material provides results on the MNIST. The shade represents the standard deviation over different samples.
	}
	\label{fig:relation}
	\vspace{-15pt}
\end{figure}

\textbf{Explaining the enhancement of the significance of the common direction caused by all training samples in a certain category.} Theorem \ref{theorem2} shows that each category has a dominating training direction.
Specifically, let us combine Theorem \ref{theorem2} and the background assumption that {\small $\Delta F_t^{(l-1)}$} and {\small $ F_t^{(l-1)}$} have similar directions, and {\small $\Delta V_t^{(l)}$} and {\small $V_t^{(l)}$} have similar directions. Thus, we can consider {\small $\cos(\alpha_c V_{t}^{(l)}, \Delta F_{t}^{(l-1)}|_{x})$ $\geq0$} in Theorem \ref{theorem2} means that features of training samples in the same category $c$ are all pushed towards a common direction {\small $\alpha_c V_{t}^{(l)}$}, making {\small $\Delta F_{t}^{(l-1)}|_{x}$} highly similar to {\small $\alpha_c V_{t}^{(l)}$}. This keeps a high similarity between features within the same category $c$. On the other hand, {\small $\cos(\alpha_c \Delta V_{t}^{(l)}|_x, F_{t}^{(l-1)}|_{x})\!\geq\!0$} in Theorem \ref{theorem2} means that training samples in the category $c$ all push {\small $V_{t}^{(l)}$} towards {\small $ \alpha_c \mathbb{E}_{x \in X_c} [F_{t}^{(l-1)}|_{x}]$}, making {\small $\Delta V_{t}^{(l)}$} roughly parallel to {\small $ \alpha_c \mathbb{E}_{x \in X_c} [F_{t}^{(l-1)}|_{x}]$}. This phenomenon is verified in Figure \ref{fig:linear}, where {\small $\cos(\alpha_c \Delta V_{t}^{(l)}, F_{t}^{(l-1)})$} is always positive over different samples of the same category. The above analysis also well explains the dynamics behind {\small $\cos(\Delta V_{t}^{(l)},F_{t}^{(l-1)}) \cdot \cos(V_{t}^{(l)}, \Delta F_{t}^{(l-1)})\geq0$} in Lemma \ref{lemma3}.

 \textbf{In sum, if we only consider training samples in the sample category $c$, then features of samples in this category would become increasingly similar to each other. On the other hand, such training samples have similar training effects, \emph{i.e.}, pushing weights of different neurons all towards the average feature.}

\textit{$\bullet$ Experimental verification of the relationship between the feature {\small $F_{t}^{(l-1)}$} and the vector {\small $V_{t}^{(l)}$}.}
To this end, we measured the change of the value {\small $o^{(l)}=\cos(\Delta\! V_{t}^{(l)}\!\!\!, F_{t}^{(l-1)}) \cdot \cos(V_{t}^{(l)}\!\!\!, \Delta F_{t}^{(l-1)})$}. Figure \ref{fig:relation} reports the average {\small $o^{(l)}$} value over different samples at each iteration. For each sample $x$, {\small $o^{(l)}$} was always positive and usually increased over iterations, which verified Lemma \ref{lemma3}. Besides, the assumption for a tiny {\small $\varepsilon_{t}^{(l)}$} in Lemma \ref{lemma3} was verified by experimental results in the supplementary material.

\begin{assumption}
\label{assp2}
\textit{We assume that the MLP encodes features of very few (a single or two) categories in the first phase, while other categories have not been learned in this phase.}
\end{assumption}

\begin{figure}[t]
    \centering
    \begin{minipage}[c]{.54\linewidth}
    \centering
        \includegraphics[width=0.99\linewidth]{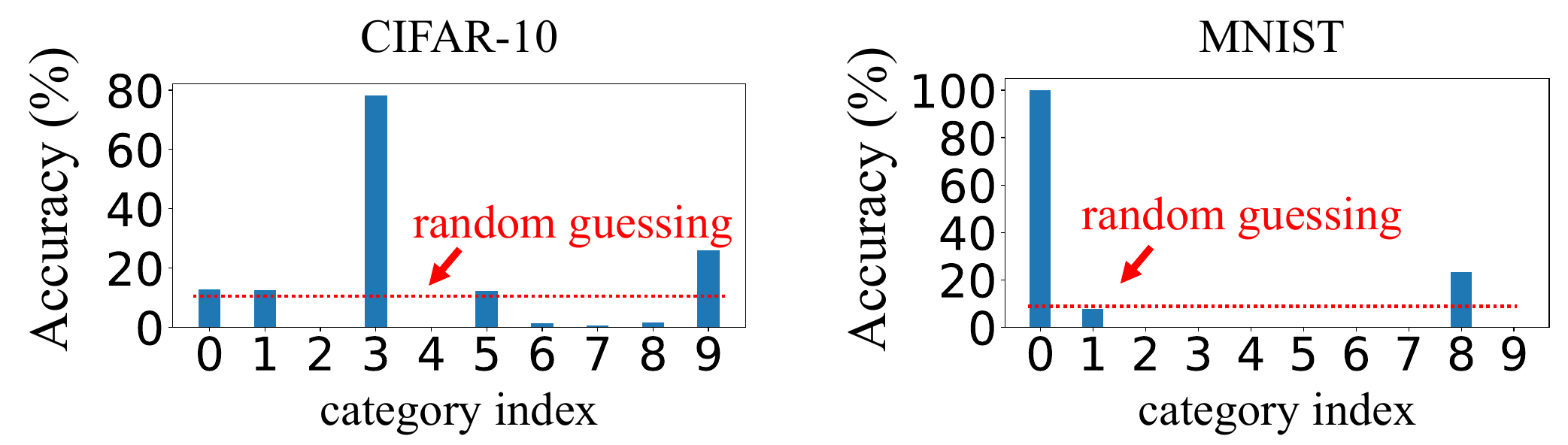}
        \vspace{-10pt}
    \end{minipage}
    \vspace{-10pt}
    \hfill
    \begin{minipage}{.45\linewidth}
        \vspace{5pt}
        \caption{The training accuracy of MLPs on the CIFAR-10 dataset and the MNIST dataset. The accuracy was evaluated at the end of the first phase. The MLP only learned features of a single or two categories in the first phase.}
        \label{fig:accuracy}
    \end{minipage}
    \vspace{-10pt}
\end{figure}

Figure \ref{fig:accuracy} verifies this assumption. It shows that only a single or two categories exhibit much higher accuracies than the random guessing at the end of the first phase. This indicates that the learning of the MLP is dominated by training samples of a single or two categories in very early iterations.

\textbf{Enhancement of the significance of the common direction caused by all training samples.} The overall learning dynamics in the first phase can be roughly described as follows, by combining Theorem \ref{theorem2} and Assumption \ref{assp2}. The overall learning effects of all training samples are dominated by very few categories {\small $\hat{c}$}. There are two effects. First, features {\small $F_t^{(l-1)}$} of different samples are all pushed towards the vector {\small $\alpha_{\hat c} V_t^{(l)}$}, where {\small $\alpha_{\hat c}$} is determined by the dominating category/categories {\small $\hat c$}. Second, {\small $V_t^{(l)}$} is pushed towards {\small $ \alpha_{\hat c} \mathbb{E}_{x\in X_{\hat c}} [F_{t}^{(l-1)}|_{x}]$}. Therefore, features {\small $F_t^{(l-1)}$} of different samples and {\small $\alpha_{\hat c} V_t^{(l)}$} enhance each other, just like a self-enhanced system, in the scenario that {\small $\Delta F_t^{(l-1)}$} and {\small $ F_t^{(l-1)}$} of most samples have similar directions, and {\small $\Delta V_t^{(l)}$} and {\small $V_t^{(l)}$} have similar directions (see the background assumption). In other words, the component along the common direction {\small $C^{(l)} \Delta V_{t}^{(l)^\top}$} in {\small  $\Delta W_{t}^{(l)}  =C^{(l)} \Delta V_{t}^{(l)^\top}+\Delta \varepsilon_{t}^{(l)^\top}$} will be further enhanced.

\textbf{Explaining the increasing feature similarity and the increasing gradient similarity.}
As aforementioned, features {\small $F_t^{(l-\!1)}$} of different samples are consistently pushed towards the same vector {\small $\alpha_{\hat c} V_t^{(l)}$}.
Therefore, the similarity between features of different samples {\small $\mathbb{E}_{x,x'\in X}\! [\cos(F_t^{(l-1)}|_x,F_t^{(l-1)}|_{x'})]$} increases in the first phase. On the other hand, the increasing similarity between feature gradients can be explained from two views. (1) The increasing feature similarity over different samples makes different training samples generate similar gating states {\small $D_t^{(l)}$} in each ReLU layer. The increasing similarity between gating states of each ReLU layer over different samples leads to the increasing similarity between feature gradients over different samples {\small $\mathbb{E}_{x,x'\in X_c}\! [\cos( \dot F_t^{(l-1)}|_x,\dot F_t^{(l-1)}|_{x'})]$} of the same category.  (2) Another view is that the component along the common direction {\small $C^{(l)} V_{t}^{(l)^\top}$} in {\small $W_{t}^{(l)}$} is enhanced in the first phase. Because {\small $C^{(l)}$} denotes the principle weight direction of each the $i$-th column {\small $w_{t,i}^{(l)}$} of {\small $W_{t}^{(l)}$}, \emph{i.e.}, \textit{each ``pseudo-neuron'' {\small $w_{t,i}^{(l)}$} is optimized towards the common direction {\small $C^{(l)}$}. Eq. \eqref{eq:232} has proved that the increasing cosine similarity between {\small $w_{t,i}^{(l)}$} and {\small $C^{(l)}$} for all ``pseudo-neurons''  will lead to the increasing similarity between feature gradients over different samples.}

\textbf{How to escape from the first phase?}
In the first phase, the MLP only discovers a single direction to optimize a single or two categories. However, the optimization of a single or two categories will soon saturate, and the gradient mainly comes from training samples of other categories, which disturbs the dominating roles of a single or two categories in the learning of the MLP. Therefore, the learning effects of training samples from different categories may conflict with each other. Thus, the self-enhanced system is destroyed, and the learning of the MLP enters the second phase.

\subsection{Theoretically alleviating the decrease of feature diversity}
\label{alleviate}

In previous sections, we have discovered and explained a fundamental yet counter-intuitive two-phase phenomenon with the MLP. This is the distinctive contribution of this study, which has not been theoretically explained for a long time. Besides, we find that we can use the above findings to explain that four typical operations can usually alleviate the decrease of feature diversity, \emph{i.e.}, normalization, momentum, initialization, and {\small $L_2$} regularization. Although these operations have been widely used, previous studies failed to theoretically explain their effectiveness. To this end, our analysis can explain a high likelihood for such operations to alleviate the decrease of feature diversity, but it is not a proof of a strict sufficient condition or a necessary condition for the feature diversity.

\textbf{Normalization.} Based on theoretical analysis, we explain that normalization operations (\emph{e.g.} batch normalization (BN)) can alleviate the decrease of feature diversity in the first phase. Specifically, according to Theorem \ref{theorem2}, the self-enhanced system of decreasing feature diversity requires features {\small $F_t^{(l)}$} of any two training samples {\small $x$} and {\small $x'$} in the same category to be similar to each other.
However, the BN layer prevents features {\small $F_t^{(l)}$} of different samples from being similar to each other, because the mean feature {\small $\bar{F}_t^{(l)}=\mathbb{E}_{x\in X}[F_t^{(l)}|_x]$} is subtracted from features of all samples, \emph{i.e.}, {\small $F_t^{\prime~(l)}|_x = F_t^{(l)}|_x-\bar{F}_t^{(l)}$}.
Therefore, features {\small $F_t^{\prime~(l)}$} of different samples are no longer similar to each other, thereby being more likely to break the self-enhanced system. Please see the supplementary material for more discussions.

\begin{figure}[t]
	\centering
	\includegraphics[width=0.96\linewidth]{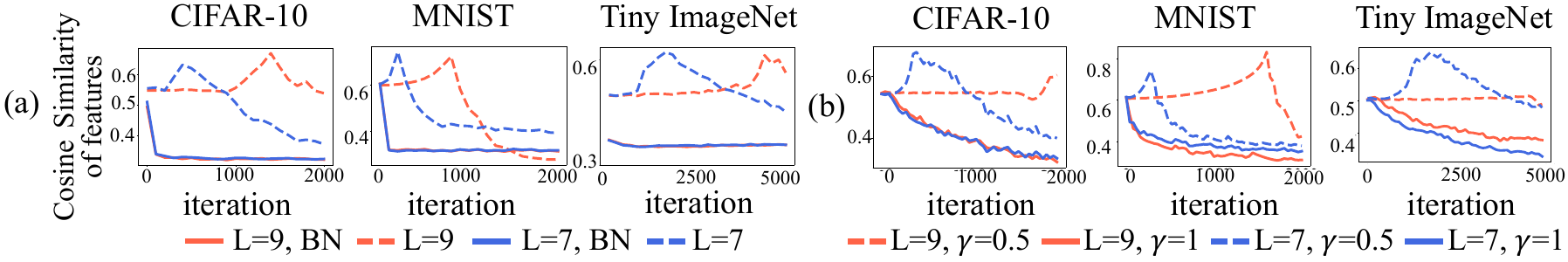}
	\vspace{-3pt}
	\caption{Effects of (a) normalization and (b) initialization. We trained $L$-layer MLPs, where each layer had 512 neurons. A shorter first phase indicates that the decrease of feature diversity is more alleviated. Effects of normalization and $L_2$ regularization are shown in the supplementary material.}
	\label{fig:effect}
	\vspace{-15pt}
\end{figure}

We conducted experiments to verify the above analysis. We compared MLPs trained with and without BN layers. Specifically, we added a BN layer after each linear layer to construct MLPs. Figure \ref{fig:effect}(a) shows that the feature similarity in MLPs with BN layers kept decreasing. This verified that BN layers alleviated the decrease of feature diversity. Besides, based on our analysis, we further simplified the BN layer, which could also alleviate the decrease of feature diversity. Please see the supplementary material for more details.

\textbf{Momentum.} Based on the theoretical explanation for the decrease of feature diversity, we can explain that momentum in gradient descent can alleviate this phenomenon. Based on Lemma \ref{lemma3}, the self-enhanced system of the decreasing of feature diversity requires weights along other directions {\small $\varepsilon_{t}^{(l)}$} to be small enough. However, because the momentum operation strengthens influences of the initialized noisy weights {\small $W_{t=0}^{(l)}$}, it strengthens singular
values of {\small $\varepsilon_{t}^{(l)}$}, to some extent, thereby alleviating the decrease of feature diversity.
Specifically, a larger coefficient of momentum {\small $m$} usually more alleviate the decrease of feature diversity.

Please see the supplementary material for the detailed discussion.
To verify the above analysis, we trained MLPs with {\small $m=0, 0.5, 0.9$}, respectively. The verification of our claim that a larger value of {\small $m$} usually more alleviates the decrease of feature diversity is introduced in the supplementary material.

\textbf{Initialization.} We explain that the initialization of MLPs also affects the decrease of feature diversity. According to Lemma \ref{lemma3}, such a self-enhanced system requires weights along other directions {\small $\varepsilon_{t}^{(l)}$} to be small enough. However, because increasing the variance of the initialized weights {\small $W_{t=0}^{(l)}$} will increase singular values of {\small $\varepsilon_{t}^{(l)}$}, alleviating the decrease of feature diversity. Please see the supplementary material for the detailed discussion.

We conducted experiments to verify the above claim by comparing MLPs trained using different initializations with different variances.
We used {\small $\gamma$} to control the variance of the initialization, \emph{i.e.}, {\small $W_{t=0}^{(l)} \sim \mathcal{N}(0,\gamma\sigma_{var}^{2})$, where {\small $\sigma_{var}$} is a constant. }
Figure \ref{fig:effect}(b) verifies that the initialization with a large variance alleviated the decrease of feature diversity.

\textbf{${L}_2$ regularization.} We also explain that a small coefficient of the ${L}_2$ regularization can alleviate the decrease of feature diversity. \emph{I.e.}, The total loss {\small  $\mathcal{L}(W_t)=\mathcal{L}^{CE}(W_t)+\lambda\|W_t\|_{2}^{2}$}, where {\small $\mathcal{L}^{CE}(W_t)$} represents the cross entropy loss and {\small $\lambda\|W_t\|_{2}^{2}$} denotes the ${L}_2$ regularization loss.
As aforementioned, the decrease of feature diversity requires singular values of {\small $\varepsilon_{t}^{(l)}$} to be small enough. However, because a smaller coefficient {\small $\lambda$} of the ${L}_2$ regularization yields larger singular values of {\small $\varepsilon_{t}^{(l)}$}, it alleviates the decrease of feature diversity.
Please see the supplementary material for the detailed discussion. The verification that a smaller coefficient {\small$\lambda$} more alleviated the decrease of feature diversity is introduced in the supplementary material.

\section{Conclusion}
In this paper, we find that in the early stage of the training process, the MLP exhibits a fundamental yet counter-intuitive two-phase phenomenon, \emph{i.e.}, the feature diversity keeps decreasing in the first phase. We explain this phenomenon by analyzing the learning dynamics of the MLP. Furthermore, we explain the reason why four typical operations can alleviate the decrease of feature diversity.


\bibliographystyle{plainnat}
\bibliography{main}

\newpage

\appendix
\setcounter{section}{0}
\setcounter{figure}{0}
\setcounter{table}{0}
\setcounter{equation}{0}
\setcounter{lemma}{0}
\setcounter{theorem}{0}

\section{Common phenomenon shared by different DNNs for different tasks.}
\label{ap:bp}

In this section, we aim to demonstrate an interesting two-phase phenomenon when we train an MLP in early iterations. Specifically, the training process of the MLP can usually be divided into the following two phases according to the training loss. In the first phase, the training loss does not decrease significantly, and the training loss suddenly begins to decrease in the second phase. More crucially, the feature diversity decreases in the first phase. This phenomenon is widely shared by different DNNs with different architectures for different tasks.

Let us take the 9-layer MLP trained on the CIFAR-10 dataset for an example, where each layer of the MLP had 512 neurons. As Figure \ref{fig:cifar}(e)(f) shows, before the 1300-th iteration (the first phase), both the feature diversity and the gradient diversity kept decreasing, \emph{i.e.}, both the cosine similarity between features over different samples and the cosine similarity between gradients kept increasing. After the 1300-th iteration (the second phase), the feature diversity and the gradient diversity suddenly began to increase, \emph{i.e.} their similarities began to decrease. Therefore, the MLP had the lowest feature diversity and the lowest gradient diversity at around the 1300-th iteration. Specifically, the training loss was evaluated on the whole training set.

\subsection{On the CIFAR-10 dataset}
\begin{figure}[h]
	\centering
	\includegraphics[width=0.98\linewidth]{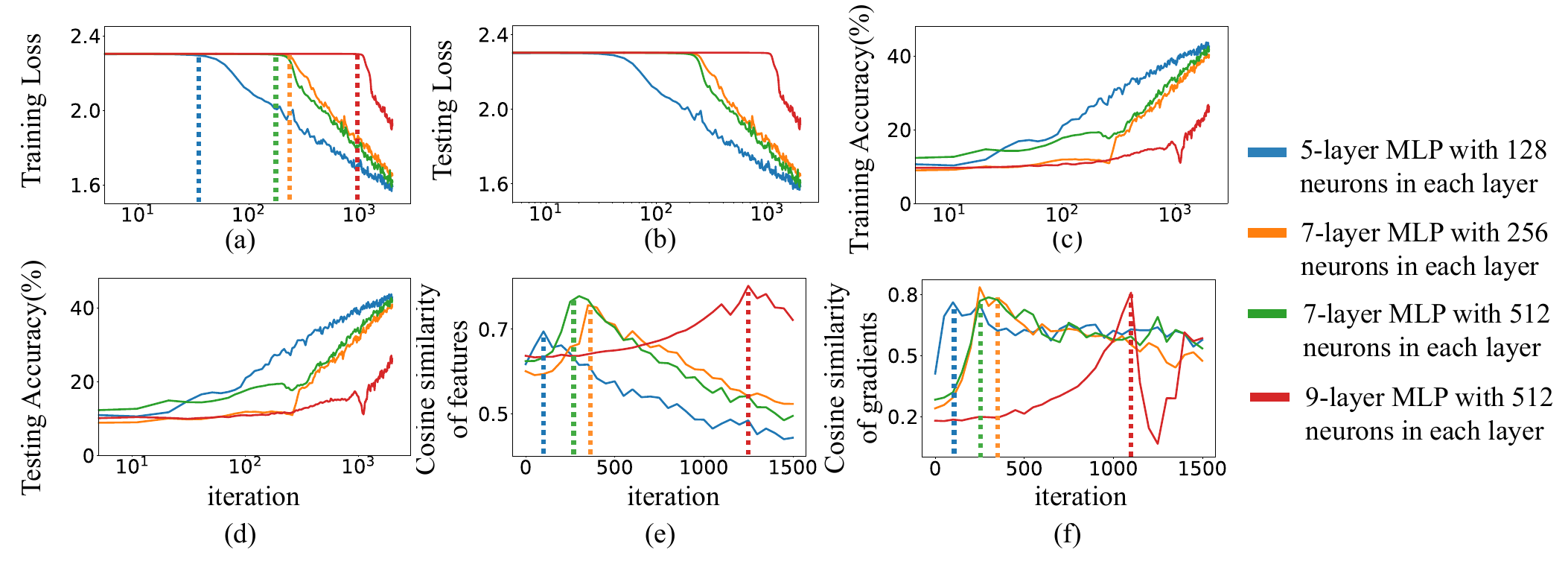}
	\vspace{-5pt}
	\caption{(a) The training loss of four MLPs trained on the CIFAR-10 dataset. (b) The testing loss of four MLPs. (c) Training accuracies of four MLPs. (d) Testing accuracies of four MLPs. (e) Cosine similarity between features of different categories. (f) Cosine similarity between gradients of different samples in a category. The feature and the feature gradient were used in the third linear layer of MLPs.}
	\label{fig:cifar}
	\vspace{-10pt}
\end{figure}

In this subsection, we demonstrated that the two-phase phenomenon was shared by different MLPs on the CIFAR-10 dataset \citep{krizhevsky2009learning}. For different MLPs, we adopted the learning rate $\eta =0.1$, the batch size $bs=100$, the SGD optimizer, and the ReLU activation function. Besides, we used two data augmentation methods, including random cropping and random horizontal flipping. The training loss, the testing loss, the training accuracy, the testing accuracy, the cosine similarity of features, and the cosine similarity of feature gradients of MLPs trained on the CIFAR-10 dataset are shown in Figure \ref{fig:cifar}.

\subsection{On the MNIST dataset}
\begin{figure}[h]
	\centering
	\includegraphics[width=0.98\linewidth]{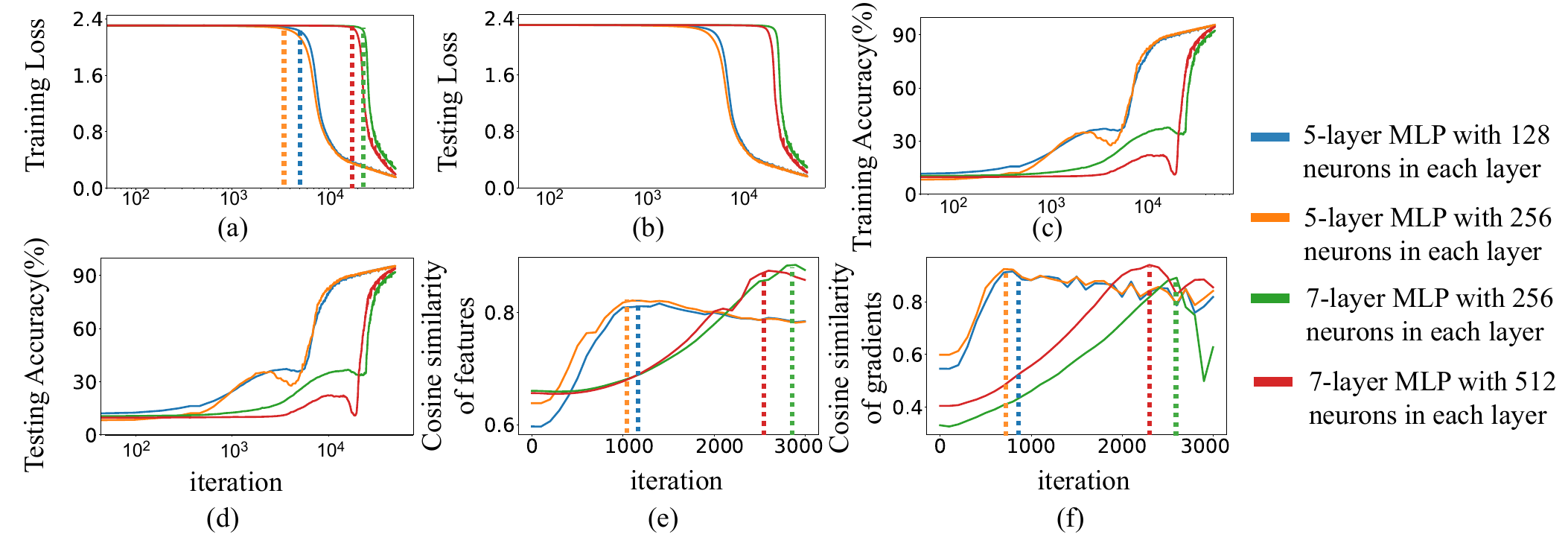}
	\vspace{-5pt}
	\caption{(a) The training loss of four MLPs tranined on the MNIST dataset. (b) The testing loss of four MLPs. (c) Training accuracies of four MLPs. (d) Testing accuracies of four MLPs. (e) Cosine similarity between features of different categories. (f) Cosine similarity between gradients of different samples in a category. The feature and the feature gradient were used in the third linear layer of MLPs.}
	\label{fig:mnist}
	\vspace{-10pt}
\end{figure}

In this subsection, we demonstrated that the two-phase phenomenon was shared by different MLPs on the MNIST dataset \citep{lecun1998gradient}. For different MLPs, we adopted the learning rate $\eta =0.01$, the batch size $bs=100$, the SGD optimizer, and the ReLU activation function. The training loss, the testing loss, the training accuracy, the testing accuracy, the cosine similarity of features, and the cosine similarity of feature gradients of MLPs trained on the MNIST are shown in Figure \ref{fig:mnist}.

\subsection{On the Tiny ImageNet dataset}
\begin{figure}[h]
	\centering
	\includegraphics[width=0.98\linewidth]{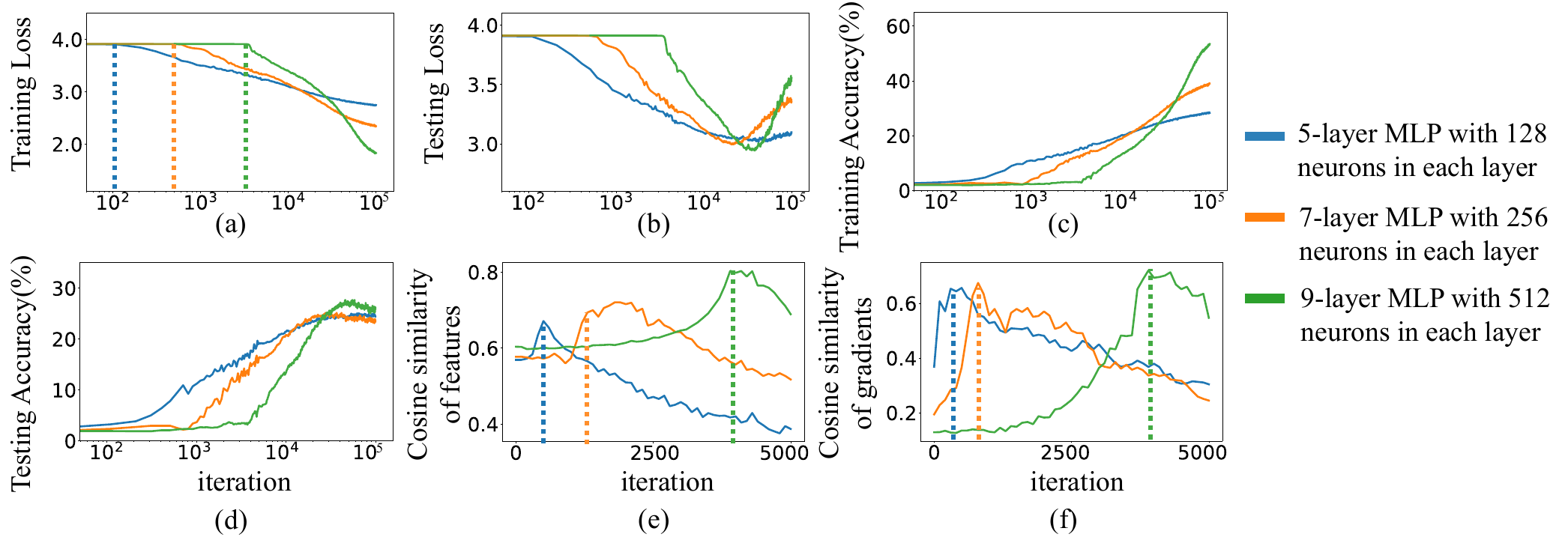}
	\vspace{-5pt}
	\caption{(a) The training loss of three MLPs tranined on the Tiny ImageNet dataset. (b) The testing loss of three MLPs. (c) Training accuracies of three MLPs. (d) Testing accuracies of three MLPs. (e) Cosine similarity between features of different categories. (f) Cosine similarity between gradients of different samples in a category. The features and the feature gradient were used in the second linear layer of MLPs.}
	\label{fig:tiny}
	\vspace{-10pt}
\end{figure}

In this subsection, we demonstrated that the two-phase phenomenon was shared by different MLPs on the Tiny ImageNet dataset \citep{le2015tiny}. Specifically, we randomly selected the following 50 categories, \textit{orangutan, parking meter, snorkel, American alligator, oboe, basketball, rocking chair, hopper, neck brace, candy store, broom, seashore, sewing machine, sunglasses, panda, pretzel, pig, volleyball, puma, alp, barbershop, ox, flagpole, lifeboat, teapot, walking stick, brain coral, slug, abacus, comic book, CD player, school bus, banister, bathtub, German shepherd, black stork, computer keyboard, tarantula, sock, Arabian camel, bee, cockroach, cannon, tractor, cardigan, suspension bridge, beer bottle, viaduct, guacamole}, and \textit{iPod} for training. For different MLPs, we adopted the learning rate $\eta =0.1$, the batch size $bs=100$, the SGD optimizer, and the ReLU activation function. Besides, we used two data augmentation methods, including random cropping and random horizontal flipping. Note that we took a random cropping with 32$\times$32 sizes.The training loss, the testing loss, the training accuracy, the testing accuracy, the cosine similarity of features, and the cosine similarity of feature gradients of MLPs trained on the Tiny ImageNet are shown in Figure \ref{fig:tiny}.

\subsection{On the Census dataset}
\begin{figure}[h]
	\centering
	\includegraphics[width=0.98\linewidth]{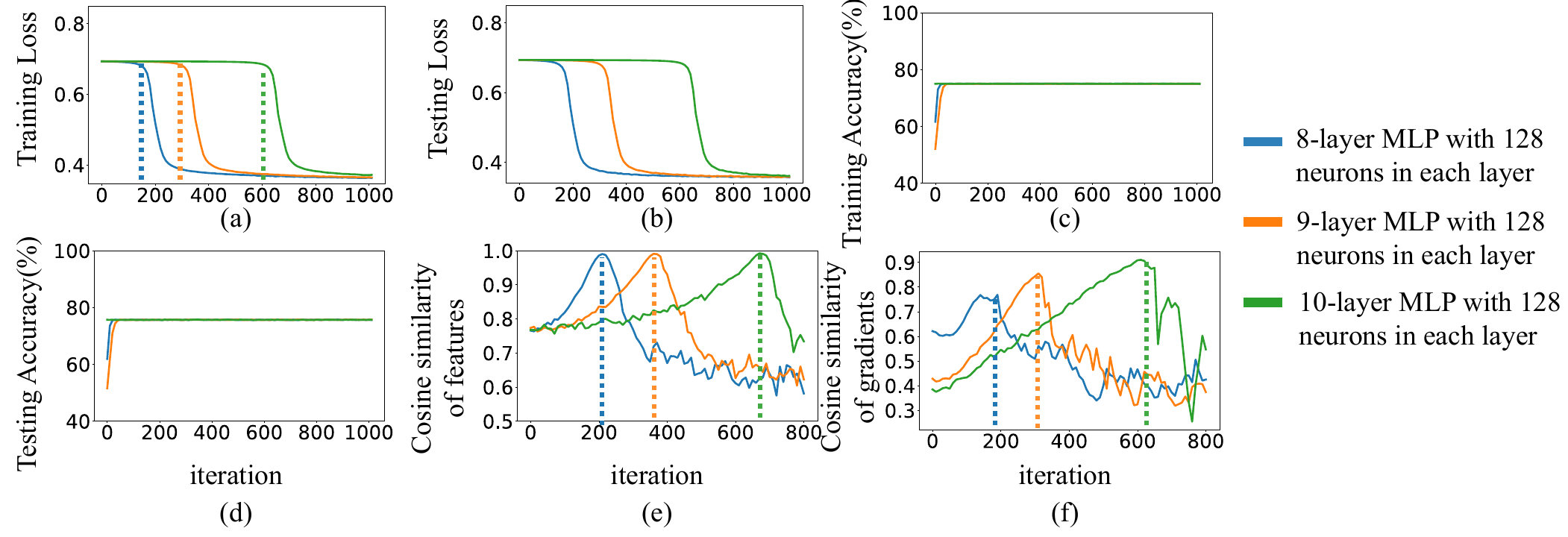}
	\vspace{-5pt}
	\caption{(a) The training loss of three MLPs trained on the Census dataset. (b) The testing loss of three MLPs. (c) Training accuracies of three MLPs. (d) Testing accuracies of three MLPs. (e) Cosine similarity between features of different categories. (f) Cosine similarity between gradients of different samples in a category. The feature and the feature gradient were used in the fifth linear layer of MLPs.}
	\label{fig:census}
	\vspace{-10pt}
\end{figure}

In this subsection, we demonstrated that the two-phase phenomenon was shared by different MLPs on the UCI census income tabular dataset (Census) \citep{asuncion2007uci}. For different MLPs, we adopted the learning rate $\eta =0.1$, the batch size $bs=1000$, the SGD optimizer, and the ReLU activation function. The training loss, the testing loss, the training accuracy, the testing accuracy, the cosine similarity of features, and the cosine similarity of feature gradients of MLPs trained on the census are shown in Figure \ref{fig:census}.

\subsection{On the Commercial dataset}
\begin{figure}[h]
	\centering
	\includegraphics[width=0.98\linewidth]{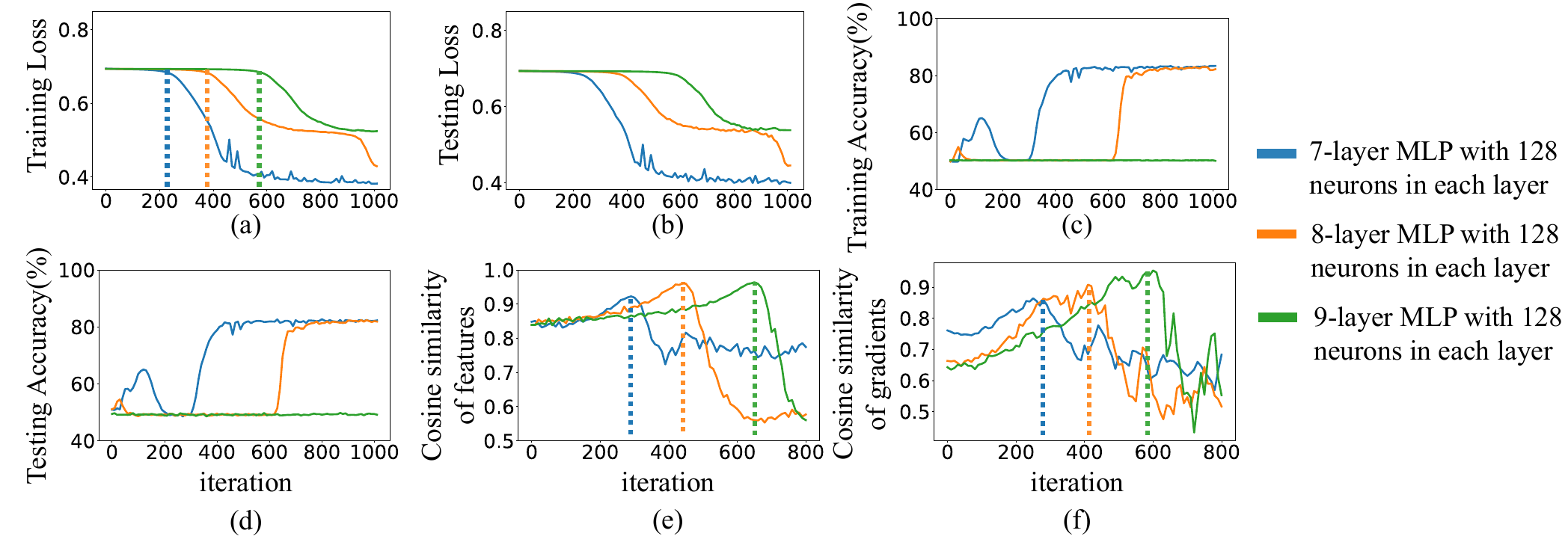}
	\vspace{-5pt}
	\caption{(a) The training loss of three MLPs trained on the Commercial dataset. (b) The testing loss of three MLPs. (c) Training accuracies of three MLPs. (d) Testing accuracies of three MLPs. (e) Cosine similarity between features of different categories. (f) Cosine similarity between gradients of different samples in a category. The feature and the feature gradient were used in the fifth linear layer of MLPs.}
	\label{fig:commercial}
	\vspace{-10pt}
\end{figure}

In this subsection, we demonstrated that the two-phase phenomenon was shared by different MLPs on the UCI TV news channel commercial detection dataset (Commercial) \citep{asuncion2007uci}. For different MLPs, we adopted the learning rate $\eta =0.1$, the batch size $bs=1000$, the SGD optimizer, and the ReLU activation function. The training loss, the testing loss, the training accuracy, the testing accuracy, the cosine similarity of features, and the cosine similarity of feature gradients of MLPs trained on the census are shown in Figure \ref{fig:commercial}.

\subsection{On the CoLA dataset}
\begin{figure}[h]
	\centering
	\includegraphics[width=0.98\linewidth]{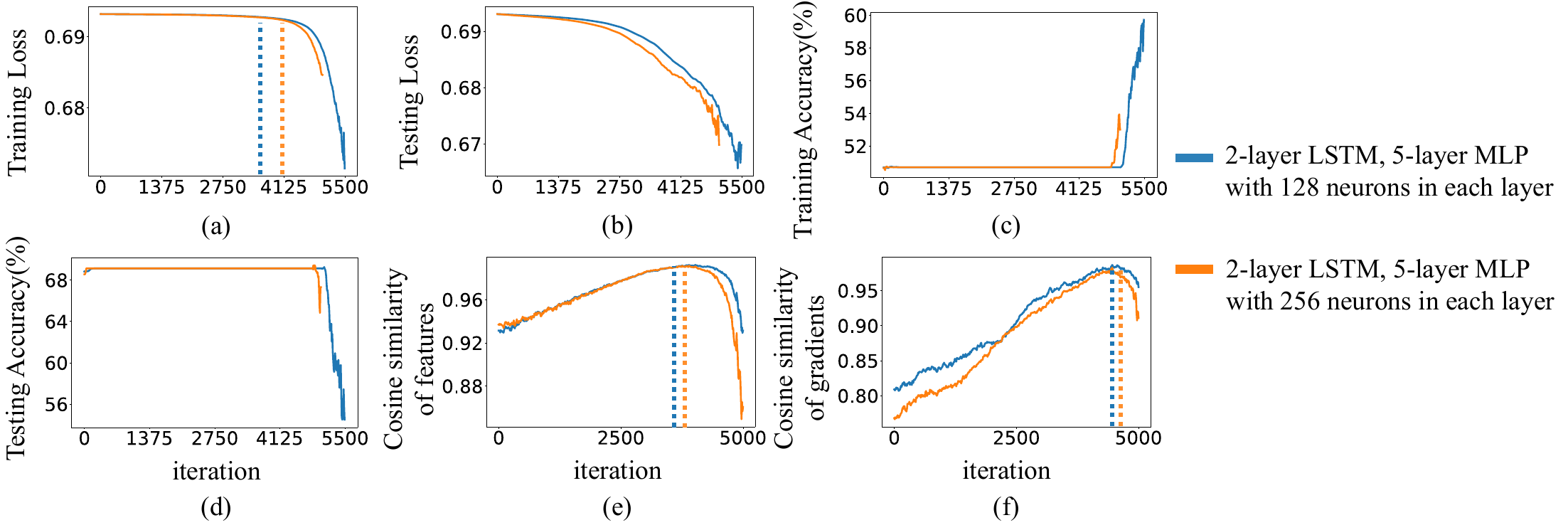}
	\vspace{-5pt}
	\caption{(a) The training loss of two LSTMs trained on the CoLA dataset. (b) The testing loss of two LSTMs. (c) Training accuracies of two LSTMs. (d) Testing accuracies of two LSTMs. (e) Cosine similarity between features of different categories. (f) Cosine similarity between gradients of different samples in a category. The feature and the feature gradient were used in the third linear layer of MLPs.}
	\label{fig:CoLA}
	\vspace{-10pt}
\end{figure}

In this subsection, we demonstrated that the two-phase phenomenon was shared by the revised LSTMs on the CoLA dataset \citep{warstadt2019neural}. We used two-layer unidirectional LSTMs concatenated with MLPs. Specifically, we trained two LSTMs with 5-layer MLPs, where each layer of the MLP has 256 and 512 neurons. We adopted the learning rate $\eta =0.1$, the batch size $bs=1000$, the SGD optimizer, and the ReLU activation function. The training loss, the testing loss, the training accuracy, the testing accuracy, the cosine similarity of features, and the cosine similarity of feature gradients of LSTMs trained on the CoLA are shown in Figure \ref{fig:CoLA}. Since training samples in the CoLA dataset were imbalanced, we constructed a new training set by randomly sampling 2000 training samples from two categories, respectively. DNNs were trained on this new training set.

\subsection{On the SST-2 dataset}
\begin{figure}[h]
	\centering
	\includegraphics[width=0.98\linewidth]{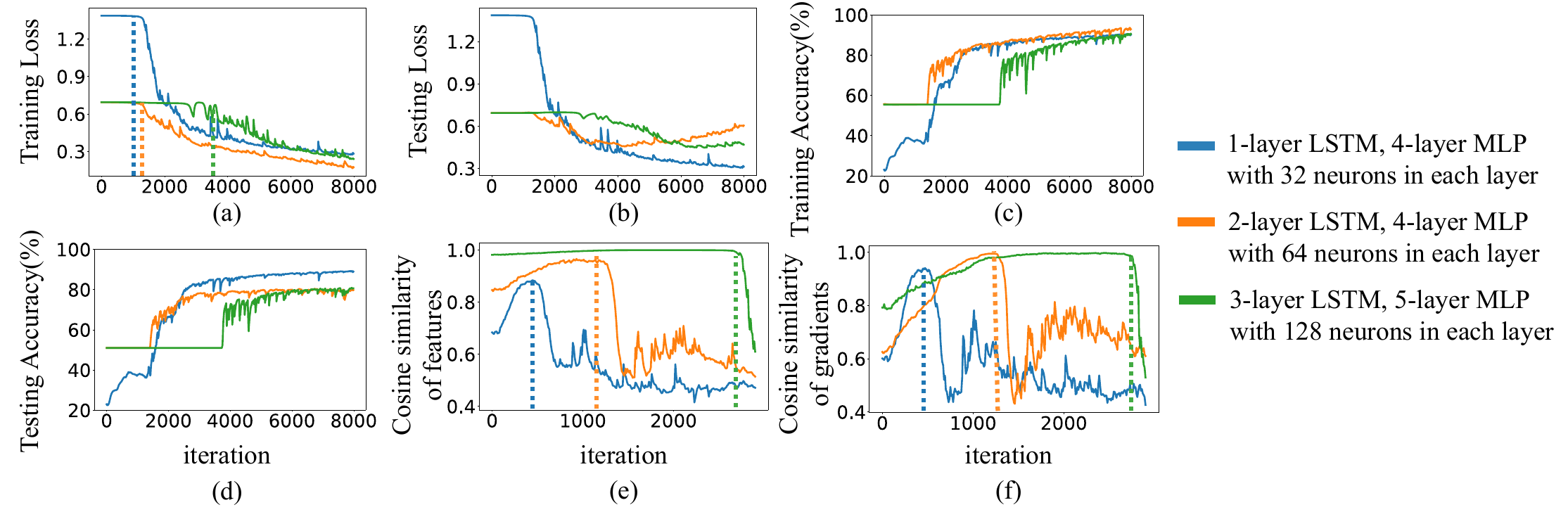}
	\vspace{-5pt}
	\caption{(a) The training loss of three LSTMs trained on the SST-2 dataset. (b) The testing loss of three LSTMs. (c) Training accuracies of three LSTMs. (d) Testing accuracies of three LSTMs. (e) Cosine similarity between features of different categories. (f) Cosine similarity between gradients of different samples in a category. The feature and the feature gradient were used in the second linear layer of MLPs.}
	\label{fig:SST-2}
	\vspace{-10pt}
\end{figure}

In this subsection, we demonstrated that the two-phase phenomenon was shared by the revised LSTMs on the SST-2 dataset \citep{socher2013recursive}. We used unidirectional LSTMs concatenated with MLPs. Specifically, we trained three LSTMs with 4-layer MLPs, 4-layer MLPs, and 5-layer MLPs, respectively, where each layer of the MLP has 32, 64, 128 neurons. We adopted the learning rate $\eta =0.1$, the batch size $bs=500$, the SGD optimizer, and the ReLU activation function. Since the training of LSTMs on the SST-2 with the SGD optimizer is unstable, we randomly selected 15000 training samples from the training set. We trained LSTMs on these 15000 training samples. The training loss, the testing loss, the training accuracy, the testing accuracy, the cosine similarity of features, and the cosine similarity of feature gradients of LSTMs trained on the SST-2 are shown in Figure \ref{fig:SST-2}.

\subsection{On the AGNews dataset}
\begin{figure}[h]
	\centering
	\includegraphics[width=0.98\linewidth]{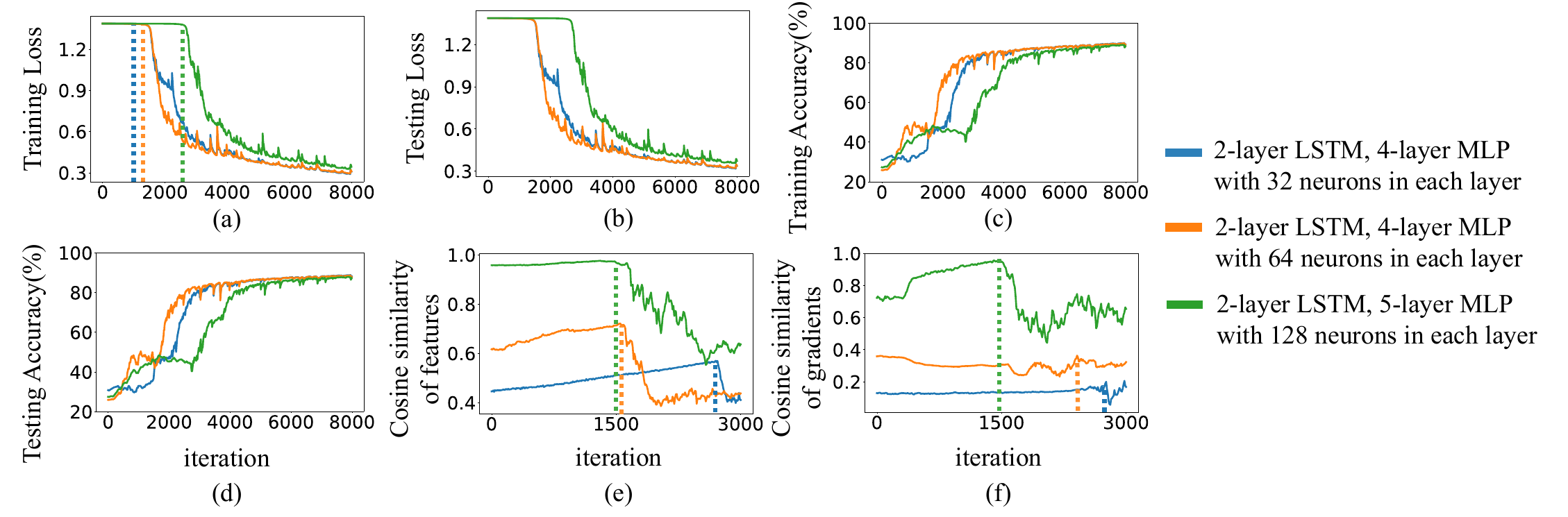}
	\vspace{-5pt}
	\caption{(a) The training loss of three LSTMs trained on the AGNEWS dataset. (b) The testing loss of three LSTMs. (c) Training accuracies of three LSTMs. (d) Testing accuracies of three LSTMs. (e) Cosine similarity between features of different categories. (f) Cosine similarity between gradients of different samples in a category. The feature and the feature gradient were used in the second linear layer of MLPs.}
	\label{fig:AGNEWS}
	\vspace{-10pt}
\end{figure}

In this subsection, we demonstrated that the two-phase phenomenon was shared by the revised LSTMs on the AGNEWS dataset. We used two-layer unidirectional LSTMs concatenated with MLPs. Specifically, we trained three LSTMs with 4-layer MLPs, 4-layer MLPs, 5-layer MLPs, respectively, where each layer of the MLP had 32, 64, and 128 neurons, respectively. We adopted the learning rate $\eta =0.1$, the batch size $bs=500$, the SGD optimizer, and the ReLU activation function. The training loss, the testing loss, the training accuracy, the testing accuracy, the cosine similarity of features, and the cosine similarity of feature gradients of LSTMs trained on the AGNEWS are shown in Figure \ref{fig:AGNEWS}.

\subsection{Different training batch sizes}
\begin{figure}[h]
	\centering
	\includegraphics[width=0.98\linewidth]{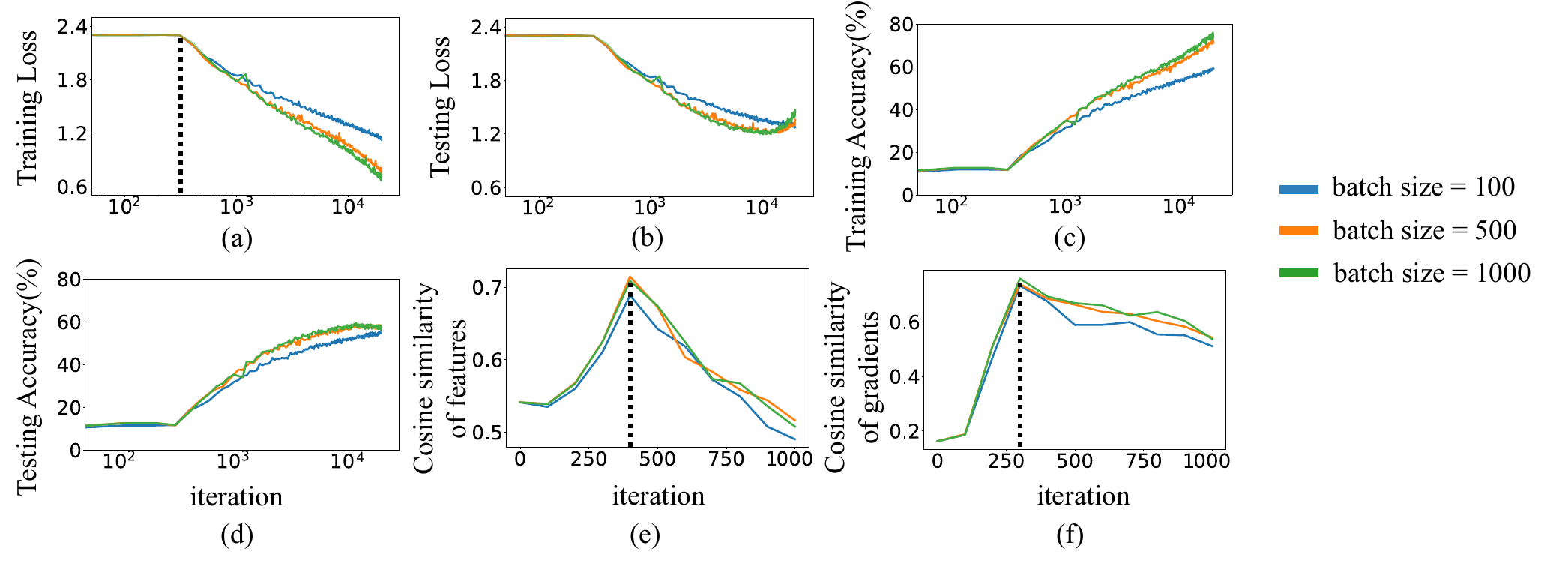}
	\vspace{-5pt}
	\caption{(a) The training loss of three MLPs trained with different batch sizes. (b) The testing loss of three MLPs. (c) Training accuracies of three MLPs. (d) Testing accuracies of three MLPs. (e) Cosine similarity between features of different categories. (f) Cosine similarity between gradients of different samples in a category. The feature and the feature gradient were used in the second linear layer of MLPs.}
	\label{fig:bs}
	\vspace{-10pt}
\end{figure}

In this subsection, we demonstrated that the two-phase phenomenon was shared by MLPs trained on the CIFAR-10 dataset with different training batch sizes. For different MLPs, we adopted the learning rate $\eta =0.1$, the SGD optimizer, and the ReLU activation function. Besides, we used two data augmentation methods, including random cropping and random horizontal flipping. We trained three 7-layer MLPs with 256 neurons in each layer, with $bs=100, 500, 1000$ respectively. The training loss, the testing loss, the training accuracy, the testing accuracy, the cosine similarity of features, and the cosine similarity of feature gradients of MLPs trained with different batch sizes are shown in Figure \ref{fig:bs}.

\subsection{Different learning rates}
\begin{figure}[h]
	\centering
	\includegraphics[width=0.98\linewidth]{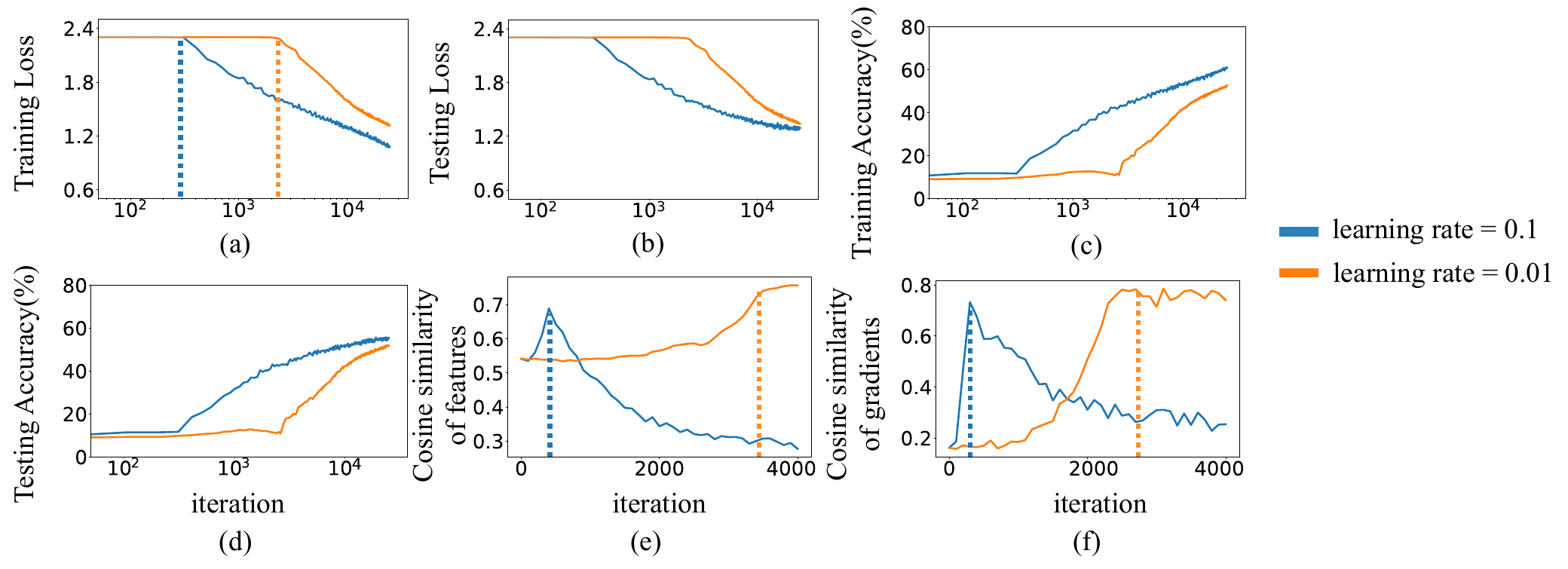}
	\vspace{-5pt}
	\caption{(a) The training loss of two MLPs trained with different learning rates. (b) The testing loss of two MLPs. (c) The training accuracies of two MLPs. (d) The testing accuracies of two MLPs. (e) Cosine similarity between features of different categories. (f) Cosine similarity between gradients of different samples in a category. The feature and the feature gradient were used in the second linear layer of MLPs.}
	\label{fig:lr}
	\vspace{-10pt}
\end{figure}

In this subsection, we demonstrated that the two-phase phenomenon was shared by MLPs trained on the CIFAR-10 dataset with different learning rates. For different MLPs, we adopted the batch size $bs =100$, the SGD optimizer, and the ReLU activation function. Besides, we used two data augmentation methods, including random cropping and random horizontal flipping. We trained two 7-layer MLPs with 256 neurons in each layer, with learning rates $\eta=0.1, 0.01$ respectively. The training loss, the testing loss, the training accuracy, the testing accuracy, the cosine similarity of features, and the cosine similarity of feature gradients of MLPs trained with different learning rates are shown in Figure \ref{fig:lr}.

\subsection{Different activation functions}
\begin{figure}[h]
	\centering
	\includegraphics[width=0.98\linewidth]{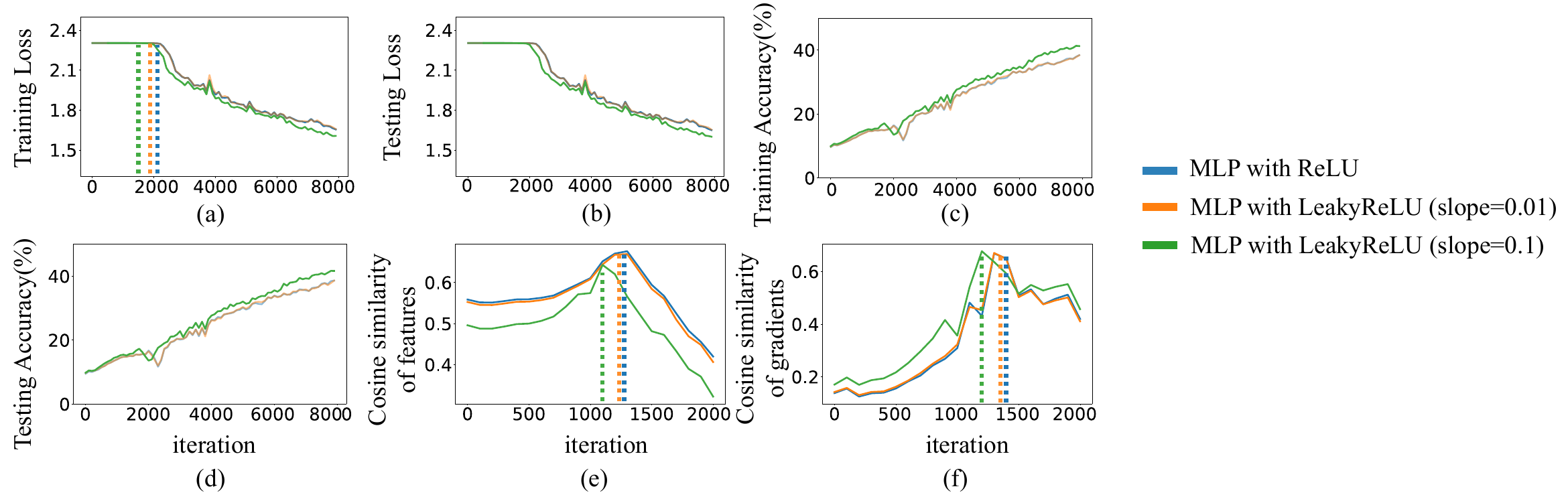}
	\vspace{-5pt}
	\caption{(a) The training loss of three MLPs with different activation functions. (b) The testing loss of three MLPs. (c) Training accuracies of three MLPs. (d) Testing accuracies of three MLPs. (e) Cosine similarity between features of different categories. (f) Cosine similarity between gradients of different samples in a category. The feature and the feature gradient were used in the second linear layer of MLPs.}
	\label{fig:relu}
	\vspace{-10pt}
\end{figure}

In this subsection, we demonstrated that the two-phase phenomenon was shared by MLPs with different activation functions. For different MLPs, we adopted the learning rate $\eta=0.1$, the batch size $bs =100$, the SGD optimizer. Besides, we used two data augmentation methods, including random cropping and random horizontal flipping. We trained three 9-layer MLPs with 512 neurons in each layer with the ReLU activation function, the Leaky ReLU (slope=0.1) activation function, and the Leaky ReLU (slope=0.01) activation function, respectively. The training loss, the testing loss, the training accuracy, the testing accuracy, the cosine similarity of features, and the cosine similarity of feature gradients of MLPs trained with different activation functions are shown in Figure \ref{fig:relu}.

\subsection{Different momentums}
\begin{figure}[t]
	\centering
	\includegraphics[width=0.98\linewidth]{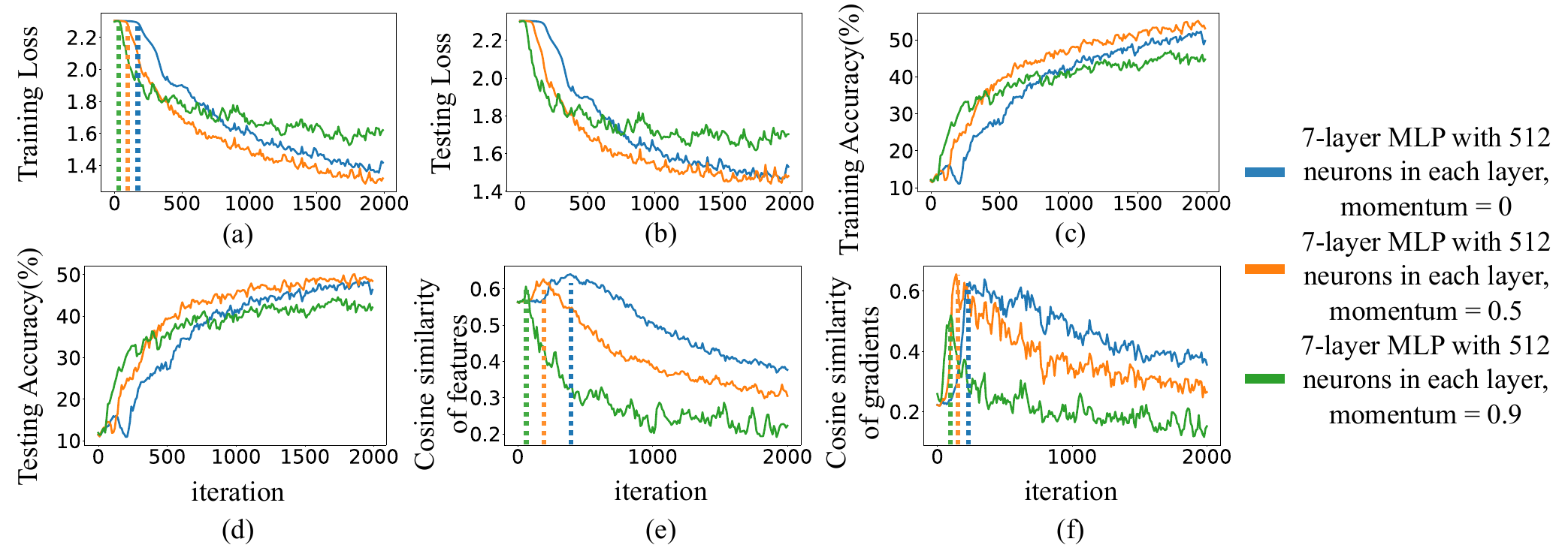}
	\vspace{-5pt}
	\caption{(a) The training loss of three MLPs with different momentums trained on the CIFAR-10 dataset. (b) The testing loss of three MLPs. (c) Training accuracies of three MLPs. (d) Testing accuracies of three MLPs. (e) Cosine similarity between features of different categories. (f) Cosine similarity between gradients of different samples in a category. The feature and the feature gradient were used in the second linear layer of MLPs.}
	\label{fig:cifar-7-mom}
	\vspace{-10pt}
\end{figure}

\begin{figure}[t]
	\centering
	\includegraphics[width=0.98\linewidth]{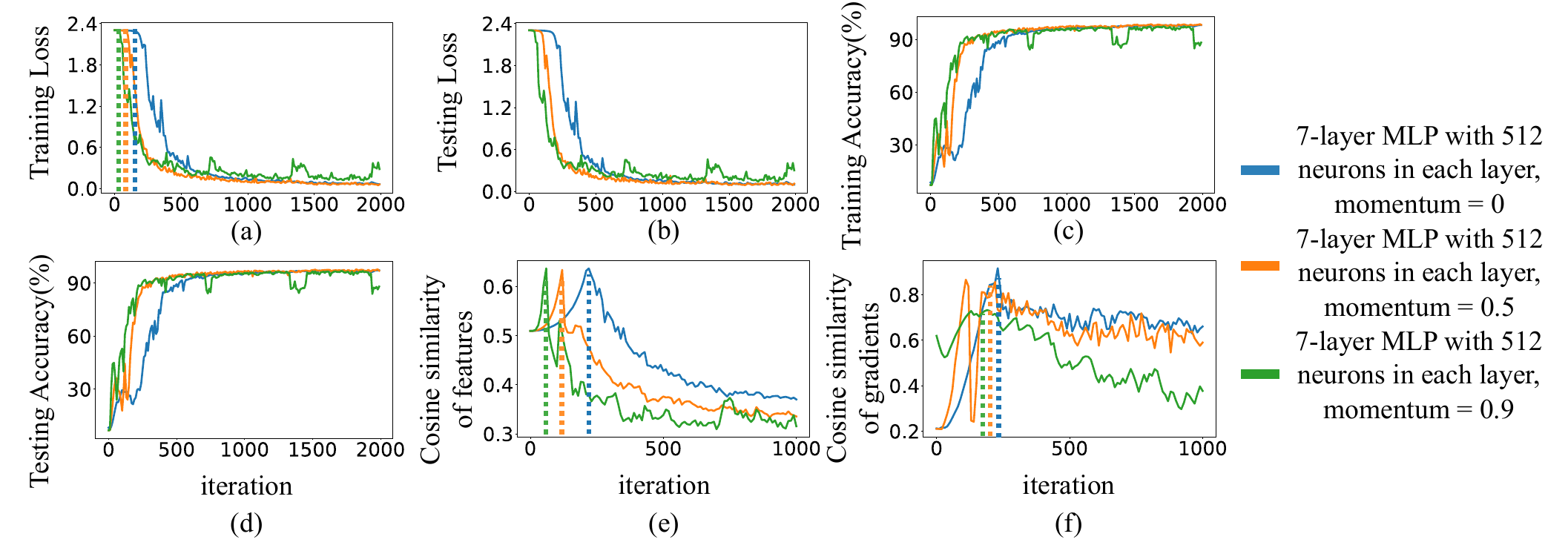}
	\vspace{-5pt}
	\caption{(a) The training loss of three MLPs with different momentums trained on the MNIST datasets. (b) The testing loss of three MLPs. (c) Training accuracies of three MLPs. (d) Testing accuracies of three MLPs. (e) Cosine similarity between features of different categories. (f) Cosine similarity between gradients of different samples in a category. The feature and the feature gradient were used in the second linear layer of MLPs.}
	\label{fig:mnist-7-mom}
	\vspace{-15pt}
\end{figure}

\begin{figure}[t]
	\centering
	\includegraphics[width=0.98\linewidth]{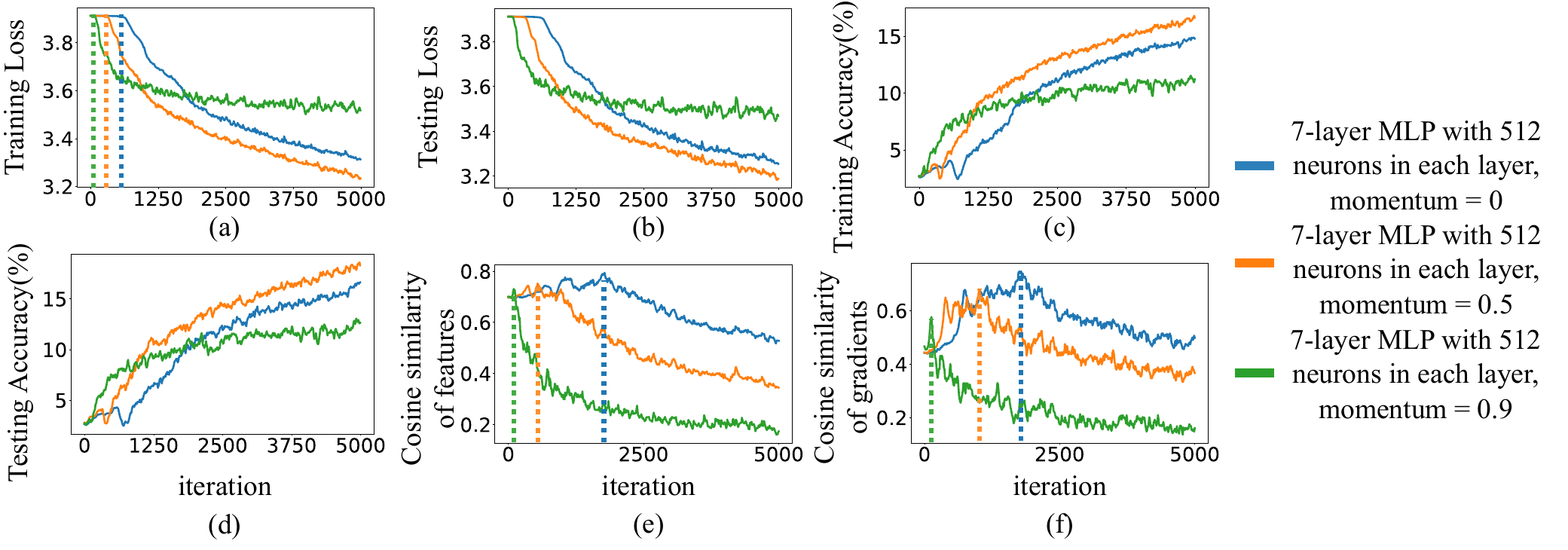}
	\vspace{-5pt}
	\caption{(a) The training loss of three MLPs with different momentums trained on the Tiny ImageNet dataset. (b) The testing loss of three MLPs. (c) Training accuracies of three MLPs. (d) Ttesting accuracies of three MLPs. (e) Cosine similarity between features of different categories. (f) Cosine similarity between gradients of different samples in a category. The feature and the feature gradient were used in the fourth linear layer of MLPs.}
	\label{fig:tiny-7-mom}
	\vspace{-10pt}
\end{figure}

\begin{figure}[t]
	\centering
	\includegraphics[width=0.98\linewidth]{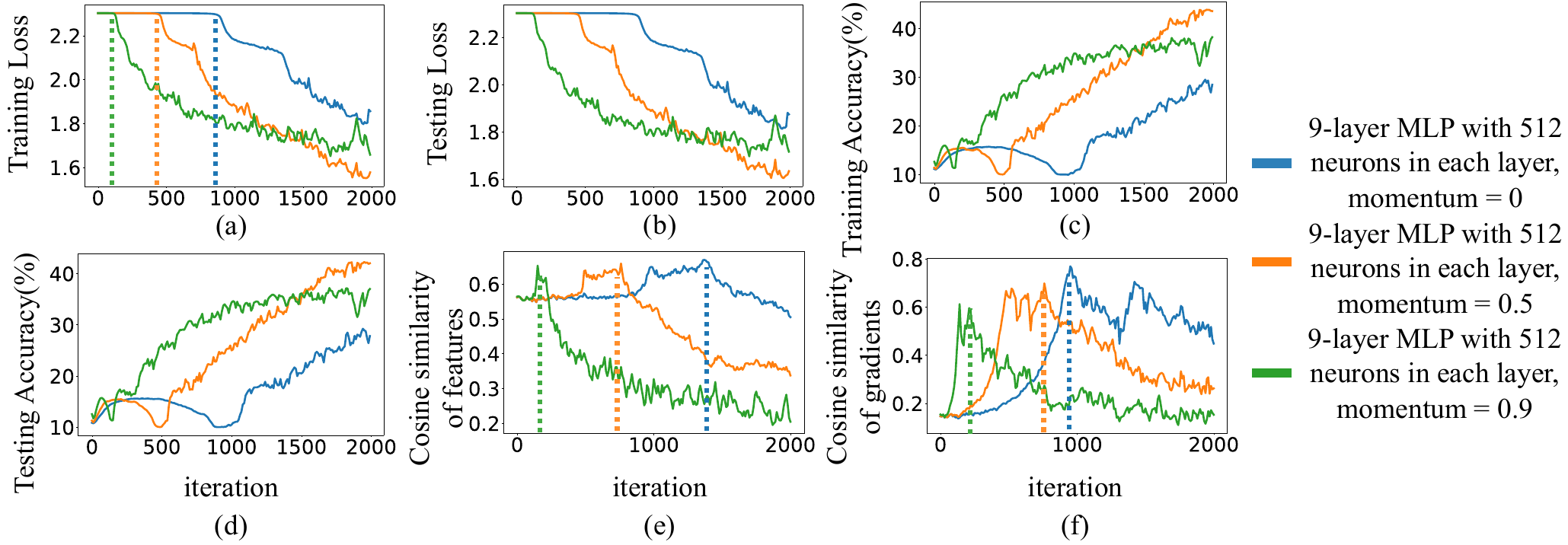}
	\vspace{-5pt}
	\caption{(a) The training loss of three MLPs with different momentums trained on the CIFAR-10 dataset. (b) The testing loss of three MLPs. (c) Training accuracies of three MLPs. (d) Testing accuracies of three MLPs. (e) Cosine similarity between features of different categories. (f) Cosine similarity between gradients of different samples in a category. The feature and the feature gradient were used in the second linear layer of MLPs.}
	\label{fig:cifar-9-mom}
	\vspace{-10pt}
\end{figure}

\begin{figure}[t]
	\centering
	\includegraphics[width=0.98\linewidth]{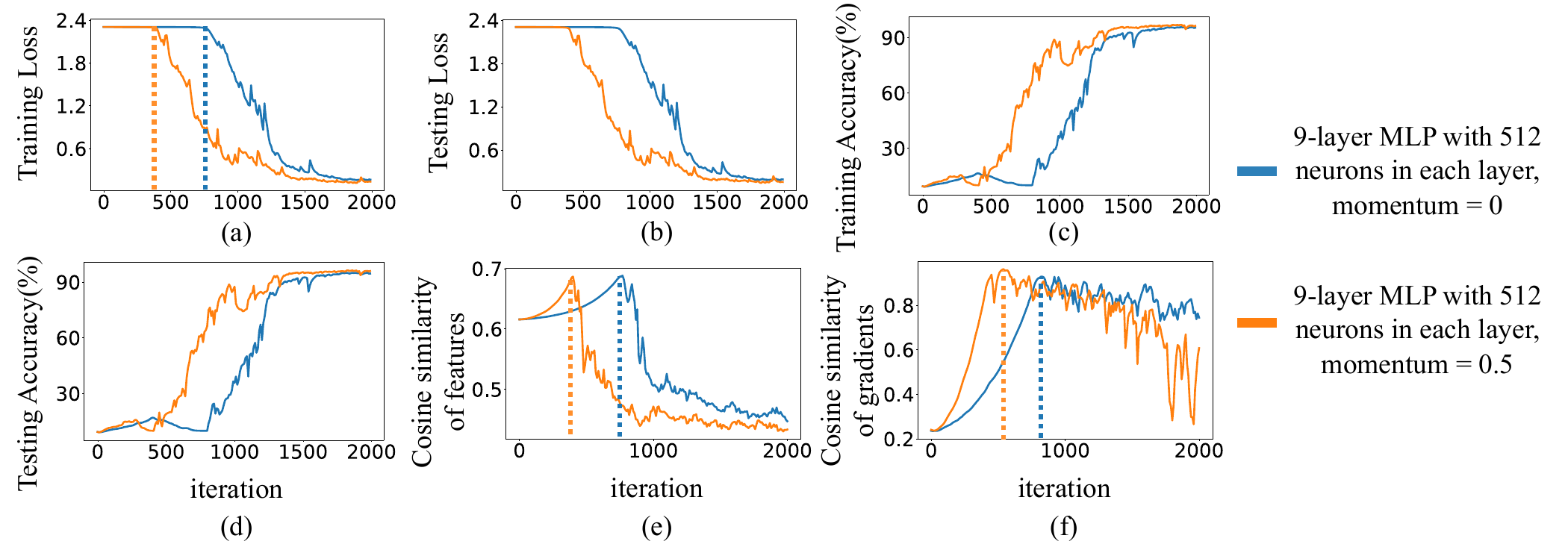}
	\vspace{-5pt}
	\caption{(a) The training loss of two MLPs with different momentums trained on the MNIST dataset. (b) The testing loss of two MLPs. (c) Training accuracies of two MLPs. (d) Testing accuracies of two MLPs. (e) Cosine similarity between features of different categories. (f) Cosine similarity between gradients of different samples in a category. The feature and the feature gradient were used in the second linear layer of MLPs.}
	\label{fig:mnist-9-mom}
	\vspace{-10pt}
\end{figure}

\begin{figure}[t]
	\centering
	\includegraphics[width=0.98\linewidth]{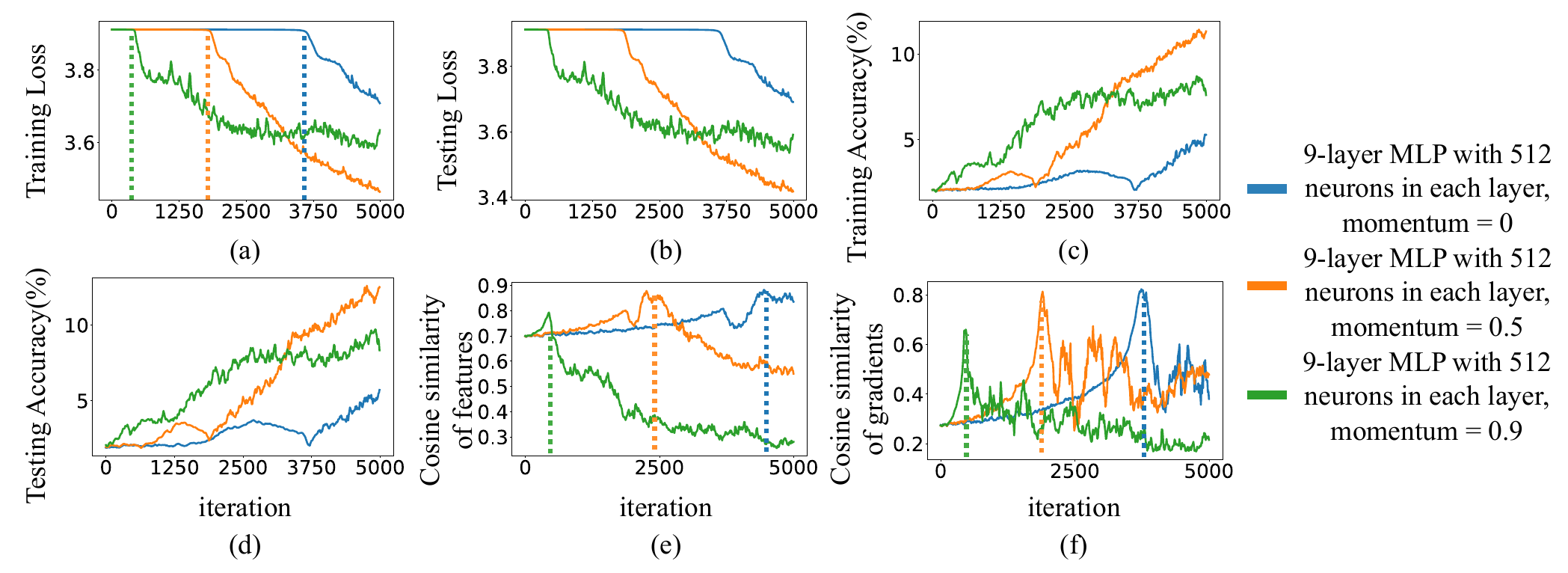}
	\vspace{-5pt}
	\caption{(a) The training loss of three MLPs with different momentums trained on the Tiny ImageNet dataset. (b) The testing loss of three MLPs. (c) Training accuracies of three MLPs. (d) Testing accuracies of three MLPs. (e) Cosine similarity between features of different categories. (f) Cosine similarity between gradients of different samples in a category. The feature and the feature gradient were used in the fourth linear layer of MLPs.}
	\label{fig:tiny-9-mom}
	\vspace{-10pt}
\end{figure}

In this subsection, we demonstrated that the two-phase phenomenon was shared by MLPs trained on the CIFAR-10, MNIST and Tiny Imagenet dataset with different momentums. For different MLPs, we adopted the learning rate $\eta=0.1$, the batch size $bs =100$, the SGD optimizer. Besides, we used two data augmentation methods, including random cropping and random horizontal flipping. We trained 7-layer MLPs and 9-layer MLPs with 512 neurons in each layer with the ReLU activation function. The training loss, the testing loss, the training accuracy, the testing accuracy, the cosine similarity of features, and the cosine similarity of feature gradients of MLPs trained with different momentum are shown in Figure \ref{fig:cifar-7-mom}, Figure \ref{fig:mnist-7-mom}, Figure \ref{fig:tiny-7-mom}, Figure \ref{fig:cifar-9-mom}, Figure \ref{fig:mnist-9-mom}, and Figure \ref{fig:tiny-9-mom}.

\clearpage

\clearpage

\subsection{Different weight decays}
\begin{figure}[h]
	\centering
	\includegraphics[width=0.98\linewidth]{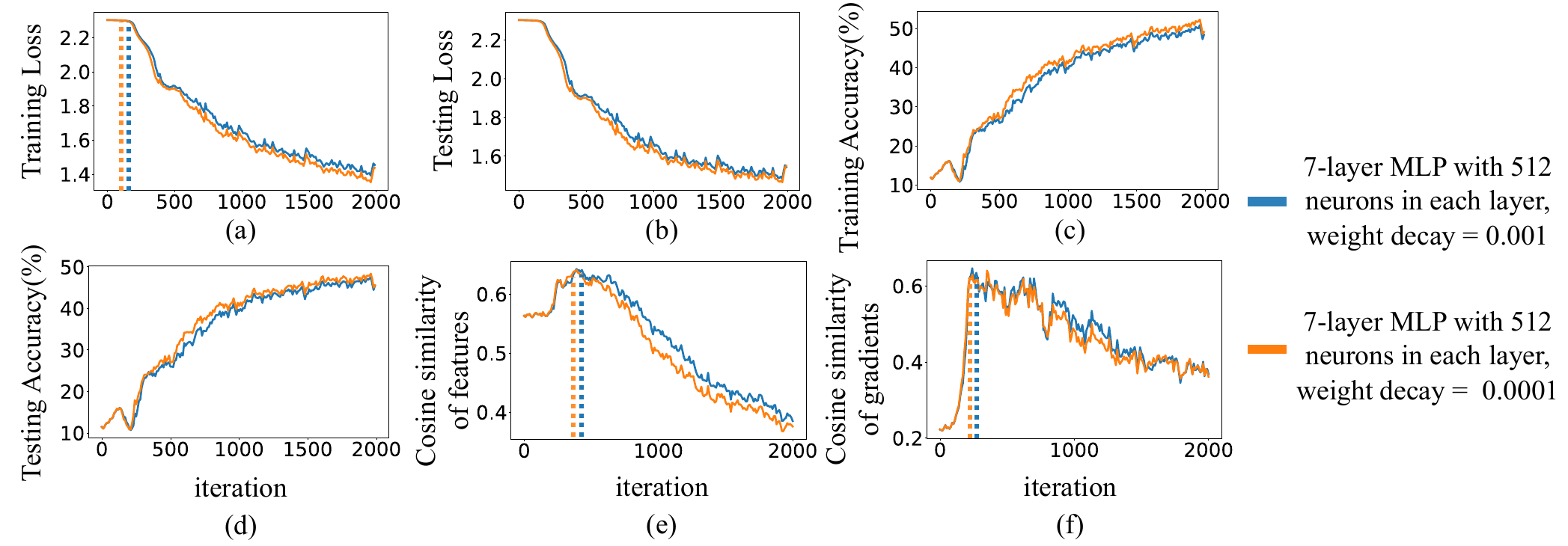}
	\vspace{-5pt}
	\caption{(a) The training loss of two MLPs with different weight decays trained on the CIFAR-10 dataset. (b) The testing loss of two MLPs. (c) Training accuracies of two MLPs. (d) Testing accuracies of two MLPs. (e) Cosine similarity between features of different categories. (f) Cosine similarity between gradients of different samples in a category. The feature and the feature gradient were used in the second linear layer of MLPs.}
	\label{fig:cifar-7-wd}
	\vspace{-10pt}
\end{figure}

\begin{figure}[h]
	\centering
	\includegraphics[width=0.98\linewidth]{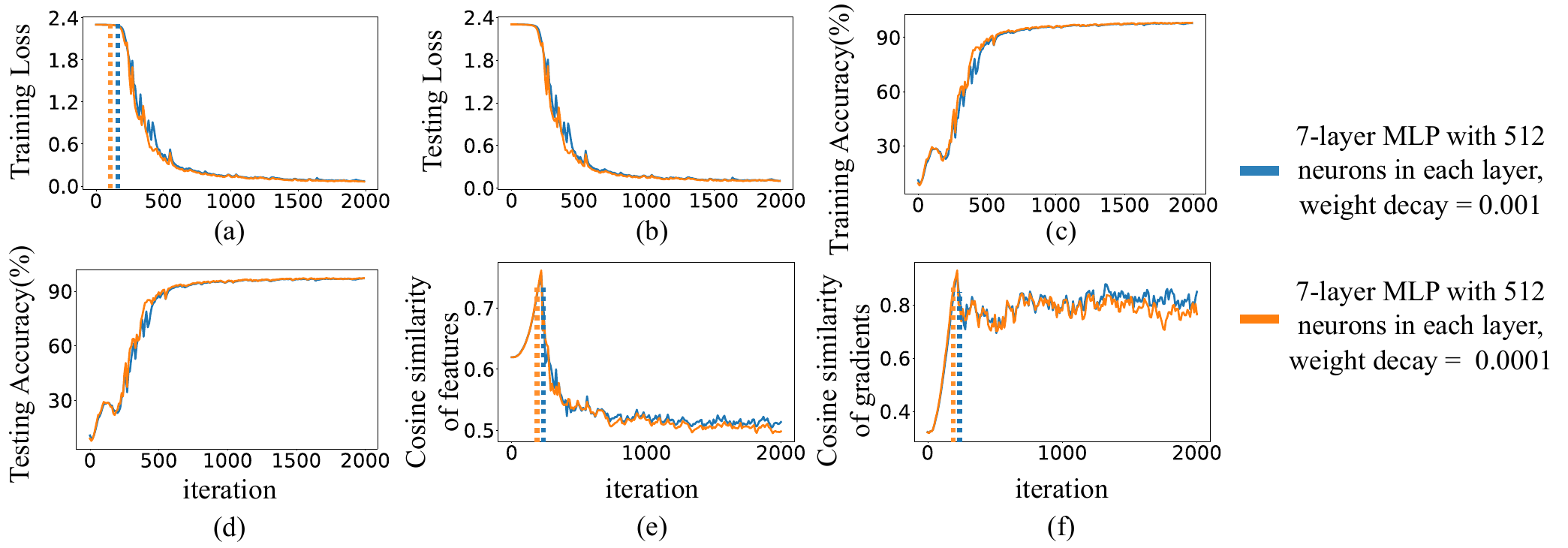}
	\vspace{-5pt}
	\caption{(a) The training loss of two MLPs with different weight decays trained on the MNIST dataset. (b) The testing loss of two MLPs. (c) Training accuracies of two MLPs. (d) Testing accuracies of two MLPs. (e) Cosine similarity between features of different categories. (f) Cosine similarity between gradients of different samples in a category. The feature and the feature gradient were used in the second linear layer of MLPs.}
	\label{fig:mnist-7-wd}
	\vspace{-10pt}
\end{figure}

\begin{figure}[t]
	\centering
	\includegraphics[width=0.98\linewidth]{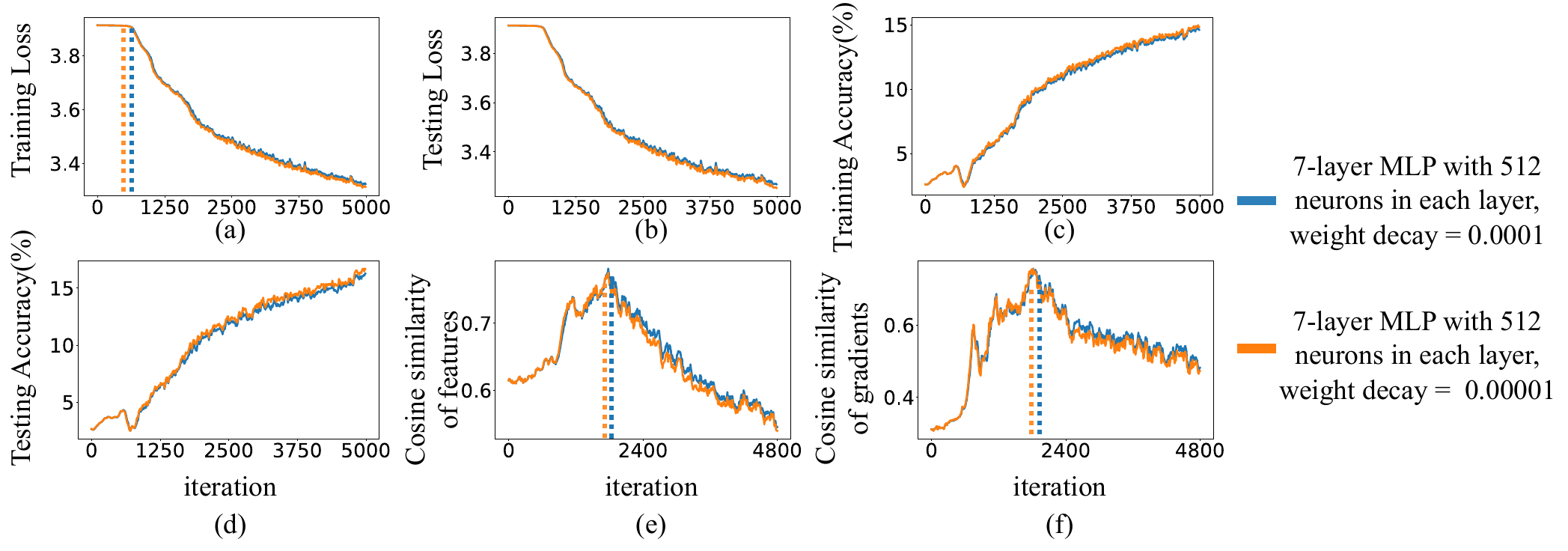}
	\vspace{-5pt}
	\caption{(a) The training loss of two MLPs with different weight decays trained on the Tiny ImageNet dataset. (b) The testing loss of two MLPs. (c) Training accuracies of two MLPs. (d) Testing accuracies of two MLPs. (e) Cosine similarity between features of different categories. (f) Cosine similarity between gradients of different samples in a category. The feature and the feature gradient were used in the third linear layer of MLPs.}
	\label{fig:tiny-7-wd}
	\vspace{-10pt}
\end{figure}

\begin{figure}[t]
	\centering
	\includegraphics[width=0.98\linewidth]{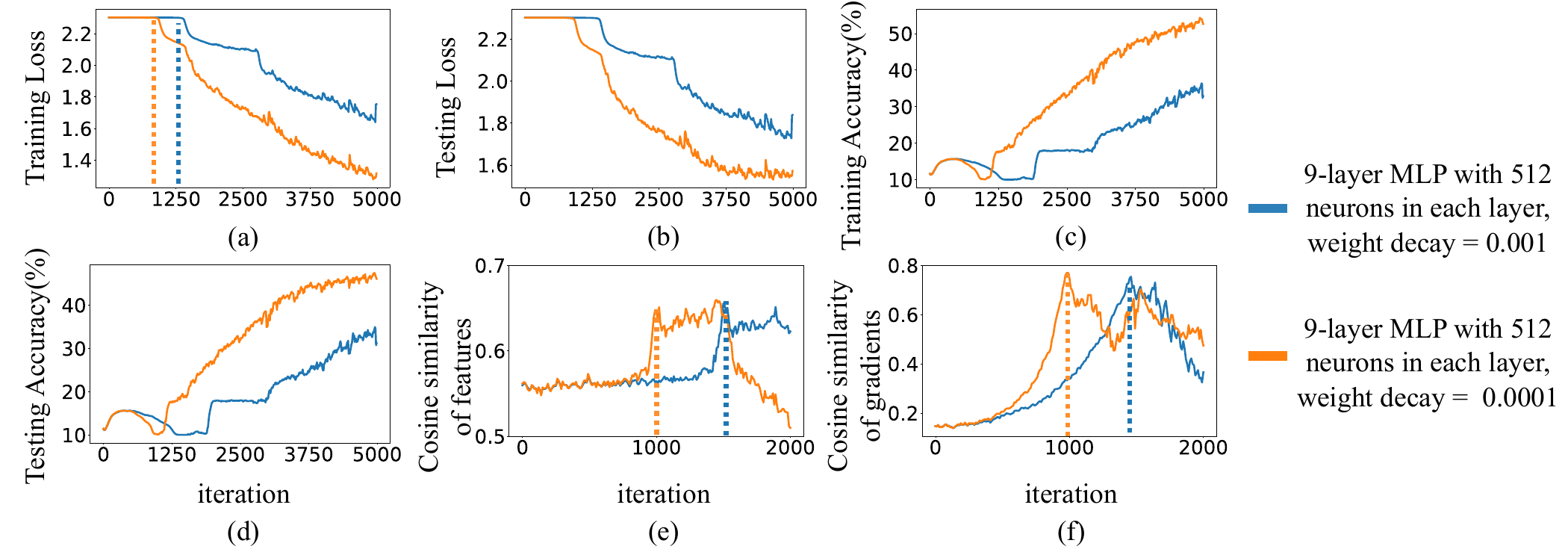}
	\vspace{-5pt}
	\caption{(a) The training loss of two MLPs with different weight decays trained on the CIFAR-10 dataset. (b) The testing loss of two MLPs. (c) Training accuracies of two MLPs. (d) Testing accuracies of two MLPs. (e) Cosine similarity between features of different categories. (f) Cosine similarity between gradients of different samples in a category. The feature and the feature gradient were used in the second linear layer of MLPs.}
	\label{fig:cifar-9-wd}
	\vspace{-10pt}
\end{figure}

\begin{figure}[t]
	\centering
	\includegraphics[width=0.98\linewidth]{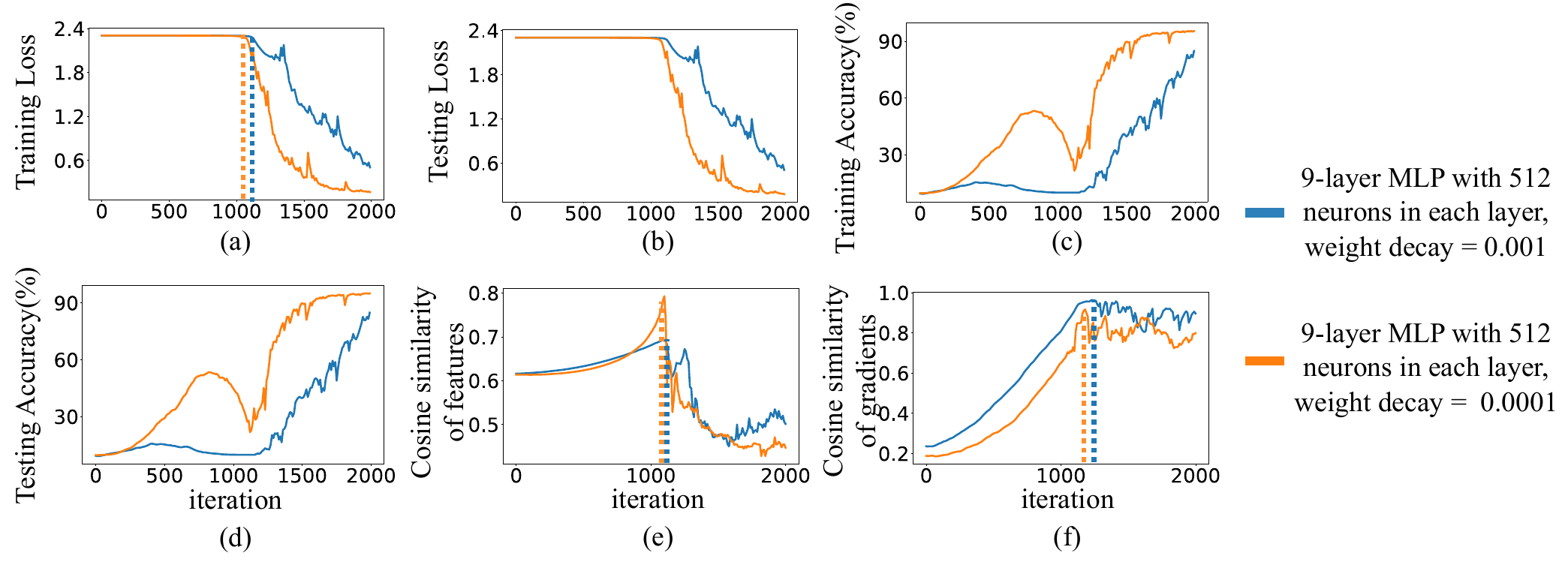}
	\vspace{-5pt}
	\caption{(a) The training loss of two MLPs with different weight decays trained on the MNIST dataset. (b) The testing loss of two MLPs. (c) Training accuracies of two MLPs. (d) Testing accuracies of two MLPs. (e) Cosine similarity between features of different categories. (f) Cosine similarity between gradients of different samples in a category. The feature and the feature gradient were used in the second linear layer of MLPs.}
	\label{fig:mnist-9-wd}
	\vspace{-10pt}
\end{figure}

\begin{figure}[h]
	\centering
	\includegraphics[width=0.98\linewidth]{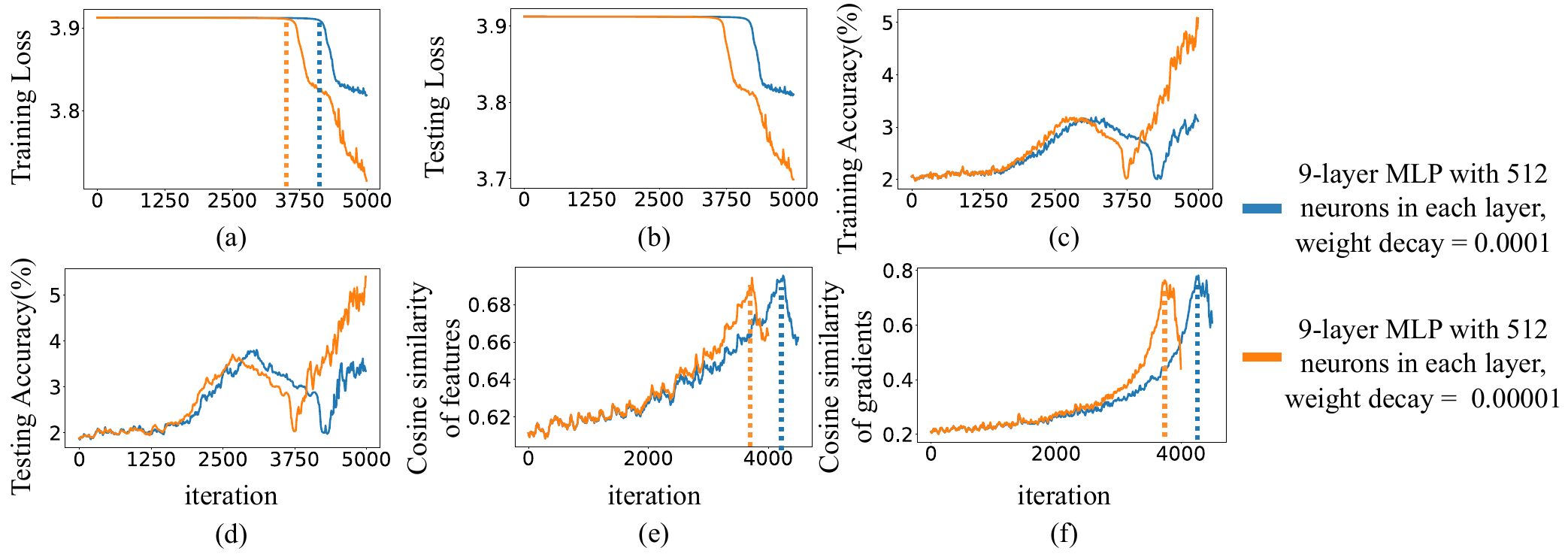}
	\vspace{-5pt}
	\caption{(a) The training loss of two MLPs with different weight decays trained on the Tiny ImageNet dataset. (b) The testing loss of two MLPs. (c) Training accuracies of two MLPs. (d) Testing accuracies of two MLPs. (e) Cosine similarity between features of different categories. (f) Cosine similarity between gradients of different samples in a category. The feature and the feature gradient were used in the third linear layer of MLPs.}
	\label{fig:tiny-9-wd}
	\vspace{-10pt}
\end{figure}

In this subsection, we demonstrated that the two-phase phenomenon was shared by MLPs trained on the CIFAR-10, MNIST and Tiny Imagenet dataset with different weight decays. For different MLPs, we adopted the learning rate $\eta=0.1$, the batch size $bs =100$, the SGD optimizer. Besides, we used two data augmentation methods, including random cropping and random horizontal flipping. We trained 7-layer MLPs and 9-layer MLPs with 512 neurons in each layer with the ReLU activation function. The training loss, the testing loss, the training accuracy, the testing accuracy, the cosine similarity of features, and the cosine similarity of feature gradients of MLPs trained with different weight decays are shown in Figure \ref{fig:cifar-7-wd}, Figure \ref{fig:mnist-7-wd}, Figure \ref{fig:tiny-7-wd}, Figure \ref{fig:cifar-9-wd}, Figure \ref{fig:mnist-9-wd}, and Figure \ref{fig:tiny-9-wd}.

\section{Discussion of the learning-sticking problem}
In this section, we aim to discuss the learning-sticking problem in the learning of MLPs. In fact, this problem appears in various DNNs, including MLPs, CNNs, and RNNs, when the task is difficult enough. In this paper, we just take the MLP as an example for discussion without loss of generality. Explaining and solving the occasional sticking of the training of MLPs are of significant values on different tasks.
Specifically, previous studies simply owed the learning-sticking problem to the difficulty of the training task and solved this problem by heuristically applying some optimization tricks.
They have no insightful analysis or theoretically supported solutions. In comparison, our study explains the learning-sticking problem as the first phase with an infinite length. Moreover, we theoretically explain mechanisms of several heuristic solutions to the learning-sticking problem.

\begin{figure}[h]
    \centering
	\includegraphics[width=0.9\linewidth]{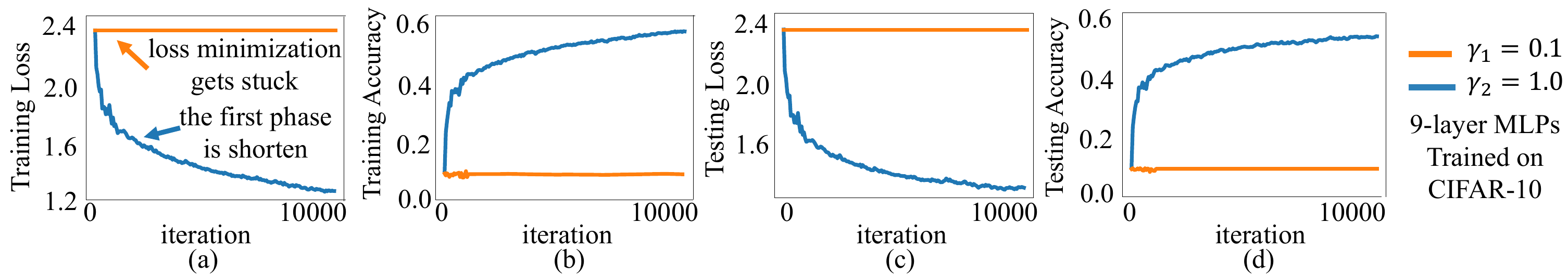}
	\vspace{-5pt}
	\caption{(a) The training loss of two MLPs trained on the CIFAR-10 dataset. When the loss minimization gets stuck (orange curve), we can consider it as the first phase with an infinite length. Therefore, the learning-sticking problem can be solved by techniques of shortening the first phase, such as the technique of increasing the variance of initial weights, which is a theoretically certificated solution in our study (blue curve). (b) The training accuracy of two MLPs. (c) The testing loss of two MLPs. (d) The testing accuracy of two MLPs.}
	\label{supp:rebuttal}
	\vspace{-5pt}
\end{figure}

To this end, the learning-sticking problem can be solved based on our study, as shown in Figure ~\ref{supp:rebuttal}. Specifically, we trained a 9-layer MLP on the CIFAR-10 dataset, where each layer of the MLP had 512 neurons and its initial weights were sample from {\small $ \mathcal{N}(0,\gamma_1\sigma_{var}^{2})$} ({\small $\gamma_1 =0.1$}). We observed that the training of the MLP got stuck (orange curve).
According to our study, the technique of increasing the variance of initial weights can shorten the first phase, thereby solving the learning-sticking problem. To this end, we trained another 9-layer MLP, and the only difference from the previous MLP is that the variance of initial weights was increased to {\small $\gamma_2\sigma_{var}^{2}$} ({\small $\gamma_2 =1$}). Figure \ref{supp:rebuttal}(a) shows that the first phase is shortened by increasing the variance of initial weights (another MLP in the blue curve), thereby the learning-sticking problem is solved.

Actually, far beyond solving the learning-sticking problem, the two-phase phenomenon of MLPs is generally considered a counter-intuitive phenomenon. In this paper, our distinctive contribution is to explain the counter-intuitive two-phase phenomenon of MLPs theoretically.

\section{Double descent}
\label{ap:dd}
There are usually two types of double-descent phenomena. The model-wise double descent behavior has emerged in many deep learning tasks, which means that as the model size increases, performance first decreases, then increases, and finally decreases \citep{advani2017high, jacot2020implicit, yang2020rethinking, d2020double}. Furthermore, some recent studies discussed the existence of the triple descent curve \citep{d2020triple, adlam2020neural}. Besides, the double descent behavior also occurs with respect to training epochs \citep{nakkiran2019deep, heckel2020early}, called epoch-wise double descent, \emph{i.e.} as the epoch increases, the testing error first decreases, then increases, and finally decreases. As Figure 1(d) in the main paper shows, the first and the second stages in the epoch-wise double descent behavior are temporally aligned with the first phase in the aforementioned two-phase phenomenon, where the training loss does not change significantly.


\section{More results on the MNIST dataset}
\begin{figure}[h]
	\centering
	\includegraphics[width=0.6\linewidth]{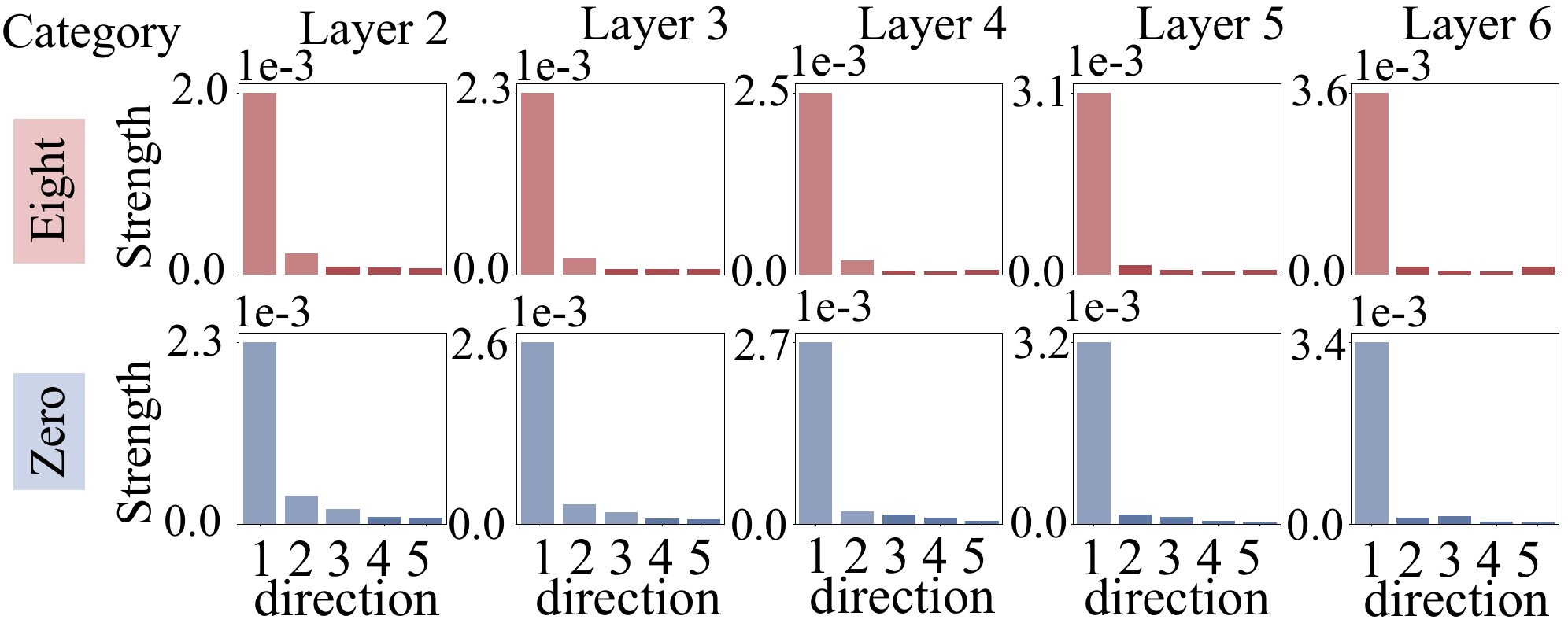}
	\vspace{-5pt}
	\caption{The strength of top-ranked common directions on the MNIST dataset. We trained a 9-layer MLP, where each layer of the MLP had 512 neurons. We computed the strength of common directions on the two categories with the highest training accuracies. {\small $s_{i} = \|C_{i}\Delta \overline V_{i}^{\top}\|_{F} $} measures the strength of weight changes along the $i$-th common direction, where {\small $\Delta \overline V_{i}=\mathbb{E}_t [\Delta \overline V_{i,t}]$}. It can be observed that the strength of the primary direction was much greater than the strength of other directions.}
	\label{fig:mnist_4}
\end{figure}

\begin{figure}[h]
	\centering
	\includegraphics[width=0.7\linewidth]{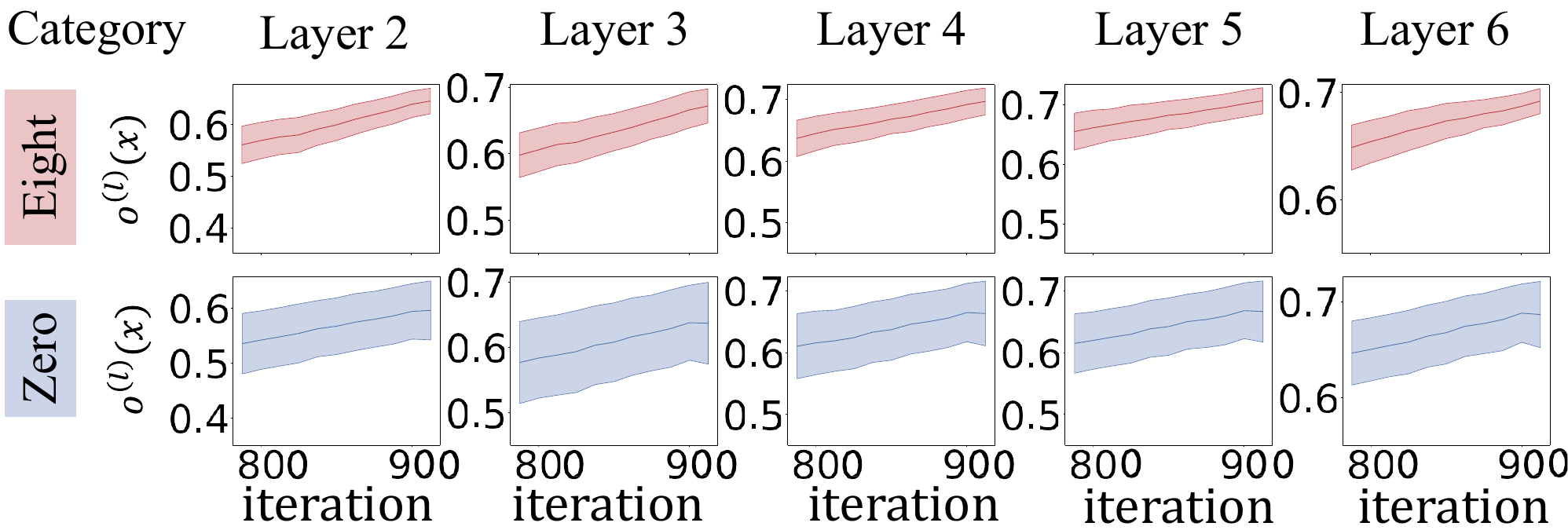}
	\vspace{-5pt}
	\caption{The change of {\small $o^{(l)}=\cos(\Delta\! V_{t}^{(l)}\!\!\!, F_{t}^{(l-1)}) \cdot \cos(V_{t}^{(l)}\!\!\!, \Delta F_{t}^{(l-1)})$} in the first phase. We trained a 9-layer MLP on the MNIST dataset. Each layer of the MLP had 512 neurons. The shade represents the standard deviation over different samples.}
	\label{fig:mnist_6}
	\vspace{-10pt}
\end{figure}

In this section, we provide more results on the MNIST dataset. Fig ~\ref{fig:mnist_4} and Tables ~\ref{table:norm} empirically verify the strength of the primary common direction, which are supplementary to Figure 4 and Table 1 in the main paper, respectively. Fig ~\ref{fig:mnist_6} illustrates the change of {\small $o^{(l)}=\cos(\Delta\! V_{t}^{(l)}\!\!\!, F_{t}^{(l-1)}) \cdot \cos(V_{t}^{(l)}\!\!\!, \Delta F_{t}^{(l-1)})$} in the first phase, which is supplementary to Figure 6 in the main paper.

\begin{table*}[h]
	\renewcommand{\arraystretch}{1}
	\caption{ Strength of components of weight changes along the primary common direction and other directions. We trained a 9-layer MLP on the MNIST dataset. Each layer of the MLP had 512 neurons. It can be observed that the strength of the primary common direction was much greater than those of other directions.}
	
	\centering
	\resizebox{0.98\linewidth}{!}{
		
		\begin{tabular}{|c|c|ccccc|ccccc|}
			\hline
			\multirow{6}{*}{\rotatebox{90}{MNIST$\quad\,\,\,$}}         & Category                     & \multicolumn{5}{c|}{Eight}                                                                                       & \multicolumn{5}{c|}{Zero}                                                                                        \\ \cline{2-12}
			& $S$ {\tiny ($\times 10^{-3}$)} & Layer 2              & Layer 3              & Layer 4              & Layer 5              & Layer 6              & Layer 2              & Layer 3              & Layer 4              & Layer 5              & Layer 6              \\ \cline{2-12}
			& $S_{\text{primary}}^{(l)}$   & ${367.1}_{\pm 56.8}$ & ${364.5}_{\pm 52.8}$ & ${381.9}_{\pm 56.3}$ & ${444.4}_{\pm 68.7}$ & ${504.0}_{\pm 81.3}$ & ${441.7}_{\pm 86.0}$ & ${448.2}_{\pm 83.5}$ & ${429.0}_{\pm 78.1}$ & ${493.1}_{\pm 87.2}$ & ${504.1}_{\pm 89.0}$ \\
			& $S_{1}^{(l)}$                & ${14.9}_{\pm 0.8}$   & ${15.9}_{\pm 1.4}$   & ${15.5}_{\pm 1.1}$   & ${15.6}_{\pm 1.5}$   & ${13.5}_{\pm 2.0}$   & ${24.6}_{\pm 3.1}$   & ${30.0}_{\pm 4.3}$   & ${18.4}_{\pm 2.6}$   & ${17.2}_{\pm 2.2}$   & ${15.6}_{\pm 1.8}$   \\
			& $S_{2}^{(l)}$                & ${16.3}_{\pm 1.7}$   & ${13.1}_{\pm 0.9}$   & ${16.4}_{\pm 0.8}$   & ${18.1}_{\pm 3.2}$   & ${11.7}_{\pm 1.6}$   & ${16.6}_{\pm 1.7}$   & ${23.9}_{\pm 4.2}$   & ${17.9}_{\pm 2.4}$   & ${14.3}_{\pm 1.5}$   & ${12.2}_{\pm 1.9}$   \\
			& $S_{3}^{(l)}$                & ${15.1}_{\pm 1.5}$   & ${16.3}_{\pm 1.7}$   & ${13.5}_{\pm 0.6}$   & ${15.1}_{\pm 1.4}$   & ${15.0}_{\pm 1.1}$   & ${29.4}_{\pm 5.2}$   & ${21.1}_{\pm 4.2}$   & ${15.5}_{\pm 1.8}$   & ${21.2}_{\pm 3.6}$   & ${14.7}_{\pm 1.6}$   \\ \hline
		\end{tabular}
	}
	\label{table:norm}
	\vspace{-5pt}
\end{table*}

\clearpage

\section{Proof for the lemma 1}
\label{ap:lemma1}

In this section, we present the detailed proof for Lemma 1.\\

\begin{lemma}
	For the decomposition { $\Delta W_{t}^{\top}\!\!=\!\!\Delta V_{t}C^\top+\Delta \varepsilon_{t}$},
	given weight changes over different samples { $\Delta W_{t}^{\top}$}, we can compute the common direction { $C$} and obtain { $\Delta V_{t}=\frac{\Delta W_{t}^{\top} C}{C^{\top}C}$} and { $\Delta\varepsilon_{t}$}= { $\Delta W_{t}^{\top}-\Delta W_{t}^{\top}\frac{C C^{\top}}{C^{\top} C} $}  \emph{s.t.} { $\Delta\varepsilon_{t}C=\bm 0$}. Such settings minimize { $\|\Delta\varepsilon_{t}\|_{F}$}.
\end{lemma}

\textit{proof}.
Let $\Delta\varepsilon_{t}^{\top}[j]$ denote the $j$-th column of the matrix $\Delta\varepsilon_{t}^{\top} \in \mathbb{R}^{h\times d}$. Given a sample $x$, we can represent $\Delta\varepsilon_{t}^{\top}[j]$ by the vector $C$ and a residual term ${\Delta\varepsilon_{t}^{\top}[j]}'$ as follows:
\begin{equation}
\Delta\varepsilon_{t}^{\top}[j]=\lambda C+{\Delta\varepsilon_{t}^{\top}[j]}',
\end{equation}

where $C^{\top}{\Delta\varepsilon_{t}^{\top}[j]}'=0$, and $\lambda$ is a scalar.
\par
Then,
\begin{equation}
\begin{split}
\left\|\Delta\varepsilon_{t}^{\top}[j]\right\|_{2}^{2}&=\left\|\lambda C+{\Delta\varepsilon_{t}^{\top}[j]}'\right\|_{2}^{2}\\
&=(\lambda C+{\Delta\varepsilon_{t}^{\top}[j]}')^{\top}(\lambda C+{\Delta\varepsilon_{t}^{\top}[j]}')\\
&=\lambda^2C^{\top}C+({{\Delta\varepsilon_{t}^{\top}[j]}'})^{\top} {\Delta\varepsilon_{t}^{\top}[j]}'\\
&=\lambda^2C^{\top}C+\left\|{\Delta\varepsilon_{t}^{\top}[j]}'\right\|_2^2
\end{split}
\end{equation}
Obviously, $\left\|\Delta\varepsilon_{t}^{\top}[j]\right\|_{2}^{2}$ is the smallest when $\lambda=0$. In other words, $\Delta\varepsilon_{t}^{\top}[j]$ does not contain the component along the direction $C$ and $C^{\top}\Delta\varepsilon_{t}^{\top}[j]=0$. Therefore, $\left\|\Delta\varepsilon_{t}^{\top}[j]\right\|_{2}^{2}$ reaches its minimum if and only if {  $\Delta\varepsilon_{t}C=\bm0$}.
\par When $\left\|\Delta\varepsilon_{t}^{\top}[j]\right\|_{2}^{2}$ reaches its minimum, $\left\|\Delta\varepsilon_{t}\right\|_{F}^{2}$ becomes the smallest. Thus, we have:
\begin{equation}
\begin{split}
\Delta W_{t}=&C\Delta V_{t}^{\top}+\Delta \varepsilon_{t}^{\top}\\
C^{\top}\Delta W_{t}=&C^{\top}C\Delta V_{t}^{\top}+C^{T}\Delta \varepsilon_{t}^{\top} \\
& = C^{\top}C\Delta V_{t}^{\top} + \bm 0
\end{split}
\end{equation}

Then, $ \Delta V_{t}^{\top}$ can be represented as follows.
\begin{equation}
\Delta V_{t}^{\top}=\frac{C^{\top}\Delta W_{t}}{C^{\top}C}
\label{132}
\end{equation}
Substituting Eq. \ref{132} into $\Delta W_{t}=C\Delta V_{t}^{\top}+\Delta \varepsilon_{t}^{\top}$, we have
\begin{equation}
\Delta \varepsilon_{t}=\Delta \mathrm{W}_{t}^{\top}-\Delta \mathrm{W}_{t}^{\top} \frac{C C^{{\top}}}{C^{\top} C}
\end{equation}

\section{Proof for the lemma 2}
\label{ap:lemma2}

In this section, we present the detailed proof for Lemma 2.\\
\begin{lemma}
	(\textbf{We can also decompose the weight { $W_{t}^{(l)}$} into the component along the common direction { $C$} and the component { $\varepsilon_{t}$} in other directions}.)
	Given the weight { $ W_{t}^{\top}$} and the common direction { $C$}, the decomposition { $ W_{t}^{\top}= V_{t}C^\top+\varepsilon_{t}$} can be conducted as { $ V_{t}=\frac{ W_{t}^{\top} C}{C^{\top}C}$} and { $\varepsilon_{t}$}= { $ W_{t}^{\top}-W_{t}^{\top}\frac{CC^{\top}}{C^{\top} C}$}  \emph{s.t.} { $\varepsilon_{t}C=\bm 0$}. Such settings minimize { $\|\varepsilon_{t}\|_{F}$}.
\end{lemma}

\textit{proof}.
Let $\varepsilon_{t}^{\top}[j]$ denote the $j$-th column of the matrix $\varepsilon_{t}^{\top} \in \mathbb{R}^{h\times d}$. We can represent $\varepsilon_{t}^{\top}[j]$ by the vector $C$ and a residual term ${\varepsilon_{t}^{\top}[j]}'$  as follows:
\begin{equation}
\varepsilon_{t}^{\top}[j]=\lambda C+{\varepsilon_{t}^{\top}[j]}',
\end{equation}

where $C^{\top}{\varepsilon_{t}^{\top}[j]}'=0$ and $\lambda$ is a scalar.
\par
Then,
\begin{equation}
\begin{split}
\left\|\varepsilon_{t}^{\top}[j]\right\|_{2}^{2}&=\left\|\lambda C+{\varepsilon_{t}^{\top}(x)[j]}'\right\|_{2}^{2}\\
&=(\lambda C+{\varepsilon_{t}^{\top}[j]}')^{\top}(\lambda C+{\varepsilon_{t}^{\top}[j]}')\\
&=\lambda^2C^{\top}C+({{\varepsilon_{t}^{\top}[j]}'})^{\top} {\varepsilon_{t}^{\top}[j]}'\\
&=\lambda^2C^{\top}C+\left\|{\varepsilon_{t}^{\top}[j]}'\right\|_2^2
\end{split}
\end{equation}
Obviously, $\left\|\varepsilon_{t}^{\top}[j]\right\|_{2}^{2}$ becomes the smallest when $\lambda=0$. In other words, $\varepsilon_{t}^{\top}[j]$ does not contain the component along the direction $C$ and $C^{\top}\varepsilon_{t}^{\top}[j]=0$. Therefore, $\left\|\varepsilon_{t}^{\top}[j]\right\|_{2}^{2}$ reaches its minimum if and only if $\varepsilon_{t}C=\bm0$.
\par When $\left\|\varepsilon_{t}^{\top}[j]\right\|_{2}^{2}$ reaches its minimum, $\left\|\varepsilon_{t}\right\|_{F}^{2}$ becomes the smallest. Thus, we have:
\begin{equation}
\begin{split}
W_{t}=&CV_{t}^{\top}+\varepsilon_{t}^{\top}\\
C^{\top} W_{t}=&C^{\top}C V_{t}^{\top}+C^{\top} \varepsilon_{t}^{\top} \\
=& C^{\top}C V_{t}^{\top}+\bm 0
\end{split}
\end{equation}

Then, $ V_{t}^{\top}$ can be written as follows.
\begin{equation}
V_{t}^{\top}=\frac{C^{\top}W_{t}}{C^{\top}C}
\label{173}
\end{equation}
Substituting Eq. \ref{173} into $W_{t}=C V_{t}^{\top}+ \varepsilon_{t}^{\top}$, we have
\begin{equation}
\varepsilon_{t}= \mathrm{W}_{t}^{\top}-\mathrm{W}_{t}^{\top}\frac{C C^{{\top}}}{C^{{\top}} C}
\end{equation}

\section{Decomposition of common directions}
\label{ap:decomposition_easy}

Actually, the estimation of the common direction $C$ is similar to the singular value decomposition (SVD), although there are slight differences.

We compute the average weight change {\small $\Delta \overline W_t=\mathbb{E}_{x \in X} \Delta W_{t}|_x$}, where {\small $\Delta W_{t}|_x$} denotes the weight change made by the sample {\small $x$}. Then, we decompose {\small $\Delta \overline W_t$} into components along five common directions as {\small $\Delta \overline W_t= C_{1}\Delta \overline V_{1,t}^{\top} + C_{2} \Delta \overline V_{2,t}^{\top} + \cdots +  C_{5} \Delta \overline V_{5,t}^{\top} + \Delta \overline \varepsilon_{5,t}^{\top}$}, where {\small $C_{1}$}={\small $C$} is termed the \textit{primary common direction}. {\small $C_{1}$}, {\small $C_{2}$}, {\small $C_{3}$}, {\small $C_{4}$}, and {\small $C_{5}$} are orthogonal to each other. {\small $C_{2}, C_{3}, C_{4}$} and {\small $C_{5}$} represent the second, third, forth, and fifth common directions, respectively. {\small $C_{i}$} represents the $i$-th common direction. {\small $\Delta \overline V_{i,t}$} denotes the average weight change along the $i$-th common direction decomposed from {\small $\Delta \overline W_t$}.

Specifically, we first decompose the average weight change {\small $\Delta \overline W_t$} after the $t$-th iteration as {\small  $\Delta \overline W_t = C\Delta \overline V_t^{\top} +\Delta \overline \varepsilon_t^{\top}$}. We remove all components along the common direction {\small $C$} from {\small $\Delta \overline W_t$}, and obtain {\small  $\Delta \overline W_{\text{\rm{new}}, t} = \Delta \overline W_{t} - C \Delta \overline V_{t}^{\top} =\Delta \overline \varepsilon_t^{\top}$}. Then, we further decompose {\small  $\Delta \overline W_{\text{\rm{new}}, t} = C_{2}\Delta V_{2,t}^{\top} +\Delta \varepsilon_{2,t}^{\top}$}. In this way, we can consider {\small $C_2$} as the secondary common direction, while {\small $C_1=C$} is termed as the primary common direction. Thus, we conduct this process recursively and obtain common directions {\small  $\{C_{1}, C_{2}, \cdots C_{5}\}$}. Accordingly, {\small  $\Delta \overline W_t$} is decomposed into {\small  $\Delta \overline W_t= C_{1}\Delta \overline V_{1,t}^{\top} + C_{2}\Delta V_{2,t}^{\top} + \cdots +  C_{5}\Delta V_{5,t}^{\top} + \Delta \varepsilon_{5,t}^{\top}$}.

\section{Decomposition of the weight change made by a sample $x$}
\subsection{Proof for Theorem 1.}
\label{ap:theorem1}
In this subsection, we present the detailed proof for Theorem 1.\\

\begin{theorem}
The weight change made by a sample  can be decomposed into {$(h+1)$} terms after the {$t$}-th iteration as follows.
	\vspace{-2pt}
	\begin{equation}
	 \small
	 \Delta W_{t}^{(l)}=  \Delta W_{\text{\rm{primary}},t}^{(l)}+ \sum\nolimits_{k=1}^h\Delta W_{\text{\rm{noise}}, t}^{(l,k)} \overset{\text{\rm{rewritten}}}{=\!\!=\!\!=\!\!=} \Gamma_t^{(l)} F_{t}^{(l-1)^{\top}}+ \kappa_t^{(l)^\top},
	\end{equation}
	
	\vspace{-6pt}
 where {\small $\Delta W_{\text{\rm{primary}},t}^{(l)}\!\!\!= D_{t}^{(l)}V_{t}^{(l+1)} $ $C^{(l+1)^{\top}} C^{(l+1)} \Delta V_{t}^{(l+1)^{\top}} F_{t}^{(l)} F_{t}^{(l-1)^{\top}}/ \|F_{t}^{(l)}\|_{2}^{2}$} denotes the component along the primary common direction,
 	and {\small $\! \Delta W_{\text{\rm{noise}}, t}^{(l,k)}\!\!=$ $D_{t}^{(l)}\varepsilon_{t}^{(l+1,k)} \Delta \varepsilon_{t}^{(l+1)^\top} F_{t}^{(l)}F_{t}^{(l-1)^{\top}}/\|F_{t}^{(l)}\|_{2}^{2}$} denotes the component along the $k$-th common direction in the noise term. {\small $\varepsilon_{t}^{(l+1, k)}\!\!\!= \boldsymbol{\Sigma}_{kk}\mathcal{U}_k\mathcal{V}_k^{\top}$}, where the SVD of {\small $ \varepsilon_{t}^{(l+1)} \in \mathbb{R}^{h\times h'} \!\!$} is given as {\small $\varepsilon_{t}^{(l+1)}=\mathcal{U}\boldsymbol{\Sigma}\mathcal{V}^{\top}$} {\small $(h \leq h')$}, and {\small $\boldsymbol{\Sigma}_{kk}$} denotes the $k$-th singular value {\small $\in \mathbb{R}$}. {\small $\varepsilon_{t}^{(l+1)}=\sum_k \varepsilon_{t}^{(l+1, k)}$}. {\small $\mathcal{U}_k$} and {\small $\mathcal{V}_k$} denote the $k$-th column of the matrix {\small $\mathcal{U}$} and {\small $\mathcal{V}$}, respectively. Besides, we have {\small $\forall k\in\{1,2,\ldots,h\},~ \mathcal{U}_k^{\top}C^{(l+1)}=0$.} Consequently, we have {\small $\Gamma_t^{(l)}\!\!\!= {D_{t}^{(l)}V_{t}^{(l+1)}C^{(l+1)^{\top}} C^{(l+1)} \Delta V_{t}^{(l+1)^{\top}} F_{t}^{(l)}} /\|F_{t}^{(l)}\|_{2}^{2} \in \mathbb{R}^{h}$}, and {\small $\kappa_t^{(l)^\top}\!\!\!=D_{t}^{(l)}$}{\small$ \varepsilon_{t}^{(l+1)}$}{\small $  \Delta \varepsilon_{t}^{(l+1)^\top}F_{t}^{(l)}F_{t}^{(l-1)^{\top}}\!\!\!\! /\|F_{t}^{(l)}\|_{2}^{2}$}{\small $  \in \mathbb{R}^{h \times d}$}.
\end{theorem}

\textit{proof.} We can represent weight matrix as $W_t^{(l)}=C^{(l)}{V_t^{(l)}}^{\top}+\varepsilon_t^{(l)^\top}$. In addition, according to the back propagation and chain rule, we have $\Delta W_t^{(l)}=-\eta D_t^{(l)} \dot F_t^{(l)}
F_t^{(l-1)^\top}$, where $\dot F_{t}^{(l)}=\frac{\partial Loss}{\partial F_{t}^{(l)}}$, and $\eta$ denotes the learning rate.

According to Lemma 1 and Lemma 2, we have $\Delta \varepsilon_{t}^{(l+1)}C^{(l+1)}=\bm0$ and $\varepsilon_{t}^{(l+1)}C^{(l+1)}=\bm0$.
After the $t$-th iteration, the weight change made by a training sample $x$ can be computed as follows.
\begin{equation}
\begin{aligned}
\Delta W_{t}^{(l)} &=-\eta D_{t}^{(l)} \dot F_{t}^{(l)} F_{t}^{(l-1)^{\top}} \\
&=-\eta D_{t}^{(l)} W_t^{(l+1)^{\top}} D_t^{(l+1)}\dot F_t^{(l+1)} F_{t}^{(l-1)^{\top}} \\
&=D_{t}^{(l)} W_{t}^{(l+1)^{\top}} \Delta W_{t}^{(l+1)} F_{t}^{(l)} F_{t}^{(l-1)^{\top}} /\left\|F_{t}^{(l)}\right\|_{2}^{2} \\
&=D_{t}^{(l)} \left [V_{t}^{(l+1)}C^{(l+1)^{\top}}+\varepsilon_{t}^{(l+1)} \right] \left [C^{(l+1)}\Delta V_{t}^{(l+1)^{\top}}+ \Delta \varepsilon_{t}^{(l+1)^{\top}} \right]  F_{t}^{(l)} F_{t}^{(l-1)^{\top}} /\left\|F_{t}^{(l)}\right\|_{2}^{2} \\
&=D_{t}^{(l)} [ V_{t}^{(l+1)}C^{(l+1)^{\top}}C^{(l+1)}\Delta V_{t}^{(l+1)^{\top}} + V_{t}^{(l+1)}C^{(l+1)^{\top}}\Delta \varepsilon_{t}^{(l+1)^{\top}} \\ &+\varepsilon_{t}^{(l+1)}C^{(l+1)}\Delta V_{t}^{(l+1)^{\top}} +\varepsilon_{t}^{(l+1)}\Delta \varepsilon_{t}^{(l+1)^{\top}} ]  F_{t}^{(l)} F_{t}^{(l-1)^{\top}} /\left\|F_{t}^{(l)}\right\|_{2}^{2} \\
&=D_{t}^{(l)} \left [ V_{t}^{(l+1)}C^{(l+1)^{\top}}C^{(l+1)}\Delta V_{t}^{(l+1)^{\top}} +\varepsilon_{t}^{(l+1)}\Delta \varepsilon_{t}^{(l+1)^{\top}} \right]  F_{t}^{(l)} F_{t}^{(l-1)^{\top}} /\left\|F_{t}^{(l)}\right\|_{2}^{2} \\
&=D_{t}^{(l)} V_{t}^{(l+1)}C^{(l+1)^{\top}}C^{(l+1)}\Delta V_{t}^{(l+1)^{\top}} F_{t}^{(l)} F_{t}^{(l-1)^{\top}} /\left\|F_{t}^{(l)}\right\|_{2}^{2} \\
&+D_{t}^{(l)} \varepsilon_{t}^{(l+1)}\Delta \varepsilon_{t}^{(l+1)^{\top}}  F_{t}^{(l)} F_{t}^{(l-1)^{\top}} /\left\|F_{t}^{(l)}\right\|_{2}^{2}\\
\end{aligned}
\end{equation}

{\small $\varepsilon_{t}^{(l+1,k)}={\Sigma}_{kk}\mathcal{U}_k\mathcal{V}_k^{\top}$}, where the singular value decomposition of {\small  $\varepsilon_{t}^{(l+1)}$} is given as {\small  $\varepsilon_{t}^{(l+1)}=\mathcal{U}\boldsymbol{\Sigma}\mathcal{V}^{\top}$}, and {\small  $\boldsymbol{\Sigma}_{kk}$} denotes the $k$-th singular value. {\small  $\mathcal{U}_k$} and { \small $\mathcal{V}_k$} denote the $k$-th column of the matrix {\small  $\mathcal{U}$} and {  \small $\mathcal{V}$}, respectively. We can derive the following equations.

\begin{equation}
\begin{aligned}
\Delta W_{t}^{(l)} &=D_{t}^{(l)} V_{t}^{(l+1)}C^{(l+1)^T}C^{(l+1)}\Delta V_{t}^{(l+1)^\top} F_{t}^{(l)} F_{t}^{(l-1)^{\top}} /\left\|F_{t}^{(l)}\right\|_{2}^{2} \\
&+D_{t}^{(l)} \varepsilon_{t}^{(l+1)}\Delta \varepsilon_{t}^{(l+1)^\top}  F_{t}^{(l)} F_{t}^{(l-1)^{\top}} /\left\|F_{t}^{(l)}\right\|_{2}^{2}\\
&=D_{t}^{(l)} V_{t}^{(l+1)}C^{(l+1)^T}C^{(l+1)}\Delta V_{t}^{(l+1)^\top} F_{t}^{(l)} F_{t}^{(l-1)^{\top}} /\left\|F_{t}^{(l)}\right\|_{2}^{2} \\
&+\sum^h_{k=1} D_{t}^{(l)}\varepsilon_{t}^{(l+1,k)} \Delta \varepsilon_{t}^{(l+1)^\top} F_{t}^{(l)} F_{t}^{(l-1)^{\top}} /\left\|F_{t}^{(l)}\right\|_{2}^{2}.\\
&=\Delta W_{\text{primary},t}^{(l)}+ \sum^h_{k=1}\Delta W_{t,\text{\rm{noise}}}^{(l,k)}
\label{eq:512}
\end{aligned}
\end{equation}

In addition, if we set {\small $\Gamma_t^{(l)}\!\!\!= {D_{t}^{(l)}V_{t}^{(l+1)}C^{(l+1)^{\top}} C^{(l+1)} \Delta V_{t}^{(l+1)^{\top}} F_{t}^{(l)}} /\|F_{t}^{(l)}\|_{2}^{2}$}, and {\small $\kappa_t^{(l)^\top}\!\!\!=D_{t}^{(l)}$}{\small$ \varepsilon_{t}^{(l+1)}$}{\small $  \Delta \varepsilon_{t}^{(l+1)^\top}F_{t}^{(l)}F_{t}^{(l-1)^{\top}}\!\!\!\! /\|F_{t}^{(l)}\|_{2}^{2}$}. Then we can re-write the Eq. \eqref{eq:512} as follows.
\begin{equation}
	 \Delta W_{t}^{(l)}   \overset{\text{\rm{rewritten}}}{=\!\!=\!\!=\!\!=} \Gamma_t^{(l)} F_{t}^{(l-1)^{\top}}+ \kappa_t^{(l)^\top}
\end{equation}

\subsection{The explanation for the phenomenon that $S_{1}^{(l)}$, $S_{2}^{(l)}$, and $S_{3}^{(l)}$ does not decrease monotonically.}

In this subsection, we explain the phenomenon that {  $S_{1}^{(l)}$}, {  $S_{2}^{(l)}$}, and {  $S_{3}^{(l)}$} does not decrease monotonically in Table \ref{table:norm} in the supplementary material and Table 1 in the main paper (Page 6). In fact, we first decompose {$\varepsilon_{t}^{(l+1)}=\sum_k \varepsilon_{t}^{(l+1, k)}$} according to the SVD. Then $\Delta W_{\text{\rm{noise}},t}^{(l,k)}$ is computed as $ \Delta W_{\text{\rm{noise}},t}^{(l,k)}= D_{t}^{(l)}\varepsilon_{t}^{(l+1,k)} \Delta \varepsilon_{t}^{(l+1)^\top} F_{t}^{(l)} F_{t}^{(l-1)^{\top}} /\left\|F_{t}^{(l)}\right\|_{2}^{2}.$ Accordingly, the strength of weight changes along the primary direction is computed as $S_{\text{\rm{primary}}}^{(l)}=\mathbb{E}_{t\in [T_\text{start}, T_\text{end}]}\mathbb{E}_{x \in X} \left [\|\Delta W_{\text{\rm{primary}},t}^{(l,k)}|_x\|_F \right ]$. The strength of weight changes along the $k$-th noise direction is computed as { $S_{k}^{(l)}=\mathbb{E}_{t\in [T_\text{start}, T_\text{end}]}\mathbb{E}_{x \in X} \left [\|\Delta W_{\text{\rm{noise}},t}^{(l,k)}|_x\|_F \right ]$}. In this way, {  $S_{1}^{(l)}$}, {  $S_{2}^{(l)}$}, and {$S_{3}^{(l)}$} do not decrease monotonically, although  $\|\varepsilon_{t}^{(l+1,1)}\|_F$, $\|\varepsilon_{t}^{(l+1,2)}\|_F$, and $\|\varepsilon_{t}^{(l+1,3)}\|_F$ are directly decomposed from {$\varepsilon_{t}^{(l+1)}$} based on the SVD and decrease monotonically.

\section{Analysis based on Eq. (3) in the main paper and explanation for the parallelism.}

According the Eq. (3) in the main paper, we have
\begin{equation}
\dot F_t^{(l-1)}=(C^{(l)^\top}D_t^{(l)}\dot F_t^{(l)}) \cdot \bm{\beta} +  \bm{\epsilon} D_t^{(l)}\dot F_t^{(l)}
\end{equation}
Thus, if { $C^{(l)^\top}D_t^{(l)}\dot F_t^{(l)}$} is large enough (\emph{i.e.}, keeping optimizing { $W_t^{(l)^\top}$} along the common direction { $ C^{(l)}$} for a long time), then the feature gradients { $ \dot F_t^{(l-1)}$} of different samples will be roughly parallel to the same vector { $\bm{\beta}$}. This is because { $C^{(l)^\top}D_t^{(l)}\dot F_t^{(l)}$} is a scalar and the term { $\bm{\epsilon} D_t^{(l)}\dot F_t^{(l)}$} is small. In other words, the diversity between feature gradients { $\dot F_{t}^{(l-1)}$} of different samples decreases. Here, { $\bm{\beta} = [\beta_1, \beta_2, \cdots, \beta_d]$}, and { $\bm{\epsilon} = [\bm{\epsilon}_1, \bm{\epsilon}_2, \cdots, \bm{\epsilon}_d]^{\top}$}.

\section{Discussion on the background assumption.}
In the above section, we demonstrate that on the ideal state, \emph{i.e.}, { $W_t^{(l)^\top}$} has been optimized towards the common direction { $ C^{(l)}$} for a long time, we can consider that the feature gradients {\small $ \dot F_t^{(l-1)}$} of different samples will be roughly parallel to the same vector {\small $\bm{\beta}$}. In this way, we can explain that the diversity between feature gradients {\small $\dot F_{t}^{(l-1)}$} of different samples decreases. \\
In comparison, in the current section, we mainly discuss the trustworthiness of the background assumption in Line 214 in the main paper. We aim to discuss that on the assumption that features { $F_{t}^{(l-1)}$} of different samples have been pushed a little bit towards a specific common direction, we can find at least one learning iteration in the first phase where $\Delta F_{t}^{(l-1)}$ and $F_{t}^{(l-1)}$ of most samples have similar directions, and $V_t^{(l)}$ and $\Delta V_t^{(l)}$ have similar directions. The assumption that features {\small $F_{t}^{(l-1)}$} of different samples have been pushed a little bit towards a specific common direction is an intermediate state between the chaotic initial state of the MLP and the ideal state introduced in the above section. In this way, we can assume that $C^{(l)^\top}D_t^{(l)} \dot F_t^{(l)}$ is large.\\
According to Eq. (2) in the main paper and Lemma 2, we have $ \dot F_t^{(l-1)} = W_t^{(l)^\top} D_t^{(l)} \dot F_t^{(l)} $ and $W_t^{(l)^\top}=V_t^{(l)}C^{(l)^\top}+\varepsilon_t^{(l)^\top}$. Thus, we have
\begin{equation}
\begin{aligned}
    \dot F_t^{(l-1)} &= W_t^{(l)^\top} D_t^{(l)} \dot F_t^{(l)} \\ &= (V_t^{(l)}C^{(l)^\top}+\varepsilon_t^{(l)^\top}) D_t^{(l)} \dot F_t^{(l)} \\
    & = V_t^{(l)}C^{(l)^\top}D_t^{(l)} \dot F_t^{(l)}+\varepsilon_t^{(l)^\top} D_t^{(l)} \dot F_t^{(l)}
\end{aligned}
\end{equation}
If the scalar $C^{(l)^\top}D_t^{(l)} \dot F_t^{(l)}$ is large, we can roughly consider
\begin{equation}
\begin{aligned}
    &\dot F_t^{(l-1)} \approx V_t^{(l)}C^{(l)^\top} D_t^{(l)} \dot F_t^{(l)} \\ &= V_t^{(l)} \cdot (C^{(l)^\top} D_t^{(l)} \dot F_t^{(l)}) \mathrel{/\mkern-5mu/}  V_t^{(l)}
\end{aligned}
\end{equation}
It means that the feature gradient $\dot F_t^{(l-1)}$ is roughly parallel to the vector $V_t^{(l)}$. Furthermore, the feature gradient $\dot F_t^{(l-1)}$ and the change of feature $\Delta F_t^{(l-1)}$ can be considered negatively parallel to each other, we have
\begin{equation}
    \Delta F_t^{(l-1)} \mathrel{/\mkern-5mu/} \dot F_t^{(l-1)}  \mathrel{/\mkern-5mu/} V_t^{(l)}
\end{equation}
 Similarly, we have $\Delta F_{t+1}^{(l-1)}  \mathrel{/\mkern-5mu/}  V_{t+1}^{(l)}$. Therefore, we can roughly consider that $V_t^{(l)} \approx k_t \Delta F_t^{(l-1)}$, and $V_{t+1}^{(l)} \approx k_{t+1} \Delta F_{t+1}^{(l-1)}$, where $k_t, k_{t+1} \in \mathbb{R}$ are two scalars. Then, we can derive that
\begin{equation}
\Delta V_t^{(l)}=V_{t+1}^{(l)} - V_t^{(l)}\approx k_{t+1}\Delta F_{t+1}^{(l-1)} - k_{t}\Delta F_{t}^{(l-1)}
\end{equation}
If features { $F_{t}^{(l-1)}$} of different samples have been pushed a little bit towards a specific common direction, then it is easy to find at least one learning iteration that $\Delta F_{t}^{(l-1)}$ and $F_{t}^{(l-1)}$ of most samples have similar directions, \emph{i.e.} $\Delta F_{t}^{(l-1)} \mathrel{/\mkern-5mu/} F_{t}^{(l-1)}$. Meanwhile, we can find at least one learning iteration in the first phase where the change of feature in $t$-th iteration $\Delta F_{t}^{(l-1)}$ and $(t+1)$-th iteration $\Delta F_{t+1}^{(l-1)}$ are roughly the same. In other words, $\Delta F_{t}^{(l-1)} \approx \Delta F_{t+1}^{(l-1)}$. 
Thus, we have
\begin{equation}
\Delta V_t^{(l)} \approx (k_{t+1}-k_t)\Delta F_t^{(l-1)} \mathrel{/\mkern-5mu/}  \Delta F_t^{(l-1)} \mathrel{/\mkern-5mu/}  V_t^{(l)}
\end{equation}
In this way, we can obtain that $V_t^{(l)}$ and $\Delta V_t^{(l)}$ have similar directions.

\clearpage
\section{Proof for Lemma 3}
\label{ap:lemma3}
In this section, we present the detailed proof for Lemma 3.
\begin{lemma}
    Given an input sample $x \in X$ and a common direction {\small $C^{(l)}$} after the {$t$}-th iteration, if the noise term {\small $\varepsilon_t^{(l)}$} is small enough to satisfy {\small $|\Delta V_{t}^{(l)^{\top}}F_{t}^{(l-1)}V_{t}^{(l)^{\top}}V_{t}^{(l)}C^{(l)^\top}C^{(l)}\Delta V_t^{(l)^\top} F_{t}^{(l-1)}|\gg|\Delta V_{t}^{(l)^{\top}}F_{t}^{(l-1)}V_{t}^{(l)^{\top}}\varepsilon_t^{(l)}\Delta \varepsilon_t^{(l)^\top}F_{t}^{(l-1)}|$}
	, we can obtain {\small $\cos(\Delta V_{t}^{(l)},F_{t}^{(l-1)}) \cdot \cos(V_{t}^{(l)}, \Delta F_{t}^{(l-1)})\geq0$}, where {\small $\Delta V_{t}^{(l)}=\frac{\Delta W_{t}^{(l)^\top} C^{(l)}}{C^{(l)^\top}C^{(l)}}$}, and {\small $V_{t}^{(l)}=\frac{W_{t}^{(l)^\top} C^{(l)}}{C^{(l)^\top}C^{(l)}}$}. {\small $\Delta F_{t}^{(l-1)}$} denotes the change of features {\small $\Delta F_{t}^{(l-1)}=F_{t+1}^{(l-1)}-F_{t}^{(l-1)}$} made by the training sample $x$ after the {\small $t$}-th iteration. To this end, we approximately consider the change of features {\small $\Delta F_{t}^{(l-1)}$} after the {\small $t$}-th iteration negatively parallel to feature gradients {\small $\dot{F}_{t}^{(l-1)}$}, although strictly speaking, the change of features is not exactly equal to the gradient \emph{w.r.t.} features.
\end{lemma}

\textit{proof.} Given a sample $x$, we can prove that $\cos(\Delta V_{t}^{(l)}, F_{t}^{(l-1)})\cdot \cos(V_{t}^{(l)}, \Delta F_{t}^{(l-1)})\geq 0$.

According to chain rule, we have
\begin{equation}
    \small
    \Delta W_t^{(l)}=-\eta D_t^{(l)} \dot F_t^{(l)} F_t^{(l-1)^T}
\end{equation}
According to Lemma 1 and Lemma 2, we have $C^{(l)^\top}\Delta\varepsilon_{t}^{(l)^\top}=0$ and $\varepsilon_{t}^{(l)}C^{(l)}=0$. Then, we have
\begin{equation}
\small
\cos(\Delta V_{t}^{(l)}, F_{t}^{(l-1)})\cdot \cos(V_{t}^{(l)}, \dot F_{t}^{(l-1)})\\
=\left[\frac{\Delta V_{t}^{(l)^{\top}}F_{t}^{(l-1)}}{\|\Delta V_{t}^{(l)}\| \cdot \|F_{t}^{(l-1)}\|}\right]\cdot \left[\frac{ V_{t}^{(l)^{\top}} \dot F_{t}^{(l-1)}}{\| V_{t}^{(l)}\| \cdot \|\dot F_{t}^{(l-1)}\|}\right]
\end{equation}
Therefore, we have
\begin{equation}
\begin{aligned}
\small
&\text{\rm{sign}}(\cos(\Delta V_{t}^{(l)}, F_{t}^{(l-1)})\cdot \cos(V_{t}^{(l)}, \dot F_{t}^{(l-1)}))\\
&=\text{\rm{sign}}([\Delta V_{t}^{(l)^{\top}}F_{t}^{(l-1)}]\cdot [V_{t}^{(l)^{\top}}\dot F_{t}^{(l-1)}]/(\|\Delta V_{t}^{(l)}\|_2 \|F_{t}^{(l-1)}\|_2 \|V_{t}^{(l)}\|_2 \|\dot F_{t}^{(l-1)}\|_2))\\
&=\text{\rm{sign}}([\Delta V_{t}^{(l)^{\top}}F_{t}^{(l-1)}]\cdot [V_{t}^{(l)^{\top}}W_t^{(l)^\top}D_t^{(l)}\dot F_t^{(l)}]/(\|\Delta V_{t}^{(l)}\|_2 \|F_{t}^{(l-1)}\|_2 \|V_{t}^{(l)}\|_2 \|\dot F_{t}^{(l-1)}\|_2))\\
&=\text{\rm{sign}}([\Delta V_{t}^{(l)^{\top}}F_{t}^{(l-1)}]\cdot [V_{t}^{(l)^{\top}}(V_{t}^{(l)}C^{(l)^\top}+ \varepsilon_{t}^{(l)})D_t^{(l)}\dot F_t^{(l)}]/(\|\Delta V_{t}^{(l)}\|_2 \|F_{t}^{(l-1)}\|_2 \|V_{t}^{(l)}\|_2 \|\dot F_{t}^{(l-1)}\|_2))\\
&=\text{\rm{sign}}([\Delta V_{t}^{(l)^{\top}}F_{t}^{(l-1)}]\cdot [V_{t}^{(l)^{\top}}(V_{t}^{(l)}C^{(l)^\top}+ \varepsilon_{t}^{(l)})(\Delta W_{t}^{(l)} F_{t}^{(l-1)} / (-\eta\left\|F_{t}^{(l-1)}\right\|_{2}^{2}))]\\
&/(\|\Delta V_{t}^{(l)}\|_2 \|F_{t}^{(l-1)}\|_2 \|V_{t}^{(l)}\|_2 \|\dot F_{t}^{(l-1)}\|_2))\\
&=\text{\rm{sign}}([\Delta V_{t}^{(l)^{\top}}F_{t}^{(l-1)}]\cdot [(V_{t}^{(l)^{\top}}V_{t}^{(l)}C^{(l)^\top}+V_{t}^{(l)^{\top}} \varepsilon_{t}^{(l)})\Delta W_{t}^{(l)} F_{t}^{(l-1)}]\\&/(-\eta\left\|F_{t}^{(l-1)}\right\|_{2}^{2}\|\Delta V_{t}^{(l)}\|_2 \|F_{t}^{(l-1)}\|_2 \|V_{t}^{(l)}\|_2 \|\dot F_{t}^{(l-1)}\|_2))\\
&=\text{\rm{sign}}([\Delta V_{t}^{(l)^{\top}}F_{t}^{(l-1)}]\cdot [(V_{t}^{(l)^{\top}}V_{t}^{(l)}C^{(l)^\top}+V_{t}^{(l)^{\top}} \varepsilon_{t}^{(l)})(C^{(l)}\Delta V_t^{(l)^\top} +\Delta \varepsilon_t^{(l)^\top}) F_{t}^{(l-1)}] \\ &/(-\eta\left\|F_{t}^{(l-1)}\right\|_{2}^{2}\|\Delta V_{t}^{(l)}\|_2 \|F_{t}^{(l-1)}\|_2 \|V_{t}^{(l)}\|_2 \|\dot F_{t}^{(l-1)}\|_2))\\
&=\text{\rm{sign}}([\Delta V_{t}^{(l)^{\top}}F_{t}^{(l-1)}]\cdot [(V_{t}^{(l)^{\top}}V_{t}^{(l)}C^{(l)^\top}C^{(l)}\Delta V_t^{(l)^\top} +V_{t}^{(l)^{\top}} \varepsilon_{t}^{(l)}\Delta \varepsilon_t^{(l)^\top}\\&+V_{t}^{(l)^{\top}}V_{t}^{(l)}C^{(l)^\top}\Delta \varepsilon_t^{(l)^\top} +V_{t}^{(l)^{\top}} \varepsilon_{t}^{(l)}C^{(l)}\Delta V_t^{(l)^\top}) F_{t}^{(l-1)}]/(-\eta\left\|F_{t}^{(l-1)}\right\|_{2}^{2}\|\Delta V_{t}^{(l)}\|_2 \|\dot F_{t}^{(l-1)}\|_2 \|V_{t}^{(l)}\|_2 \|F_{t}^{(l-1)}\|_2))  \\
&=\text{\rm{sign}}([\Delta V_{t}^{(l)^{\top}}F_{t}^{(l-1)}]\cdot [(V_{t}^{(l)^{\top}}V_{t}^{(l)}C^{(l)^\top}C^{(l)}\Delta V_t^{(l)^\top} +V_{t}^{(l)^{\top}} \varepsilon_{t}^{(l)}\Delta \varepsilon_t^{(l)^\top}) F_{t}^{(l-1)}]\\ &/(-\eta\left\|F_{t}^{(l-1)}\right\|_{2}^{2}\|\Delta V_{t}^{(l)}\|_2 \|F_{t}^{(l-1)}\|_2 \|V_{t}^{(l)}\|_2 \|\dot F_{t}^{(l-1)}\|_2))\\
&=\text{\rm{sign}}([\Delta V_{t}^{(l)^{\top}}F_{t}^{(l-1)}]\cdot [V_{t}^{(l)^{\top}}V_{t}^{(l)}C^{(l)^\top}C^{(l)}\Delta V_t^{(l)^\top}F_{t}^{(l-1)} +V_{t}^{(l)^{\top}} \varepsilon_{t}^{(l)}\Delta \varepsilon_t^{(l)^\top}F_{t}^{(l-1)}]\\ &/(-\eta\left\|F_{t}^{(l-1)}\right\|_{2}^{2}\|\Delta V_{t}^{(l)}\|_2 \|F_{t}^{(l-1)}\|_2 \|V_{t}^{(l)}\|_2 \|\dot F_{t}^{(l-1)}\|_2))\\
&=\text{\rm{sign}}([\Delta V_{t}^{(l)^{\top}}F_{t}^{(l-1)}V_{t}^{(l)^{\top}}V_{t}^{(l)}C^{(l)^\top}C^{(l)}\Delta V_t^{(l)^\top}F_{t}^{(l-1)} +\Delta V_{t}^{(l)^{\top}}F_{t}^{(l-1)}V_{t}^{(l)^{\top}} \varepsilon_{t}^{(l)}\Delta \varepsilon_t^{(l)^\top}F_{t}^{(l-1)}]\\ &/(-\eta\left\|F_{t}^{(l-1)}\right\|_{2}^{2}\|\Delta V_{t}^{(l)}\|_2 \|F_{t}^{(l-1)}\|_2 \|V_{t}^{(l)}\|_2 \|\dot F_{t}^{(l-1)}\|_2))\\
\end{aligned}
\end{equation}

\begin{figure}[t]
	\centering
	\vspace{-5pt}
	\includegraphics[width=0.98\linewidth]{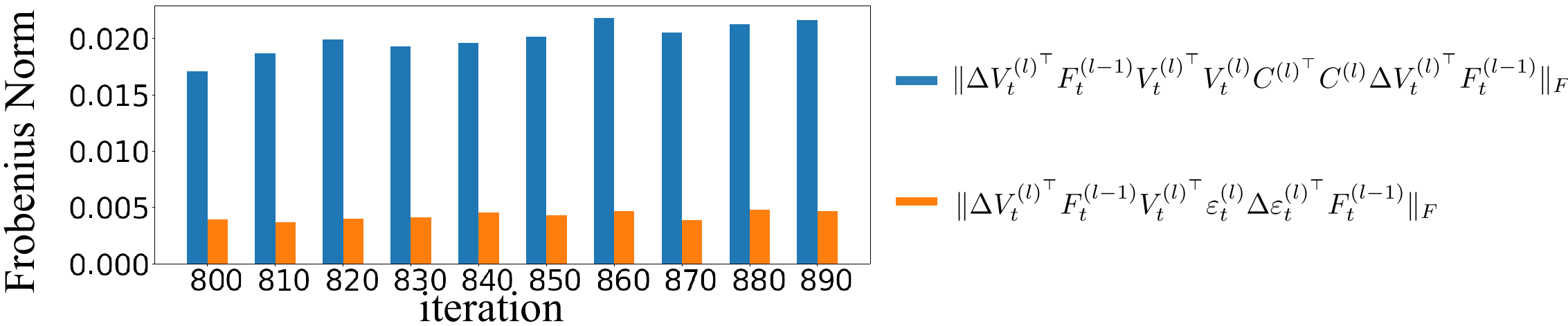}
	\vspace{-5pt}
	\caption{ Visualization of the Frobenius norm of the two components $\Delta V_{t}^{(l)^{\top}}\!\!F_{t}^{(l-1)}V_{t}^{(l)^{\top}}V_{t}^{(l)}C^{(l)^\top}C^{(l)}\Delta V_t^{(l)^\top} \!\!F_{t}^{(l-1)}$ and $\Delta V_{t}^{(l)^{\top}}F_{t}^{(l-1)}V_{t}^{(l)^{\top}}\varepsilon_t^{(l)}\Delta \varepsilon_t^{(l)^\top}F_{t}^{(l-1)}$. We trained a 9-layer MLP on the MNIST dataset, where each layer had 512 neurons. Iterations were chosen at the end of the first phase.}
	\label{fig:singular}

\end{figure}

According to our assumption, the noise term  $\varepsilon_t^{(l)}$ is small enough to satisfy $|\Delta V_{t}^{(l)^{\top}}F_{t}^{(l-1)}V_{t}^{(l)^{\top}}V_{t}^{(l)}C^{(l)^\top}C^{(l)}\Delta V_t^{(l)^\top} F_{t}^{(l-1)}|\gg|\Delta V_{t}^{(l)^{\top}}F_{t}^{(l-1)}V_{t}^{(l)^{\top}}\varepsilon_t^{(l)}\Delta \varepsilon_t^{(l)^\top}F_{t}^{(l-1)}|$. This assumption is verified in Figure ~\ref{fig:singular}. Then we can ignore the last term and obtain
\begin{equation}
    \begin{aligned}
            &\text{\rm{sign}}([\Delta V_{t}^{(l)^{\top}}F_{t}^{(l-1)}V_{t}^{(l)^{\top}}V_{t}^{(l)}C^{(l)^\top}C^{(l)}\Delta V_t^{(l)^\top}F_{t}^{(l-1)} +\Delta V_{t}^{(l)^{\top}}F_{t}^{(l-1)}V_{t}^{(l)^{\top}} \varepsilon_{t}^{(l)}\Delta \varepsilon_t^{(l)^\top}F_{t}^{(l-1)}]
            \\&/(-\eta\left\|F_{t}^{(l-1)}\right\|_{2}^{2}\|\Delta V_{t}^{(l)}\|_2 \|F_{t}^{(l-1)}\|_2 \|V_{t}^{(l)}\|_2 \|\dot F_{t}^{(l-1)}\|_2))
            \\&\approx \text{\rm{sign}}([\Delta V_{t}^{(l)^{\top}}F_{t}^{(l-1)}V_{t}^{(l)^{\top}}V_{t}^{(l)}C^{(l)^\top}C^{(l)}\Delta V_t^{(l)^\top}F_{t}^{(l-1)}]\\&(-\eta\left\|F_{t}^{(l-1)}\right\|_{2}^{2}\|\Delta V_{t}^{(l)}\|_2 \|F_{t}^{(l-1)}\|_2 \|V_{t}^{(l)}\|_2 \|\dot F_{t}^{(l-1)}\|_2))\leq0
            \end{aligned}
\end{equation}
Thus,
\begin{equation}
\text{\rm{sign}}(\cos(\Delta V_{t}^{(l)}, F_{t}^{(l-1)})\cdot \cos(V_{t}^{(l)}, \dot F_{t}^{(l-1)}))\leq 0
\end{equation}
 In this paper, we approximately consider $\Delta F_{t}^{(l-1)}$ and $\dot F_{t}^{(l-1)}$ are negatively parallel to each other. Thus, we have $\text{\rm{sign}}(\cos(\Delta V_{t}^{(l)}, F_{t}^{(l-1)})\cdot \cos(V_{t}^{(l)}, \Delta  F_{t}^{(l-1)}))=\text{\rm{sign}}(\cos(\Delta V_{t}^{(l)}, F_{t}^{(l-1)})\cdot (-\cos(V_{t}^{(l)}, \dot F_{t}^{(l-1)})))\geq 0$.

\section{Proof for Theorem 2}
\label{ap:theorem2}

In this section, we aim to prove that training samples of the same category have the same effect in the first phase.

\begin{theorem}
    Under the background assumption, for any training samples $x, x' \!\in\! X_c$ in the category $c$, if { $[C^{(l)^\top}\!\!\!D_t^{(l)}|_x \dot F_t^{(l)}|_x]\cdot[C^{(l)^\top} D_t^{(l)}|_{x'} \dot F_t^{(l)}|_{x'}]>0$} (means that { $F_t^{(l)}|_{x}$} and { $ F_t^{(l)}|_{x'}$} have kinds of similarity in very early iterations), then { $\cos(\alpha_c \Delta V_{t}^{(l)}|_x, F_{t}^{(l-1)}|_{x})\!\geq\!0$}, and { $\cos(\alpha_c V_{t}^{(l)}\!\!\!, \Delta F_{t}^{(l-1)}|_{x})$ $\geq0$}, where { $\alpha_c\! \in\!\! \{-1,+1\}$} is a constant shared by all samples in category $c$.
\end{theorem}

\textit{proof.} Given a sample $x$ and a sample $x'$ from the same category, we can prove that $\cos(\Delta V_{t}^{(l)}|_x, F_{t}^{(l-1)}|_x)\cdot \cos(\Delta V_{t}^{(l)}|_{x'}, F_{t}^{(l-1)}|_{x'})\geq0$.\\
\begin{equation}
\begin{aligned}
&\text{\rm{sign}}(\cos(\Delta V_{t}^{(l)}|_x, F_{t}^{(l-1)}|_x)\cdot \cos(\Delta V_{t}^{(l)}|_{x'}, F_{t}^{(l-1)}|_{x'}))\\
&=\text{\rm{sign}}([\Delta V_{t}^{(l)^{\top}}|_xF_{t}^{(l-1)}|_x]\cdot[\Delta V_{t}^{(l)^{\top}}|_{x'}F_{t}^{(l-1)}|_{x'}])\\
&=\text{\rm{sign}}([\frac{C^{(l)^\top}\Delta W_{t}^{(l)}|_x}{C^{(l)^\top}C^{(l)}}F_{t}^{(l-1)}|_x]\cdot[\frac{C^{(l)^\top}\Delta W_{t}^{(l)}|_{x'}}{C^{(l)^\top}C^{(l)}}F_{t}^{(l-1)}|_{x'}])\\
&=\text{\rm{sign}}([C^{(l)^\top}\Delta W_{t}^{(l)}|_xF_{t}^{(l-1)}|_x]\cdot[C^{(l)^\top}\Delta W_{t}^{(l)}|_{x'}F_{t}^{(l-1)}|_{x'}])\\
&=\text{\rm{sign}}([C^{(l)^\top}(-\eta D_t^{(l)}|_x \dot F_t^{(l)}|_x
F_t^{(l-1)^\top}|_x)F_{t}^{(l-1)}|_x]\cdot[C^{(l)^\top}(-\eta D_t^{(l)}|_{x'} \dot F_t^{(l)}|_{x'}
F_t^{(l-1)^\top}|_{x'})F_{t}^{(l-1)}|_{x'}])\\
&=\text{\rm{sign}}([C^{(l)^\top} D_t^{(l)}|_x \dot F_t^{(l)}|_x]\cdot[C^{(l)^\top} D_t^{(l)}|_{x'} \dot F_t^{(l)}|_{x'}])\\
\end{aligned}
\end{equation}

According to the assumption that { $F_t^{(l)}|_{x}$} and { $ F_t^{(l)}|_{x'}$} have kinds of similarity, we can consider { $[C^{(l)^\top}\!\!\!D_t^{(l)}|_x \dot F_t^{(l)}|_x]\cdot[C^{(l)^\top} D_t^{(l)}|_{x'} \dot F_t^{(l)}|_{x'}]>0$}. In this way, for the category $c$, there exists a constant $\alpha_c$, which satisfies $\text{\rm{sign}}(  \cos(\alpha_c \Delta V_{t}^{(l)}|_x, F_{t}^{(l-1)}|_x)\geq0$, where $\alpha_c \in \{-1,+1\}$ and training sample $x \in X_c$ belongs to the category $c$.

According to Lemma 3, we have $\cos(\Delta V_{t}^{(l)}|_x, F_{t}^{(l-1)}|_x)\cdot \cos(V_{t}^{(l)}, \Delta F_{t}^{(l-1)}|_x)\geq 0$. Thus, we have $\text{\rm{sign}}(\cos(\alpha_c \Delta V_{t}^{(l)}|_x, F_{t}^{(l-1)}|_x)\cdot \cos(\alpha_c V_{t}^{(l)}, \Delta F_{t}^{(l-1)}|_x))\geq 0$. In addition, the above proof indicates that $\text{\rm{sign}}(  \cos(\alpha_c \Delta V_{t}^{(l)}|_x, F_{t}^{(l-1)}|_x)\geq0$. Therefore, we have $\text{\rm{sign}}( \cos( \alpha_c  V_{t}^{(l)}|_x, \Delta F_{t}^{(l-1)}|_x)\geq0$

\section{Discussion for four typical operations}

\subsection{Normalization}
\label{ap:normalization}

The output feature of the {$l$}-th linear layer \emph{w.r.t.} the input sample $x$ can be described as { $[f_1,f_2,\ldots,f_h]=W_{t}^{(l)}F_{t}^{(l-1)}\in \mathbb{R}^h$}, where { $f_i$} denotes the $i$-th dimension of the feature. In this way, the batch normalization operation can be formulated as { $\textit{BN}(f_i)=\gamma_{scale}[(f_i-\mu_i)/\sigma_{i}]+\beta_{shift}$}, where $\gamma_{scale}$ and $\beta_{shift}$ denote the scaling and the shifting parameters, respectively. In this way, the batch normalization operation subtracts the mean feature { $\bar{F}_t^{(l)}=\mathbb{E}_{x\in X}[F_t^{(l)}|_x]$} from features of all samples. Therefore, features of different samples in a same category are no longer similar to each other.

We also propose a simplified normalization operation to alleviate the decrease of feature diversity in the first phase.
The simplified normalization operation is given as { $\textit{norm}_1(f_i)=(f_i-\mu_i)/\sigma_{i}$}, where {$\mu_i$} and { $\sigma_i$} denote the mean value and the standard deviation of { $f_i$} over different samples, respectively.
This operation is similar to the batch normalization~\cite{ioffe2015batch}, but we do not compute the scaling and shifting parameters in the batch normalization.

\begin{figure}[t]
	\centering
	\includegraphics[width=0.98\linewidth]{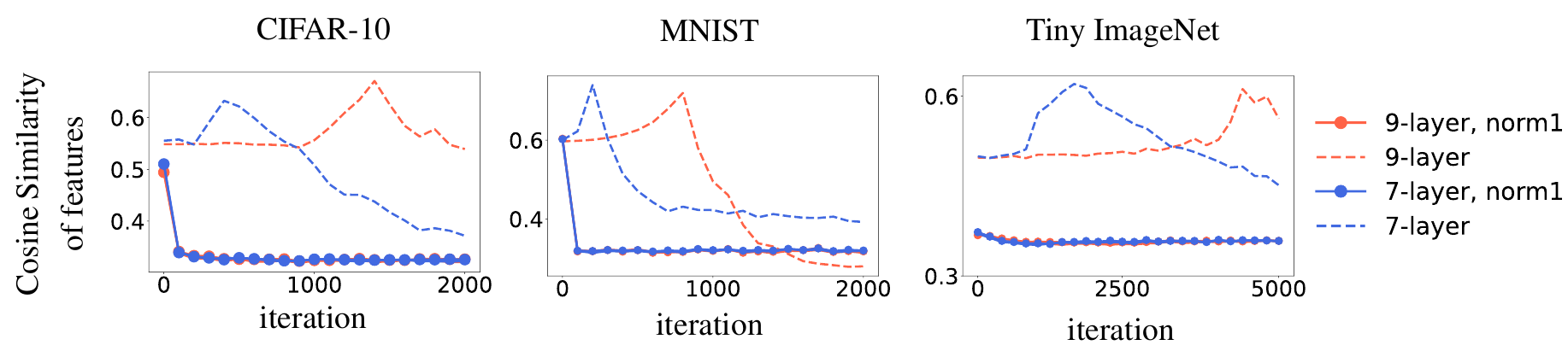}
	\vspace{-5pt}
	\caption{ Cosine similarity of features between samples in different categories. We trained 7-layer MLPs and 9-layer MLPs on the CIFAR-10, the MNIST, and the Tiny ImageNet dataset.}
	\label{fig:norm}
	\vspace{-10pt}
\end{figure}

In order to verify the simplified normalization operation can alleviate the decrease of feature diversity during the training process of the MLP, we trained 7-layer MLPs and 9-layer MLPs with and without the normalization operation. Specifically, for the normalization operation \textit{ $norm_1$}, we added the normalization operation after each linear layer, except the last linear layer. Each linear layer in the MLP had 512 neurons. Figure \ref{fig:norm} shows that the feature similarity in MLPs with normalization operations kept decreasing, while the feature similarity of the MLP without normalization operations kept increasing.
This indicated that normalization operations alleviate the decreasing of feature diversity.

\subsection{Momentum}
\label{ap:momentum}

We can explain that momentum in gradient descent can alleviate this phenomenon. Based on Lemma 3, the self-enhanced system of the decreasing of feature diversity requires singular values of weights along other directions $\varepsilon_{t}^{(l)}$ to be small enough. However, because the momentum operation strengthens influences of the initialized noisy weights $W_{t=0}^{(l)}$, it strengthens singular values of $\varepsilon_{t}^{(l)}$, to some extent, thereby alleviating the decrease of feature diversity.

\begin{figure}[t]
	\centering
	\includegraphics[width=0.98\linewidth]{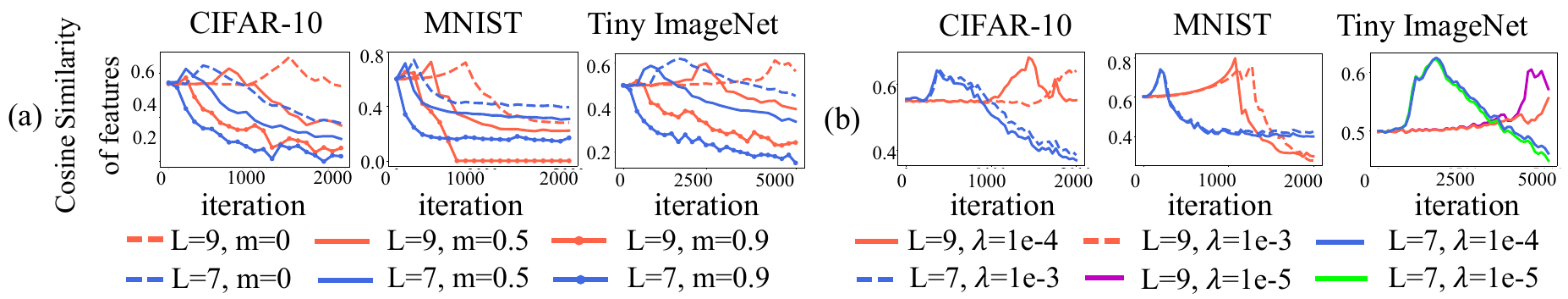}
	\vspace{-5pt}
	\caption{Effects of (a) momentum and (b) $L_2$ regularization. We trained $L$-layer MLPs, where each layer had 512 neurons. A shorter first phase indicates that the decrease of feature diversity is more alleviated.}
	\label{fig:mom_l2}
	\vspace{-10pt}
\end{figure}

Specifically, consider the momentum with the coefficient $m$, the dynamics of weights $W_{t+1}$ can be described as,
\begin{equation}
W_{t+1} = W_t-\eta \frac{\partial Loss}{\partial W_{t}}-m \frac{\partial Loss}{\partial W_{t-1}},
\end{equation}
where $\eta$ denotes the learning rate. Because we only focus on weights in a single layer, without causing ambiguity, we omit the superscript $(l)$ to simplify the notation in this subsection. In this way, we can write the gradient descent as
\begin{equation}
W_{T+1} = W_0+ \eta \sum_{t}^{T} \frac{1-m^{T+1-t}}{1-m} \frac{\partial Loss}{\partial W_{t}}.
\end{equation}
Since $0<m<1$, the coefficient $\frac{1-m^{T+1-t}}{1-m}$ decreases when the variable $t$ increases. Thus, a large $m$ represents that influences of $W_0$ on $W_{T+1}$ are significant. Because $\varepsilon_{T+1}$ is decomposed from $W_{T+1}$ and singular values of $\varepsilon_{T+1}$ are mainly determined by the noisy $W_0$. Accordingly, singular values of $\varepsilon_{T+1}$ are relatively large, which disturb the self-enhanced system and alleviate the decrease of feature diversity. Figure ~\ref{fig:mom_l2}(a) verifies that a larger value of $m$ usually more alleviates the decrease of feature diversity.

\subsection{Initialization}
\label{ap:initialization}

We explain that the initialization of MLPs also affects the decrease of feature diversity. According to Lemma 3, such self-enhanced system requires singular values of weights along other directions $\varepsilon_{t}^{(l)}$ to be small enough. However, because increasing the variance of the initialized weights $W_{0}^{(l)}$ will increase singular values of $\varepsilon_{t}^{(l)}$ based on Lemma 2, alleviating the decrease of feature diversity. Specifically, we initialize weights with Xavier normal distribution \cite{glorot2010understanding}, \emph{i.e.} $W_{0} \sim \mathcal{N}(0,\gamma\sigma_{var}^{2})$, where $\sigma_{var} = \sqrt{\frac{2}{fan_{out}+fan_{in}}}$. $fan_{in}$ and $fan_{out}$ denote the input dimension and the output dimension of the linear layer, respectively. In this way, a large $\gamma$ yields large singular values of initial weights $W_0$. Based on Lemma 2, we also have $\varepsilon_{0}^{(l)}= W_{0}^{(l)^\top}-W_{0}^{(l)^\top}\frac{C^{(l)} C^{(l)^\top}}{C^{(l)^{\top}} C^{(l)}}$. Large singular values of initial weights $W_0$ lead to large singular values of $\varepsilon_{0}^{(l)}$. Therefore, a large variance of initialized weights disturbs the self-enhanced system and alleviates the decrease of feature diversity.

\subsection{$L_2$ regularization}
\label{ap:regularization}
$L_2$ regularization is equivalent to the weight decay in the case of gradient descent. The loss function with $L_2$ regularization and cross entropy loss can be formulated as
{ $\mathcal{L}_{t}(W_t)=\mathcal{L}^{CE}_{t}(W_t)+ \lambda\|W_t\|_{2}^{2}$}, where $W_t$ denotes weights of MLPs.
In this way, we have the following iterates by using gradient descent
\begin{align}
 	W_{t+1} &= W_{t}-\eta \nabla \mathcal{L}_{t}\left(W_{t}\right)
 	\nonumber \\
	&=W_{t}-\eta \nabla \mathcal{L}^{CE}_{t}\left(W_{t}\right)-2\eta \lambda W_{t} \nonumber \\
 	&=(1-2\eta\lambda) W_{t}-\eta \nabla \mathcal{L}^{CE}_{t}\left(W_{t}\right), \label{eq:1125}
\end{align}
According to Lemma 3, such self-enhanced system requires singular values of weights along other directions $\varepsilon_{t}^{(l)}$ to be small enough. Based on Lemma 2, we also have $\varepsilon_{t}^{(l)}= W_{t}^{(l)^\top}-W_{t}^{(l)^\top}\frac{C^{(l)} C^{(l)^\top}}{C^{(l)^{\top}} C^{(l)}}$. In this way, a smaller $\lambda$ yields larger singular values of $\varepsilon_{t}^{(l)}$, which disturbs the self-enhanced system and alleviates the decrease of feature diversity. Figure ~\ref{fig:mom_l2}(b) Figure 9(d) verifies that a smaller coefficient $lambada$ more alleviated
the decrease of feature diversity.

\end{document}